\documentclass[10pt]{article} %
\usepackage[accepted]{tmlr}

\usepackage{amsmath,amsfonts,bm}

\def\eqref#1{equation~\ref{#1}}

\def\1{\bm{1}}

\DeclareMathAlphabet{\mathsfit}{\encodingdefault}{\sfdefault}{m}{sl}
\SetMathAlphabet{\mathsfit}{bold}{\encodingdefault}{\sfdefault}{bx}{n}

\usepackage{graphicx} %
\usepackage[utf8]{inputenc} %
\usepackage[T1]{fontenc}    %
\usepackage{hyperref}       %
\usepackage{url}            %
\usepackage{booktabs}       %
\usepackage{amsfonts}       %
\usepackage{nicefrac}       %
\usepackage[protrusion=true,expansion=true]{microtype}      %
\usepackage{subcaption}
\usepackage{wrapfig}
\usepackage{tabularx}
\usepackage[export]{adjustbox}
\usepackage{color,soul}
\usepackage{hyperref}
\usepackage{xcolor}
\usepackage{enumitem}
\usepackage{setspace}
\usepackage{algorithm}
\usepackage{amsmath}
\usepackage{amsthm}
\usepackage{algpseudocode}
\usepackage{booktabs}       %
\usepackage{xspace}         %
\usepackage{longtable}
\usepackage{siunitx}
\usepackage{cleveref}
\usepackage{listings}
\usepackage{multibib}
\usepackage{seqsplit}
\newcites{Supp}{Appendix References}

\usepackage{thm-restate}
\usepackage[colorinlistoftodos,prependcaption,textsize=small]{todonotes}

\newtheorem{lemma}{Lemma}

\newcommand{\methodname}{\textsc{StructureFlow}\xspace}
\newcommand{\sfm}{$[$SF$]^2$M\xspace}
\newcommand{\best}[1]{\textbf{#1}}        %
\newcommand{\second}[1]{\underline{#1}}%

\makeatletter
\let\rev\relax  %
\DeclareRobustCommand{\rev}{\@ifnextchar2{\rev@two}{\rev@old}}
\long\def\rev@old#1{#1}
\long\def\rev@two2#1{\textcolor{purple}{#1}}
\makeatother
\definecolor{rev2color}{rgb}{0.0,0.45,0.45}

\newcommand{\diff}{\mathrm{d}}

\title{Simulation-free Structure Learning for Stochastic Population Dynamics}

\author{Noah El Rimawi-Fine$^{1,2}$\thanks{Joint first authorship.
$^\dagger$Correspondence to: \url{atanacko@ualberta.ca}. \\ \parindent 1.75em\indent Our code is available at: \url{https://github.com/NoahElRimawiFine/StructureFlow}}
\quad Adam Stecklov$^{1,2*}$ \quad Lucas Nelson$^{1,2}$
\quad Alexander Tong$^{2,3,4}$
\\[0.3em]
Mathieu Blanchette$^{1,2}$
\quad Stephen Y. Zhang$^{5}$
\quad Lazar Atanackovic$^{6,7,8,\dagger}$
\\[0.9em]
{\normalfont\small
$^1$McGill University \quad $^2$Mila -- Quebec AI Institute \quad $^3$Universit\'{e} de Montr\'{e}al  \quad $^4$AITHYRA \\
$^5$University of Melbourne
\quad $^6$University of Alberta \quad
$^7$Alberta Machine Intelligence Institute \\ $^8$Canada CIFAR AI Chair
}}

\def\month{8}  %
\def\year{2026} %
\def\openreview{\url{https://openreview.net/forum?id=Kj9cibMkV3}} %

\begin{document}
\maketitle

\begin{abstract}
Modeling dynamical systems and unraveling their underlying structural dependencies is central to many domains in the natural sciences. Various physical systems, such as those arising in cell biology, are inherently high-dimensional and stochastic in nature, and admit only partial, noisy state measurements. Our primary motivating setting is single-cell biology, where destructive measurements yield unpaired population snapshots rather than longitudinal trajectories of the same cells. This poses a significant challenge for addressing the problems of modeling the underlying dynamics and inferring the network structure of these systems. Existing methods are typically tailored either for structure learning or modeling dynamics at the population level, but are limited in their ability to address both problems together. In this work, we address both problems simultaneously: we present \methodname, a novel and principled simulation-free training approach for jointly learning the structure and stochastic population dynamics of physical systems. We showcase the utility of \methodname for the tasks of structure learning from interventions and dynamical (trajectory) inference of conditional population dynamics. We empirically evaluate our approach on high-dimensional synthetic systems, a set of biologically plausible simulated systems, and an experimental single-cell dataset. 
We show that \methodname can learn the structure of underlying systems while simultaneously modeling their conditional population dynamics --- a key step toward model-based mechanistic understanding of systems behavior.
\end{abstract}

\vspace{-10pt}
\begin{figure}[h]
    \centering
    \includegraphics[width=\textwidth, height=\textheight, keepaspectratio]{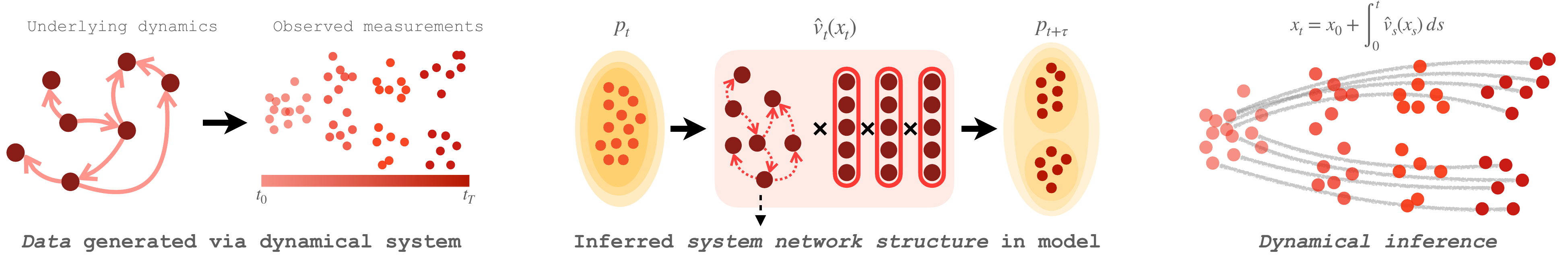}
    \caption{\textbf{Overview of \methodname for joint structure learning and dynamical inference.} Left: an underlying system, described by a directed interaction graph, generates the (visible) data; destructive measurement yields unpaired population snapshots at times $t_0,\ldots,t_T$, not individual trajectories. Middle: from snapshots $p_t$ and $p_{t+\tau}$, \methodname learns an autonomous drift $\hat{v}(x_t)$ from a neural graphical model whose first layer encodes the interaction network. The network structure is recovered directly from the learned weights. Right: integrating the learned field, $x_t = x_0 + \int_0^t \hat{v}_s(x_s)\,ds$, gives continuous trajectories. The network is read directly from the model, so structure and dynamics are recovered jointly.}
    \label{fig:structure_flow}
    \vspace{-10pt}
\end{figure}

\section{Introduction}
Unraveling the underlying structure of physical systems from their stochastic dynamics, given noisy and partial measurements, is a central problem in many areas of science. 
Many physical systems, notably in cell and molecular biology, operate in out-of-equilibrium regimes, are high-dimensional, and are subject to intrinsic stochasticity \citep{elowitz2002stochastic, schiebinger2019optimal}. 
This poses a significant challenge for the task of deciphering the underlying system structure and modeling the resulting dynamics. 
Effectively addressing this problem is a crucial step toward gaining a mechanistic understanding of systems and the ability to predict their behavior under natural and perturbational conditions. This would, in turn, provide practitioners with a map for how to control and guide a system's response towards desirable states \citep{dixit2016perturb, norman2019exploring}. 
In cellular and molecular biology, such a tool would have significant implications in advancing our ability to interpret and model cell fate, development, response to disease~\citep{Rizvi2017, Binnewies2018, Gulati2020, Mol2021}, predict perturbational responses in tissues and patients~\citep{ji_machine_2021,bunne2023learning,peidli2024scperturb,atanackovic2024meta}, and facilitate experimental design~\citep{zhang2023active, huang2024sequential}.

Currently, there exist two broad classes of approaches for deciphering the stochastic dynamics of physical systems from \textit{data}. The first class, which we refer to as \textbf{dynamical inference}\footnote{This task is also commonly labeled and referred to as \emph{trajectory inference}~\citep{hashimoto_learning_2016,weinreb_fundamental_2018,schiebinger2019optimal, tong2020trajectorynetdynamicoptimaltransport, neklyudov2022action}. Since the concept of ``trajectory'' has been frequently used in the literature to refer to several non-equivalent things, in this work, we prefer to name this task \emph{dynamical inference}.}, commonly deals with constructing an estimate of the underlying vector field from partial and noisy measurements. 
\rev{From a good estimate of the vector fields, qualitative properties of dynamics can be interpreted, valuable for understanding cell dynamics and cell fate \citep{qiu2022mapping, macarthur2022geometry}.} 
\rev{In this work, we focus on the preceding task of learning good estimates of vector fields from partially observed population snapshot data.}

The second (and in our work, concurrent) class, \textbf{structure learning}\footnote{Also commonly referred to in literature as a \emph{network inference}, \emph{causal discovery}, and \emph{system identification}.}, aims to reconstruct the \textit{directional} relationships between each of the variables of a given system. Throughout the paper, the learned graph should be interpreted as a directed structural-dependency estimate under the Neural Graphical Model (NGM) and population-dynamics assumptions that are discussed in Section~\ref{subsec:structural_dependencies} and Appendix~\ref{ap:ngm_architecture}, rather than as unrestricted causal recovery from snapshots alone. Existing methods are typically designed to tackle either task in isolation and are limited in their ability to address both problems together~\citep{zheng2018dags, huynh2018dynGENIE3, gao2018sincerities, brouillard2020differentiable, atanackovic2023dyngfn, zhang2024joint}. We posit that knowledge of the underlying vector field is advantageous for this task, and thus we address both problems simultaneously under a single-model.

A key feature of many of the systems we wish to model is that the measurement is \emph{destructive}. For example, profiling a cell via
single-cell RNA sequencing requires lysing the cell, precluding repeated observation of
the same individual over time. Experimental data therefore arrive as population-level
\emph{snapshots} at discrete timepoints, with no temporal correspondence maintained
between cells across timepoints \citep{schiebinger2019optimal,
weinreb_fundamental_2018}. While snapshot data is the canonical regime for single-cell biology, it also arises whenever longitudinal sampling is impractical, e.g., in costly survey designs or other settings with destructive readouts. We note that, while the methodology we develop is generally applicable to underdamped Langevin systems, the experimental scope of this work is centered on the single-cell setting.

To achieve this, we propose \methodname, a principled and \emph{simulation-free} training method for joint dynamical inference and structure learning. 
Building on recent advances in score and flow matching ([SF]²M)~\citep{tong2023simulation} and entropy-regularized optimal transport (EOT)~\citep{cuturi_sinkhorn_2013,shi_diffusion_2023}, \methodname learns a probability flow ordinary differential equation (PF-ODE) from snapshot data. This allows it to model the continuous evolution of a system's population while simultaneously inferring the underlying network structure embedded directly within the model's parameterization. The novelty is the integration and specialization of score/flow matching, an NGM-based structural drift, and intervention-conditioned population modeling for the joint snapshot setting; we do not claim that intervention masks or NGMs alone are new.

The simulation-free framing of \methodname is critical for high-dimensional systems common in cellular biology, as it avoids computationally expensive trajectory simulations during training. This is distinct from inference-time rollout, where we integrate the learned probability-flow ODE (PF-ODE) or stochastic differential equation (SDE) to evaluate held-out distributions. To best facilitate the joint learning task, we introduce a novel parameterization: an autonomous (time-independent) vector field represented by a Neural Graphical Model (NGM)~\citep{ bellot2021neural, dinh2020consistent} which captures the stationary system structure, while a time-dependent score function captures the evolving stochastic dynamics. We present an overview of our framework in Figure~\ref{fig:structure_flow} and outline our core contributions below: 
\begin{itemize}[noitemsep,topsep=0pt,parsep=0pt,partopsep=0pt,leftmargin=*]
    \item We formulate joint dynamical inference and structure learning as a \rev{multi-timepoint} Schr\"odinger Bridge (SB) problem.
    
    \item We introduce \methodname, a novel simulation-free training approach tailored for simultaneously learning the underlying network structure and conditional population dynamics of a stochastic dynamical system from noisy and partial observations.
    
    \item We construct a comprehensive empirical evaluation for the joint inference task over an assortment of systems: (i) on high-dimensional synthetic systems, (ii) a collection of biologically plausible simulated systems, and (iii) an experimental single-cell dataset with genetic interventions. We use our evaluation pipeline to showcase the application of \methodname for the joint structure learning and dynamical inference tasks, all while building an extensive benchmark of state-of-the-art methods.
\end{itemize}

\vspace{-5pt}
\section{Background and Preliminaries}
\label{sec:prelims}

\subsection{The Single-Cell Setting}
As single-cell biology is our primary application, we briefly discuss and connect it to the formalism we use.
  Single-cell RNA sequencing~\citep{macosko2015dropseq, klein2015indrop} measures the expression of $d \gg 1$ genes for each cell in a sampled population. Each measured cell generates a vector of gene expression levels, represented as a point $\bm{x} \in \mathbb{R}^d$ in ``gene expression space''. A cell population can therefore be viewed as an empirical distribution, or point cloud, in this space \citep{schiebinger2019optimal}, and its evolution over time can be modeled as a stochastic process, such as the SDE in~\eqref{eq:sde}. Under this model, the population observed at any instant $t$ is the time marginal of the process at $t$.
  Because sequencing a cell destroys it, the same cell cannot be measured twice; experiments therefore yield \emph{unpaired} population snapshots at discrete timepoints, rather than tracked single-cell trajectories.
  This gives rise to the statistical problem of recovering dynamics from snapshot data described below.
  Perturbational single-cell experiments add controlled changes to this snapshot setting: rather than observing only a natural time course, one also observes populations after specific genes are perturbed and compares how their distributions evolve relative to the unperturbed condition.
  Genetic interventions (e.g. CRISPR knockouts~\cite{jinek2012crispr, dixit2016perturb}) thus define the conditions $c$; recovering the directed graph $\bm{A}$ corresponds to gene regulatory network inference, and dynamical inference corresponds to predicting how the population evolves over time and under perturbation.

\vspace{-5pt}
\subsection{Problem: Modeling Stochastic Population Dynamics}

\vspace{-5pt}
We consider a dynamical model in $\mathbb{R}^d$ described by the stochastic differential equation (SDE):
\begin{equation}
    \diff \bm{x}_t = \bm{v}_t(\bm{x}_t) \diff t + \bm{\sigma} \diff \bm{B}_t,  
    \label{eq:sde}
\end{equation}
where $\bm{x}_t \in \mathbb{R}^d$ is the state at time $t$, $\bm{v}_t : [0, 1] \times \mathbb{R}^d  \to \mathbb{R}^d$  is the drift, $\diff \bm{B}_t$ are Brownian motion increments, and $\bm{\sigma} \in \mathbb{R}^{d \times d}$ is the diffusion coefficient matrix. For simplicity, we shall assume that $\bm{\sigma} = \sigma \mathbf{I}$ for some $\sigma > 0$, although the concepts we introduce can all be generalized to the setting of anisotropic noise (see Appendix~\ref{ap:anisotropic} for details).

Given $\bm{x}_0 \sim p_0$, where $p_0$ is a density over $\mathbb{R}^d$, the dynamics of \eqref{eq:sde} gives rise to a family of \emph{marginals} $(p_t)_{t \in [0, 1]}$, where $p_t$ denotes the marginal probability distribution of the random variable $\bm{x}_t$ at time $t$, which are characterized by the accompanying \emph{Fokker--Planck equation} (FPE):
\begin{equation}
    \partial_t p_t = - \nabla \cdot (p_t(\bm{x}_t) \bm{v}_t(\bm{x}_t)) + \textstyle\frac{1}{2}\nabla \cdot (\sigma^2 \nabla p_t(\bm{x}_t)).
    \label{eq:fpe}
\end{equation}  
The solution to the FPE at time $t$, $p_t(\bm{x})$, gives the probability density of the population at time $t$, starting from $p_0(\bm{x})$. Importantly, \eqref{eq:fpe} can be reformulated as an equivalent \emph{probability flow} ordinary differential equation (ODE) \citep{song2021scorebasedgenerativemodelingstochastic, Maoutsa_2020}:
\begin{equation}
  \partial_t p_t(\bm{x}) = -\nabla \cdot \left( p_t(\bm{x}) \bm{u}_t(\bm{x}) \right), \qquad \bm{u}_t(\bm{x}) = \bm{v}_t(\bm{x}) - \textstyle\frac{1}{2} \sigma^2 \nabla_{\bm{x}} \log p_t(\bm{x}).  \label{eq:PFODE}
\end{equation}
In the above, $\bm{u}_t$ is the \emph{probability flow field} and is related to the drift of the SDE \eqref{eq:sde} up to the addition of a term involving the score, $\bm{s}_t(\bm{x}) = \nabla_{\bm{x}} \log p_t(\bm{x})$. As will be apparent in Section \ref{sec:sbp}, the probability flow formulation of the FPE \eqref{eq:fpe}, and hence the SDE \eqref{eq:sde}, opens up an avenue for addressing dynamics inference without the need for costly simulation-based methods. 

\paragraph{Inference problem setup.}
We consider a setting where empirical snapshot observations are available at multiple timepoints $t_0, \ldots, t_{T-1}$ under different conditions $c \in \mathcal{C}$, yielding a set of empirical marginal distributions \smash{$\{ \hat{p}^{(c)}_{t_i} \}_{i=0}^{T-1}$} over $\bm{x}_{t_i} \in \mathbb{R}^d$ for each condition $c$. %
Each marginal comprises $N_i$ i.i.d. samples assumed to arise from some unknown SDE of the form~\eqref{eq:sde}. We assume the noise level $\bm{\sigma}$ is known, while the vector field $\bm{v}_t$ is unknown\footnote{Throughout we \emph{assume} this data‑generation model holds for every condition $c$.}.
From here, our objective is twofold: \textbf{(1; \textit{dynamical inference})} approximate the (\textit{conditional}) vector field $\bm{v}_t$ of the underlying stochastic dynamics, and \textbf{(2; \textit{structure learning})} recover a directed graph $\bm{A} \in \mathbb{R}^{d \times d}$ that captures dependencies among $d$ variables and represents the underlying data generative process of the system.
We formulate this joint inference task through the lens of Schrödinger Bridges.

\subsection{Simulation-Free Schr\"odinger Bridges via Score and Flow Matching}\label{sec:sbp}

The Schrödinger Bridge Problem (SBP) is concerned with finding the most likely stochastic evolution transporting a source density $q_0$ to a target density $q_1$, given a reference process that encodes prior knowledge of the dynamics. In its dynamical form, the SBP is typically formulated in terms of
\emph{laws of stochastic processes}, i.e.\ probability measures on the path
space $C([0,1],\mathbb{R}^d)$ that describe the distribution of entire sample
trajectories. Writing $\mathbb{P}$ to be the law of a process transporting $q_0$ to $q_1$ and $\mathbb{Q}$ to be the reference measure, we seek a stochastic process $\mathbb{P}^*$ satisfying:
\begin{equation}\label{eq:sbp}
    \mathbb{P}^* = \underset{\mathbb{P}:p_0=q_0,p_1=q_1}{\arg\min} \text{KL}(\mathbb{P}\|\mathbb{Q})
\end{equation}
for marginals $p_t$ of $\mathbb{P}$, and $\text{KL}(\mathbb{P}\|\mathbb{Q}) = \int \diff \mathbb{P} \log(\diff \mathbb{P} / \diff \mathbb{Q})$ is the Kullback--Leibler divergence. When the reference process $\mathbb{Q}$ is the Brownian motion, a key result from Schrödinger Bridge theory \citep{follmer1988} is that the solution of \eqref{eq:sbp} takes the form of a mixture of Brownian bridges $\mathbb{Q}_{xy}$ with respect to a \emph{coupling} $\pi$ of the distributions $(q_0, q_1)$:
\begin{equation}\label{eq:eot}
    \mathbb{P}^\star = \int \mathbb{Q}_{xy} \diff \pi(x, y) .
\end{equation}
Furthermore, the SBP coupling $\pi$ amounts to the solution of an entropic optimal transport problem \citep{leonard2014survey}:
\begin{equation}
    \pi^\star = {\arg\min}_{\pi \in \Pi(q_0, q_1)} \: \textstyle{\frac{1}{2}} \mathbb{E}_{\pi} \| x - y \|_2^2 + \varepsilon \text{KL}(\pi | q_0 \otimes q_1) 
\end{equation}
with regularization parameter $\varepsilon = \sigma^2 > 0$, where $\Pi(q_0, q_1)$ denotes the set of admissible couplings between $q_0$ and $q_1$. In practice when $(q_0, q_1)$ are discrete distributions, this can be solved extremely efficiently using the Sinkhorn algorithm \citep{cuturi_sinkhorn_2013}. 
While the coupling $\pi$ is easy to compute from samples, finding a solution to the dynamical SBP in continuous time is desirable for modelling continuous dynamics. With a notable exception being the case of Gaussian measures \citep{bunne_recovering_2022} where analytical expressions are available, the dynamical SBP \eqref{eq:sbp} does not admit a straightforward solution for $\bm{v}_{\mathrm{SB}}(t, \bm{x})$, defined as the drift of the SDE, $\diff \bm{x}_t = \bm{v}_{\mathrm{SB}}(t, \bm{x}_t) \diff t + \sigma \diff \bm{B}_t$, which underlies the Schrödinger Bridge process $\mathbb{P}$.

Numerous works aim to build approximations to $\bm{v}_{\mathrm{SB}}$  and hence solutions to the dynamical SBP~\citep{chen_likelihood_2022,debortoli2021diffusion,shi_diffusion_2023,tong2023simulation}. In particular, \citet{tong2023simulation} proposes to use score matching \citep{JMLR:v6:hyvarinen05a} and flow matching \citep{lipman2023flowmatchinggenerativemodeling} ([SF]$^2$M) as a natural methodology for building the approximation to $\bm{v}_{\mathrm{SB}}$ in order to solve for $\mathbb{P}$. Note $\bm{v}_{\mathrm{SB}}(t, \bm{x})$ simply corresponds to the drift $\bm{v}_t$ defined by the SBP.
The key observation is to leverage the reciprocal process characterization of $\mathbb{P}$ \citep{leonard2014survey} as a mixture of Brownian bridges, and the fact that Brownian bridges conditioned on endpoints $(\bm{x}_0, \bm{x}_1) =: \bm{z}$ admit closed form expressions for their probability flow $\bm{v}^\circ_t(\bm{x} | \bm{z})$ and score $\nabla \log p_t(\bm{x}|\bm{z})$:
\begin{equation}
    \bm{v}^\circ_t(\bm{x}|\bm{z}) = \frac{1 - 2t}{2t (1-t)} (\bm{x} - (t \bm{x}_1 + (1 - t) \bm{x}_0)) + (\bm{x}_1 - \bm{x}_0); \quad \nabla \log p_t(\bm{x}|\bm{z}) = \frac{t \bm{x}_1 + (1 - t) \bm{x}_0 - \bm{x}}{\sigma^2 t (1 - t)}. 
\end{equation}
Together, these provide the probability flow characterisation of the Brownian bridge. 
With this in hand, and leveraging the fact that the dynamical SBP can be constructed as a mixture of Brownian bridges (\eqref{eq:eot}), \citet{tong2023simulation} propose to construct a neural approximation to the SB flow field \smash{$\bm{v}^\theta(t, \bm{x}) \approx \bm{v}_t^\circ(\bm{x}) := \bm{v}_{SB}(t, \bm{x}) - \tfrac{\sigma^2}{2} \nabla \log p_t(\bm{x})$}. 
The flow field $\bm{v}^\theta(t,\bm{x})$ is trained together with a score field $\bm{s}^\theta(t, \bm{x})$ using the conditional score and flow matching loss
\begin{equation}
    \mathcal{L}_{\mathrm{[SF]^2M}}(\theta) = \mathbb{E}_{t, \bm{z}, \bm{x}} \left[ \|\bm{v}^\theta_t(\bm{x}) - \bm{v}^\circ_t(\bm{x}|\bm{z})\|^2 +\ \lambda(t) \|\bm{s}^\theta_t(\bm{x}) - \nabla \log p_t(\bm{x}|\bm{z})\|^2 \right].
    \label{loss}
\end{equation}
In the above, $\lambda (t) > 0$ weighs the score as a function of time $t$, and $\bm{z} := (\bm{x}_0, \bm{x}_1) \sim \pi(\bm{x}_0, \bm{x}_1)$ where $\pi$ is the entropic OT coupling in \eqref{eq:eot} obtained via the Sinkhorn algorithm. Following Proposition 3.4 of \citet{tong2023simulation}, minimising $\mathcal{L}_{\mathrm{[SF]^2M}}(\theta)$ then approximates the solution to the SBP in \eqref{eq:sbp} under mild conditions. Additionally, \citet{chen2023deepmomentummultimarginalschrodinger, theodoropoulos2025momentummultimarginalschrodingerbridge, rohbeck2025modeling} model a multi-marginal SBP, where successive time-marginal couplings are not computed independently, but rather globally across all timepoints.

\section{\methodname: Joint Inference of Structure and Dynamics}
\label{sec:method}

Our objective is to jointly infer the network structure and population dynamics of the underlying stochastic system. \methodname achieves this in a novel simulation-free manner during training. In this section we introduce the central components of our framework, namely: \textbf{(i)} improved parameterization tailored for the joint task, \textbf{(ii)} modeling conditional stochastic population dynamics from interventional data, and \textbf{(iii)} simulation-free training for learning conditional stochastic population dynamics. 

\subsection{Structural Vector Field Parameterization}
\label{subsec:paramaterization_auto}

We consider a time-independent (or \emph{autonomous}) vector field, which models the underlying structure of a system, and a time-dependent score function accounting for stochasticity. 
We assume that the underlying system structures (graphs) are stationary between adjacent marginals \smash{$(\hat{p}^{(c)}_{t_i}, \hat{p}^{(c)}_{t_{i+1}})$}. We posit that this assumption yields an easier \emph{joint} inference problem, allowing us to parameterize \textit{a single} structure within the autonomous vector field.

\paragraph{Autonomous vector field and time-dependent score parameterization.} From \eqref{eq:sde} we seek to recover the autonomous vector field $\bm{v}$, while the objective in \citep{tong2023simulation} targets the SB flow $\bm{v}^{\circ}_t$.
The quantities of the ground truth process \eqref{eq:sde} are related via \smash{$\bm{v}^{\circ}_t(\bm{x}_t) = \bm{v}(\bm{x}_t) - \frac{\sigma^2}{2}\nabla_{\bm{x}_t} \log p_t(\bm{x}_t)$}, where $\bm{v}^{\circ}_t, \nabla \log p_t$ are the probability flow field and the score of the Brownian bridge pinned at $(t_i, x_i), (t_{i+1}, x_{i+1})$ respectively.
We therefore make the parametrization choice \smash{$\hat{\bm{v}}_t(\bm{x}_t) = \bm{v}^{\theta}(\bm{x}_t) - \frac{\sigma^2}{2} \bm{s}^\phi_t(\bm{x}_t)$}. Since time-independent vector fields, $\bm{v}$, are a function of state, and not spurious temporal correlations, they are better suited for modeling underlying structures which are stationary over time, as we assume. This leads to a more principled modeling approach.

\paragraph{Modeling dynamic structural dependencies via neural graphical vector fields.} 
\label{subsec:structural_dependencies}
We consider a neural structural model following the definition of \cite{bellot2021neural}, where we assume there exists functions $\bm{v}_1, \dots, \bm{v}_d$ such that for $j = 1, \dots, d$, and $\bm{v}_j : \mathcal{X}^d \rightarrow \mathbb{R}$, where $\mathcal{X}^d$ is a bounded subset of $\mathbb{R}^d$. 
We can rewrite \eqref{eq:sde} under the neural dynamic structural model formalism as
\begin{equation}
    \diff x_j(t) = \bm{v}_j(\bm{x}(t)) \diff t + \bm{\sigma}_j \diff B_j(t), \quad \bm{x}(0) \sim p_0,
    \label{eq:structural_sde}
\end{equation}
where $\diff x_j(t)$ denotes the structural dependency for the instantaneous rate of change of the $j^{\text{th}}$ variable dependent on all state observations $\bm{x}$ at time $t$.\footnote{We note that we are using $\bm{x}(t)$ interchangeably with $\bm{x}_t$.} We use an NGM \citep{bellot2021neural} to parameterize the dynamic structural relationships defined in \eqref{eq:structural_sde}.

NGMs are a class of neural graphical model parameterizations designed to represent complex variable (features) dependencies in a computationally efficient manner \citep{shrivastava2023neural}. \cite{bellot2021neural} extend this formulation for the structure learning of continuous-time dynamical systems, motivating their application in our framework.
The NGM lets us parameterize the structural relationships of a system with $d$ variables directly within the \emph{autonomous} vector field $\bm{v}(\bm{x}_t)$.  We use an NGM to approximate the \emph{autonomous} component of the ground truth process described in \eqref{eq:sde} ($\bm{v}(\bm{x}_t)$ instead of $\bm{v}_t(\bm{x}_t)$). The NGM parameterization of the \emph{autonomous} vector field is defined as: 
\begin{equation}
    \bm{v}^{\theta_j}(\bm{x}_t) = \psi(\cdots \psi(\psi(\bm{x}_t \theta_j^{A}) \theta_j^{1}) \cdots)\theta_j^{K}, \quad j = 1, \dots, d,
    \label{eq:dyn_ngm}
\end{equation}
where \smash{$\theta_j^A \in \mathbb{R}^{d \times h}$} denotes a weight matrix (or graph layer) representing the dynamic dependencies between $\diff x_j(t)$ and $\bm{x}(t)$, \smash{$\theta_j^k \in \mathbb{R}^{h \times h}$} for $k = 1, \dots, K-1$ are the weight matrices of each corresponding hidden layer, $\theta^K_j \in \mathbb{R}^{h \times 1}$ is the final layer's weight matrix yielding a scalar output, and $\psi(\cdot)$ is an activation function. We assume variable dependencies defined by \smash{$\theta_j^A$} are sparse and enforce sparsity on \smash{$\theta_j^A$} using group lasso regularization during training. 
We include further details regarding the NGM and its optimization for dynamic structure learning in Appendix \ref{ap:ngm_architecture}\rev{, as well as principled theoretical justification for recovering the underling network structure in Appendix \ref{ap:identify}.}

\subsection{Modeling Conditional Population Dynamics}
\label{subsec:intervention_model}

\methodname explicitly learns conditional stochastic dynamics (trajectories) using both interventional and observational data and simultaneously infers the underlying system structure. We model interventions as ideal \textit{knockouts}, i.e. where for a given intervention, a variable is entirely removed from the system, and the intervention does not directly affect other variables of the system. 
For an intervention $c$, we define a binary mask $\bm{M}^{(c)} \in \{0,1\}^{d\times d}$
as: 
\begin{equation}
M_{ji}^{(c)} =
\begin{cases}
0 \text{ if } \space i=c , \space j \neq c\\
1 \text{ otherwise}
\end{cases}
\end{equation}
Where $i$ is the possible source of influence and $j$ is the target. The mask is structured to represent the severed outgoing influences of the variable $c$. In the observational setting, no mask is applied. Under condition $c$, the vector field's parameters become \smash{$\theta^{A(c)}_j$}, obtained by an element-wise product with $\bm{M}^{(c)}$: 
$\theta^{A(c)}_j=\bm{M}^{(c)}\odot \theta^A_j$. The drift is then evaluated as:
\begin{equation}
    \bm{v}^{\theta_j}_c(\bm{x}_t | c) = \psi(\cdots \psi(\psi(\bm{x}_t \theta_j^{A(c)}) \theta_j^{1}) \cdots)\theta_j^{K}.
\end{equation}

We can then recover our underlying matrix estimate \smash{$\bm{A} \in \mathbb{R}^{d \times d}$ as $A_{ji} = \|(\theta_j^A)_{i,:}\|_2$}.
Similarly, we condition the score \smash{$\bm{s}_t^{\phi}(\bm{x}_t)$} on $c$, i.e., \smash{$\nabla_{\bm{x}_t}\log{p_t(\bm{x}_t|\bm{k}^{(c)})}$} where \smash{$\bm{k}^{(c)}\in\{0,1\}^d$} is a conditional input vector with \smash{$k_{i}^{(c)} = 1 \text{ if } i \text{ is perturbed under } c, \text{ otherwise } k_{i}^{(c)} = 0$}.

\subsection{Simulation-free end-to-end Training}
\label{subsec:conditional_sim_free_learn}

\methodname builds on the \sfm framework introduced by \citet{tong2023simulation} to learn a neural approximation of $\bm{v}$ without simulation.
We use Entropic Optimal Transport (EOT) to pair points drawn from temporally adjacent snapshots \smash{$\hat{p}^{(c)}_{t_i}$} and \smash{$\hat{p}^{(c)}_{t_{i+1}}$}.
We use the Sinkhorn algorithm \citep{cuturi_sinkhorn_2013} to estimate the probabilistic EOT couplings between pairs of distributions \rev{only a single time before training starts, and do not need to be recomputed.}
From each coupling, we draw paired samples $(x_0,x_1)$ across $[t_i,t_{i+1}]$ and compute the corresponding target drift of the probability flow $\bm{v}^\circ_t$ and target score $\bm{s}_t = \nabla_{\bm{x}}\log p_t(\bm{x}_t\mid c)$.
We parametrize the drift of the probability flow as:
\begin{equation}    
\hat{\bm{v}}_{t,c}(\bm{x}_t;\Theta|\bm{M}^{(c)},\bm{k}^{(c)}) = \bm{v}^{\theta}_c(\bm{x}_t|\bm{M}^{(c)}) -\frac{\sigma^2}{2}\bm{s}^{\phi}_t\left(\bm{x}_t|\bm{k}^{(c)}\right).
\end{equation}
for conditional probability flow parametrization $\hat{\bm{v}}_{t,c}$ under condition $c$. We optimize model parameters $\Theta =(\theta, \phi)$ by regressing to the targets $\bm{v}^\circ_t$ and $\bm{s}_t$ via the \methodname loss
\begin{equation} 
\mathcal{L}_{\text{SF}}(\Theta)=\sum_c \mathbb{E}\left[\left(1-\alpha\right)||\hat{\bm{v}}_{t,c}(\bm{x}_t;\Theta |\bm{M}^{(c)},\bm{k}^{(c)}) t-\bm{v}^{\circ}_t||^2+\alpha||\bm{s}_{t,c}^{\phi}(\bm{x}_t|\bm{k}^{(c)})-\bm{s}_t||^2\right],
\end{equation}
where the expectation is over $t \sim \mathcal{U}(0,1)$ and $\bm{x}_t \sim p_t(\bm{x})$, and $0 \leq \alpha \leq 1$ weighs the vector field loss and score loss. 
During training, we apply group lasso on $\theta_j^A$ to encourage the model to learn sparse dependencies.
We include detailed pseudocode for training \methodname in Algorithm \ref{alg:sf2m}.

\rev{We note that we choose to solve the EOT problem to pair successive time marginals (\smash{$\hat{p}^{(c)}_{t_i}$} and \smash{$\hat{p}^{(c)}_{t_{i+1}}$}) independently of the other marginals \smash{$\hat{p}_t \mid t \notin \{i, i+1\}$}, not globally across all time marginals like is done in \citet{chen2023deepmomentummultimarginalschrodinger, theodoropoulos2025momentummultimarginalschrodingerbridge, rohbeck2025modeling}.} As noted by \citet{theodoropoulos2025momentummultimarginalschrodingerbridge}, coupling adjacent snapshots independently, as \methodname does, can limit the capture of long-range temporal dependencies and the coherence of inferred trajectories across the full sequence of timepoints. Global multi-marginal couplings \citep{chen2023deepmomentummultimarginalschrodinger, rohbeck2025modeling, theodoropoulos2025momentummultimarginalschrodingerbridge} instead condition on all timepoints jointly to learn smoother, globally consistent trajectories. We do not investigate the use of such couplings in \methodname, and leave this as potential direction for future work.

\section{Related Work}
\label{sec:related_work}

\vspace{-5pt}
\paragraph{Structure learning for continuous dynamics.} There exist several works for structure learning of continuous dynamics. Notably, \cite{bellot2021neural} introduce a continuous-time framework for inferring structure of the underlying dynamical system from time-series data. \cite{wang2023neural} extend this approach to the stochastic and Bayesian setting. However, these approaches rely on computationally expensive neural ODE solvers during training,
Recent lines of work have introduced structure learning methods for continuous dynamics specifically tailored for biological systems and the inference of gene regulatory networks (GRNs) \citep{huynh2010inferring, huynh2018dynGENIE3, gao2018sincerities, ishikawa2023renge}. However, all these methods assume a selection of fixed ODE parameterizations, linear dynamics, or perturbation-specific recovery, limiting their ability to model stochastic trajectories and complex network interactions. \rev{Symbolic regression-based approaches offer a complementary line of work \citep{steven2016discovering, sun2023symbolicphysicslearnerdiscovering, dakhmouche2025robust}.
\cite{dakhmouche2025robust} also solve for continuous population dynamics (dynamical/trajectory inference) and provide an interpretable model via the learned symbolic equations, but do not explicitly recover a network structure. 
\cite{steven2016discovering, sun2023symbolicphysicslearnerdiscovering} also use symbolic regression to recover closed-form equations, but assume fully observed trajectories.}

\paragraph{Structure learning from time-series data.}
A substantial body of work addresses structure learning when longitudinal
observations of individual particles are available. Classical approaches infer
dependencies via Granger causality or lag-based conditional independence testing,
with PCMCI and its variants~\citep{runge2019detecting} being a widely used
representative. \cite{etesami2014directed, etesami2016measuring}
formalize this perspective through directed information graphs for general dynamical
systems. More recent neural approaches include neural relational
inference~\citep{kipf2018neuralrelationalinferenceinteracting}, which jointly learns
interaction graphs and trajectories under deterministic dynamics, and
SCOTCH~\citep{wang2023neural}, which extends Bayesian structure learning to SDEs. A complementary line of work directly recovers governing equations from observed trajectories via sparse regression. SINDy~\citep{steven2016discovering} fits a sparse linear combination of candidate basis functions to time derivatives of state observations, with the recovered nonzero coefficients revealing which variables drive each other's dynamics, and stochastic extensions~\citep{boninsegna2018sparse} adapt this
framework to SDEs. These methods are powerful in their setting but rely on maintained particle identity across timepoints, an assumption violated in the destructive-measurement regime considered here. We also empirically compare against PCMCI adapted to snapshots via OT pseudo-trajectories in Appendix~\ref{ap:pcmci}, table~\ref{tab:pcmci_comparison}

\paragraph{Joint inference of structure and dynamics.} 
While both dynamical inference and structure learning are central to understanding stochastic systems, many existing methods treat these tasks in a decoupled fashion \citep{qiu2022mapping, sha2024reconstructing}. 
These procedures are centered around learning a dense vector field from the observed data, subsequently extracting structural information post-hoc. Moreover, they do not incorporate any prior on sparsity or explicitly model structure during training.
\cite{tong2023simulation} provide a preliminary investigation into the use of the NGM architecture in conjunction with \sfm for jointly inferring structure and dynamics, but do so in a limited setting, i.e. don't consider interventions and assume access to a significant quantity of time-points.
We demonstrate later (Section \ref{sec:experiments} Figure \ref{fig:AP_AUROC_interventional}) that the trivial application of the NGM architecture under the \sfm setup does not achieve competitive performance. 
\cite{lin2025interpretable} propose a neural ODE-based approach for the joint inference task, but do not model stochastic dynamics, only consider two time-points, and require simulation during training (akin to \cite{bellot2021neural}). \rev{Concurrent work also explores masking-based intervention modeling in single-cell
generative models \citep{jiang2025dspin, lorch2026latent}, highlighting joint
structure-and-dynamics learning under interventions as an actively studied problem.}
\methodname addresses these limitations for the joint task.

\vspace{-3pt}
\section{Experiments}
\label{sec:experiments}

\begin{wrapfigure}{R}{0.5\textwidth}
    \vspace{-3em}
    \centering
    \includegraphics[width=\linewidth]{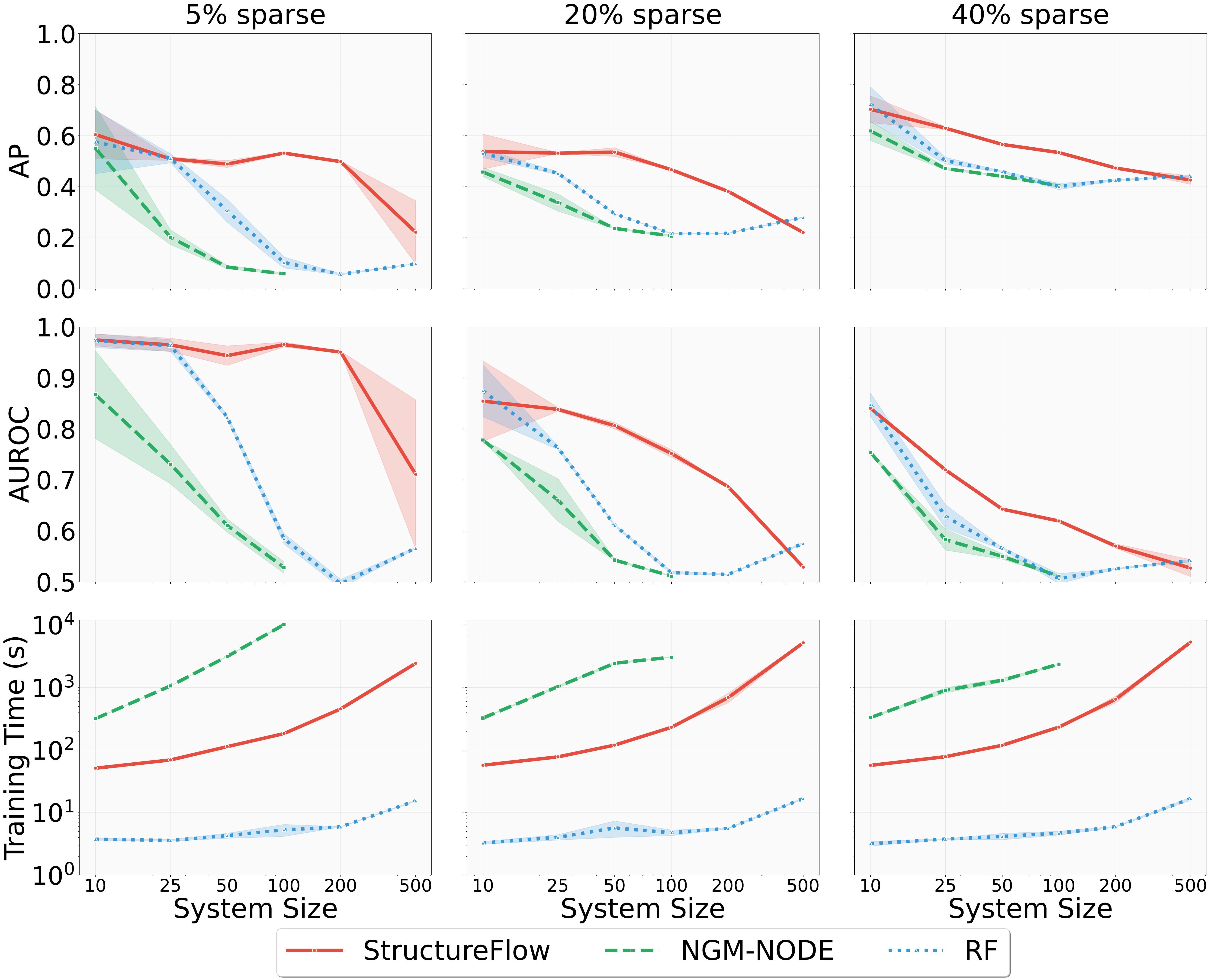}
    \caption{\textbf{\methodname yields a favorable structure-learning accuracy--runtime tradeoff when scaled to high-dimensional systems.} We compare with NGM-NODE and RF on synthetic linear systems with varying dimensionality ($d$) and system (graph) sparsity levels (5\%, 20\%, 40\%).}
    \vspace{-3em}
    \label{fig:linear_scaling}
\end{wrapfigure}

\vspace{-5pt}
We evaluate \methodname on both synthetic and real-world datasets to assess its ability to recover underlying network structure and infer stochastic population dynamics. Our experiments focus on three key capabilities: structure learning of dynamical systems from population data, dynamical inference across left-out time-points, and prediction of responses to unseen interventions (knockouts). We also include a scaling study to highlight our method's computational efficiency. We provide further details for these experiments in Appendix~\ref{ap:expt_details}, \ref{ap:implementation_details}, and \ref{ap:additional_results}.

\subsection{Experiments on Synthetic High-dimensional Systems}

We evaluate \methodname for the structure learning task across varying system dimensionality using randomly generated Erd\H{o}s--R\'{e}nyi graphs \citep{erdos59a} and derive synthetic data via linear SDE simulation (see Appendix \ref{ap:synthetic_linear_system} for details). We vary the dimensionality of the system $d$ from 10 to 500 variables and the sparsity (proportion of non-zero edges) from 5\% to 40\% of the underlying graph which defines the data-generative process of the dynamical system. We consider a simulation-based approach, NGM-NeuralODE (NGM-NODE) \citep{bellot2021neural}, and Reference Fitting (RF) \citep{zhang2024joint}, as baselines, since they are capable of the joint structure learning and trajectory inference task. 
We report results in Figure ~\ref{fig:linear_scaling} showing area under the receiver operating characteristic curve (AUROC), average precision (AP), and training time (seconds). We observe that \methodname exhibits favorable scaling performance as dimensionality of the system increases. While the baseline methods are competitive on small systems, we see that \methodname consistently maintains strong performance in high-dimensional settings while exhibiting improved computational efficiency in terms of training time compared to NGM-NODE, especially for large $d$. Thus, our scaling claim is about the accuracy vs. runtime tradeoff against alternative simulation-based approaches, rather than uniformly cheaper than every baseline. We remark that RF is the most computationally efficient in this regard as it makes assumptions of linearity of the underlying system. Due to this, RF suffers from poor expressivity and exhibits poor performance on the second element of the joint task -- dynamical (trajectory) inference. We show this in the following sections. 

\subsection{Experiments on Simulated Biological Systems}

We consider six simulated biological systems: trifurcating (\textbf{TF}), cyclical (\textbf{CY}), long linear (\textbf{LL}), swirling (\textbf{SW}), bifurcating (\textbf{BF}), and a curated system that mimics hematopoietic stem cell (\textbf{HSC}) differentiation. All datasets are simulated via BoolODE \citep{pratapa2020benchmarking} under both observational and interventional (knockout) conditions across multiple time-points (see details in Appendix~\ref{ap:expt_details}). See Appendix~\ref{ap:additional_results} for extended synthetic system results and ablations. In this section, we empirically evaluate \methodname for the joint structure learning and dynamics inference tasks.  

\begin{figure}[!t]
    \centering
    \includegraphics[width=\textwidth, height=\textheight, keepaspectratio]{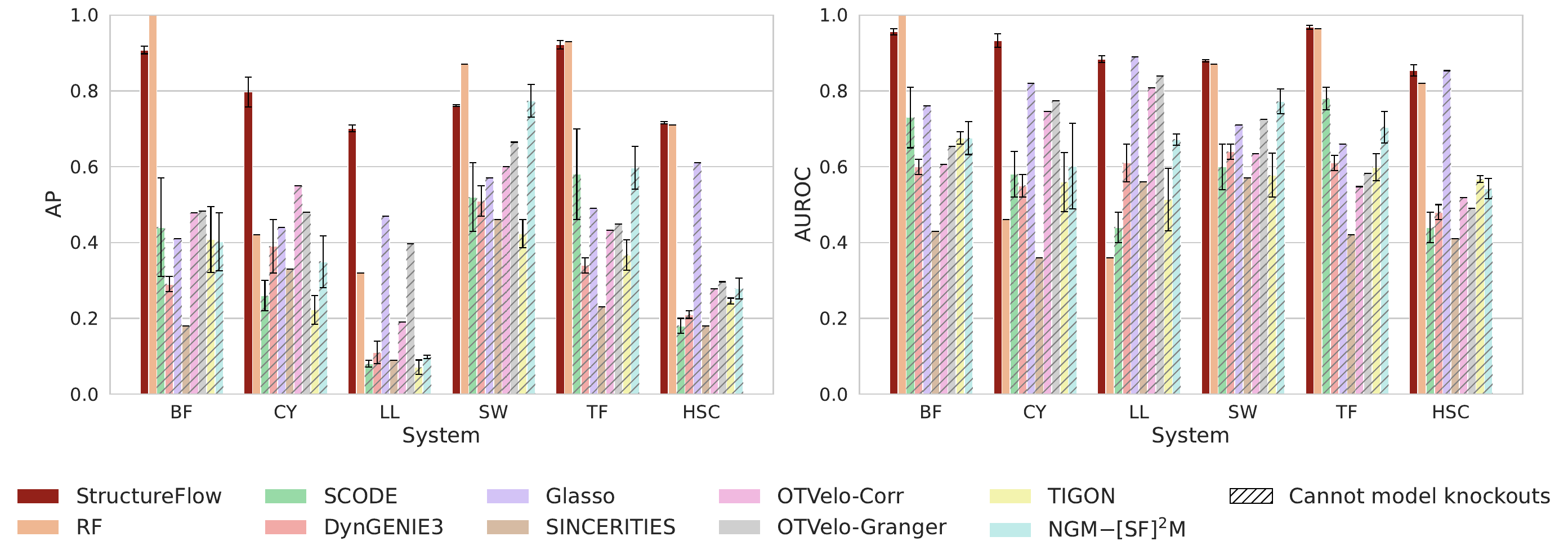}
    \caption{\textbf{\methodname is consistently a top performing \textit{structure learning} method across simulated biological systems.} Here, we use interventional (with \textit{knockouts}) and observational (no \textit{knockouts}) data, and report average precision (AP) and area under the ROC curve (AUROC) scores.}
    \label{fig:AP_AUROC_interventional}
\end{figure}

\begin{figure}[!th]
  \centering
  \begin{subfigure}{\textwidth}
  \centering
    \begin{minipage}[c]{0.05\textwidth}
    \end{minipage}%
    \begin{minipage}[c]{0.95\textwidth}
      \includegraphics[width=\textwidth]{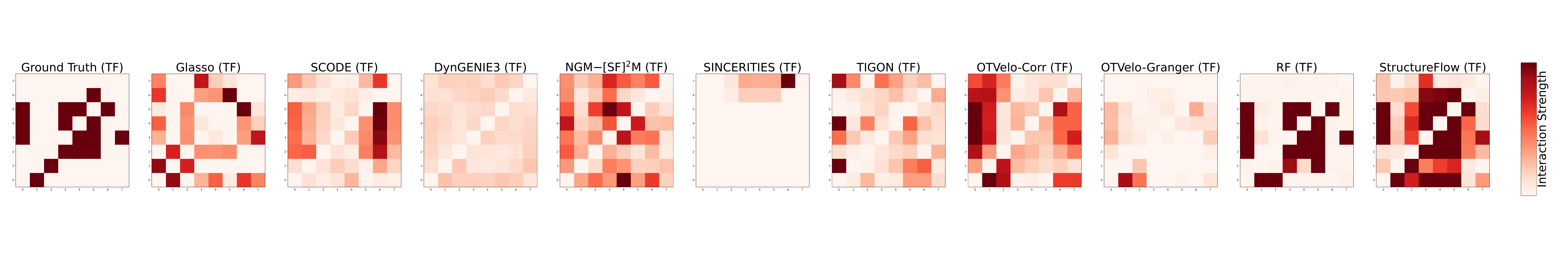}
    \end{minipage}
    \label{fig:tf_heatmaps}
  \end{subfigure}%

  \vspace{-0.5em}
  \centering
  \begin{subfigure}[b]{0.49\textwidth}
    \centering
    \includegraphics[width=\textwidth]{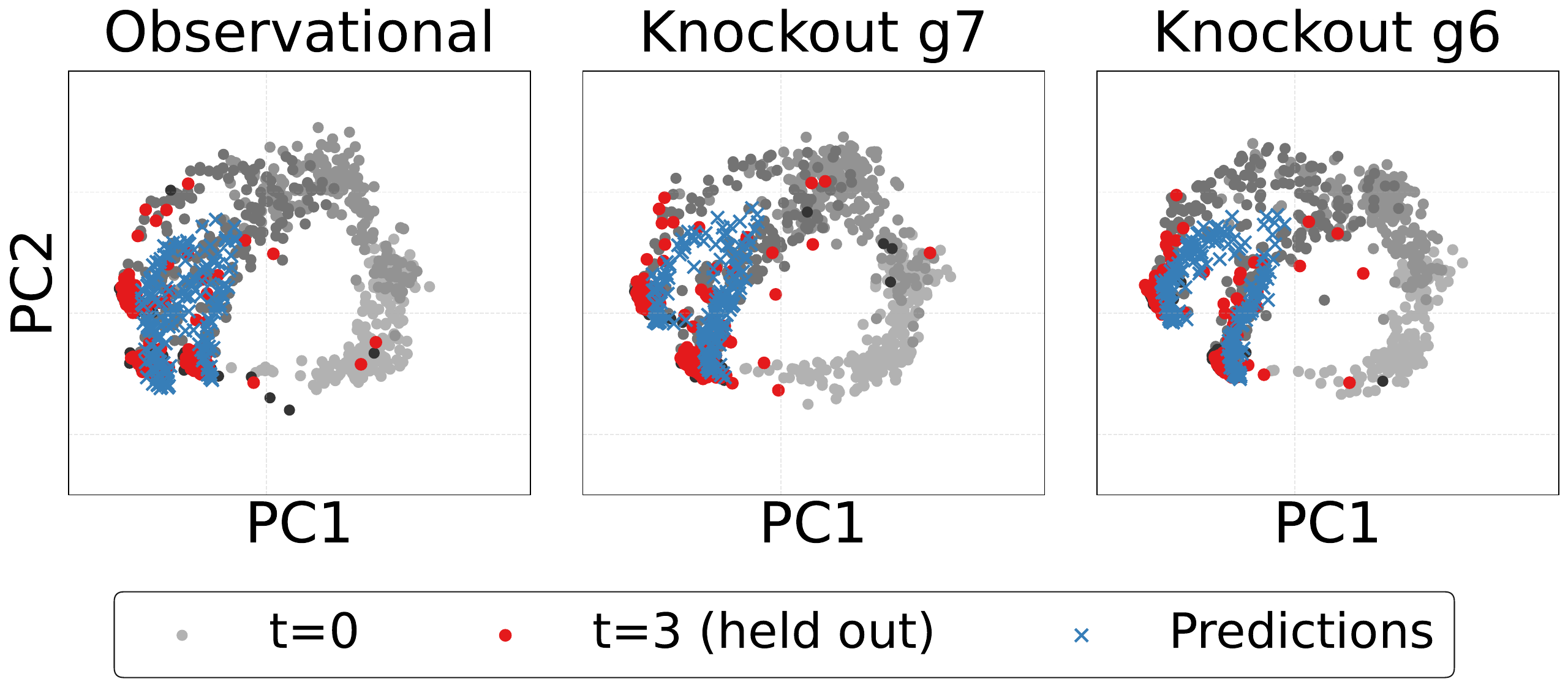}
    \caption{\textbf{\methodname (ours)}}
    \label{fig:rf_trajectory}
  \end{subfigure}
  \hfill
  \begin{subfigure}[b]{0.49\textwidth}
    \centering
    \includegraphics[width=\textwidth]{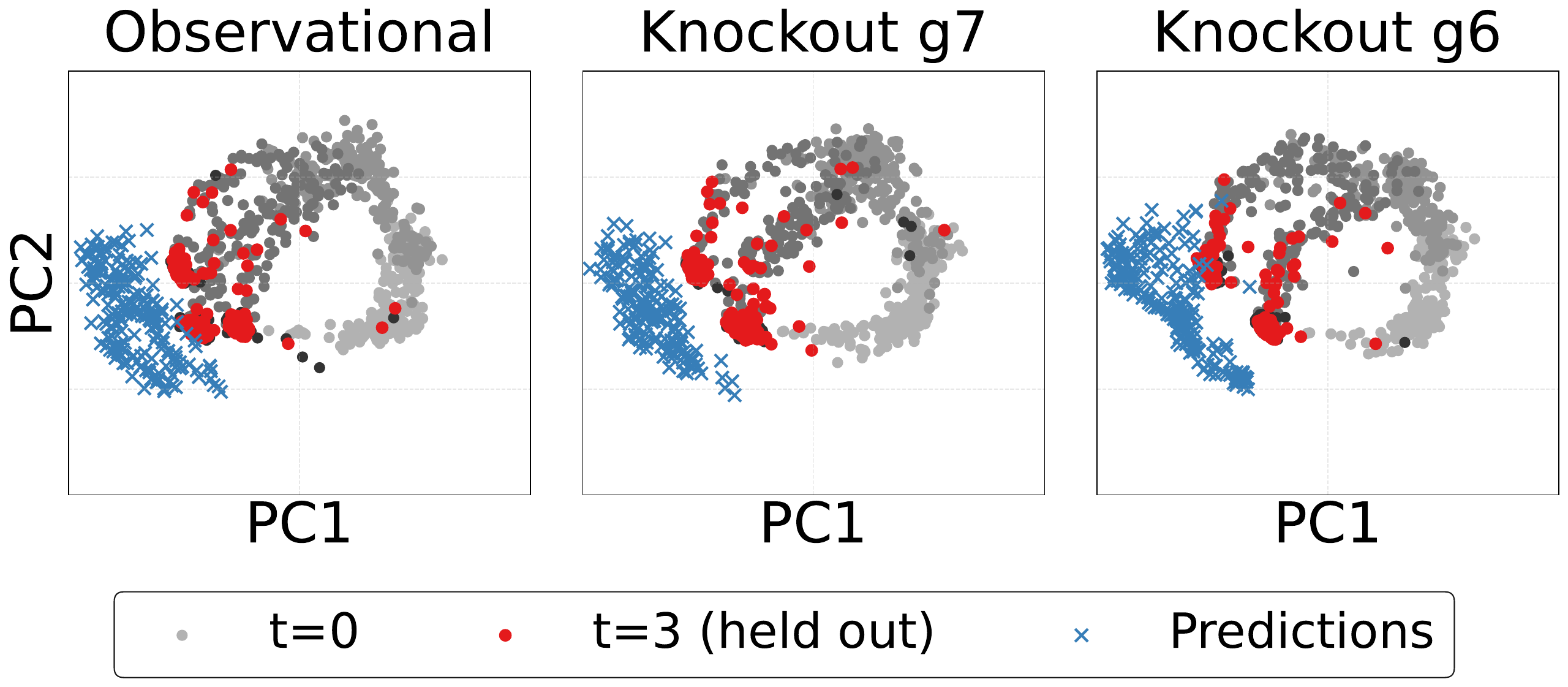}
    \caption{\textbf{RF}}
    \label{fig:structure_flow_trajectory}
  \end{subfigure}
    
  \vspace{0.3em}
  \begin{subfigure}[b]{0.49\textwidth}
    \centering
    \includegraphics[width=\textwidth]{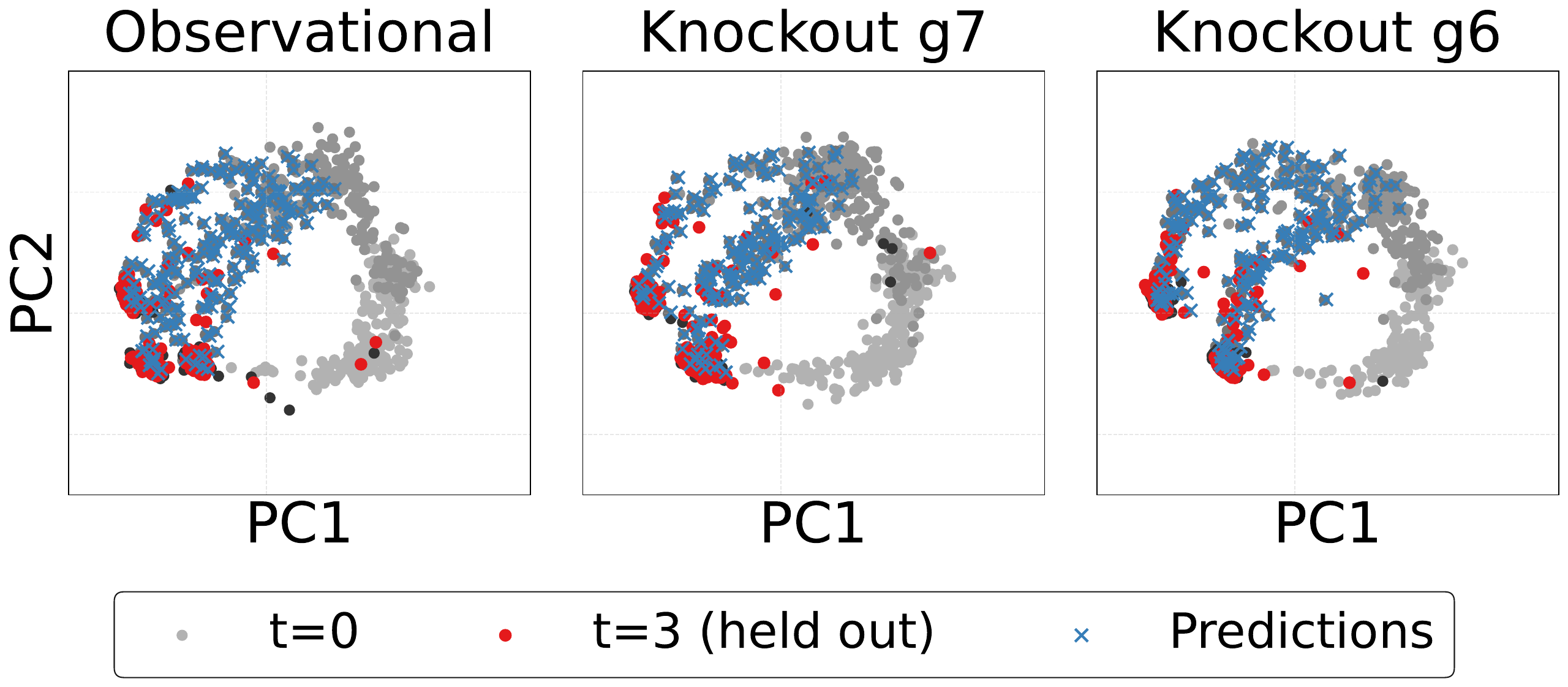}
    \caption{\textbf{TIGON}}
    \label{fig:tigon_trajectory}
  \end{subfigure}
  \hfill
  \begin{subfigure}[b]{0.49\textwidth}
    \centering
    \includegraphics[width=\textwidth]{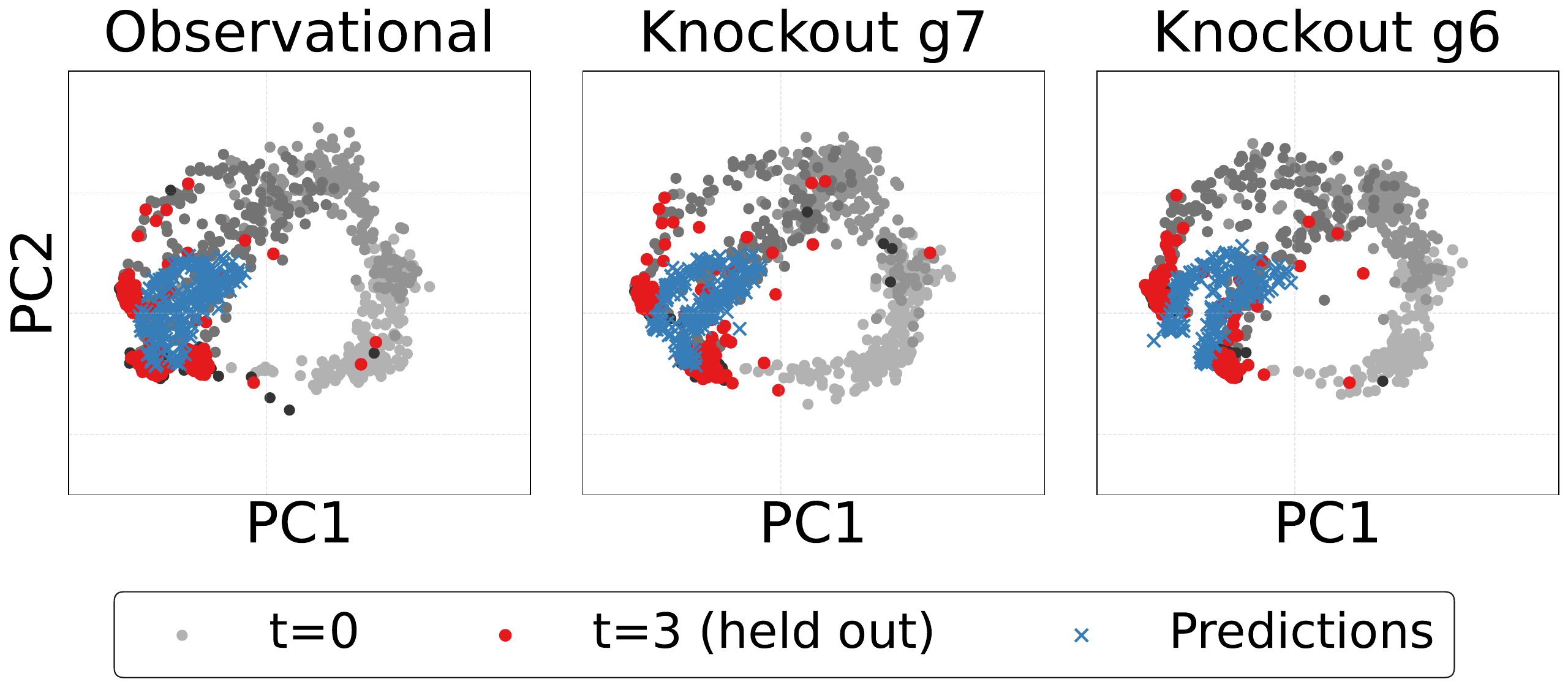}
    \caption{\textbf{OTVelo}}
    \label{fig:otvelo_trajectory}
  \end{subfigure}

    \vspace{-0.5em}
    \caption{\textbf{Summary of inferred structure and visualization of model predicted dynamics.} (top) connectivity matrices (heatmaps) showing ground truth and inferred graphs for the \textbf{TF} (trifurcating) system. The x/y-axis labels correspond to system variables and shading indicates edge interaction strength. (bottom) visual comparison of dynamical inference using four different methods: \methodname, RF, TIGON, and OTVelo, respectively. We show 2D plots of left-out timepoint prediction for the observational (wild-type) setting and for two \textit{seen} interventions (knockouts) selected for their diverse trifurcating paths. This figure is intended as a qualitative visualization of structure and representative trajectory behavior; the performance comparison is quantified in Figure~\ref{fig:AP_AUROC_interventional} and Table~\ref{tab:timepoint_results}.}

    \vspace{-1em}
  \label{fig:combined}
\end{figure}

\begin{table}[!t]
\centering
\caption{\textbf{\methodname achieves competitive performance for \textit{dynamical inference} on simulated biological systems.} 
Shown is a comparison of dynamical inference methods for learning conditional population dynamics across synthetic biological systems. 
We include \sfm for comparison, but note that \sfm does not infer underlying network structure (i.e. cannot address joint task).
Colors indicate \best{1$^{\text{st}}$ best performer} and \second{2$^{\text{nd}}$ best performer} for models that perform the joint inference task.
}
\vspace{-0.7em}
\label{tab:timepoint_results}
\resizebox{\columnwidth}{!}{
\begin{tabular}{@{}lcccccccccccc@{}}
\toprule
 & \multicolumn{2}{c}{\textbf{TF}} & \multicolumn{2}{c}{\textbf{CY}} & \multicolumn{2}{c}{\textbf{LL}} & \multicolumn{2}{c}{\textbf{SW}} & \multicolumn{2}{c}{\textbf{BF}} & \multicolumn{2}{c}{\textbf{HSC (Curated)}} \\
\cmidrule(lr){2-3}
\cmidrule(lr){4-5}
\cmidrule(lr){6-7}
\cmidrule(lr){8-9}
\cmidrule(lr){10-11}
\cmidrule(lr){12-13}
 & $W_2$↓ & MMD↓ & $W_2$↓ & MMD↓ & $W_2$↓ & MMD↓ & $W_2$↓ & MMD↓ & $W_2$↓ & MMD↓ & $W_2$↓ & MMD↓ \\
 
\midrule
\sfm & 0.761 $\pm$ 0.014 & 0.134 $\pm$ 0.005 & 0.517 $\pm$ 0.015 & 0.153 $\pm$ 0.002 & 0.847 $\pm$ 0.053 & 0.190 $\pm$ 0.010 & 0.572 $\pm$ 0.008 & 0.215 $\pm$ 0.005 & 0.660 $\pm$ 0.007 & 0.140 $\pm$ 0.004 & 0.664 $\pm$ 0.011 & 0.083 $\pm$ 0.003 \\

\midrule

RF & \second{1.376 $\pm$ 0.000} & 0.191 $\pm$ 0.000 & 2.086 $\pm$ 0.000 & 0.326 $\pm$ 0.000 & 2.120 $\pm$ 0.007 & 0.296 $\pm$ 0.000 & 1.371 $\pm$ 0.000 & 0.290 $\pm$ 0.000 & 1.495 $\pm$ 0.000 & 0.242 $\pm$ 0.000 & 0.950 $\pm$ 0.000 & 0.066 $\pm$ 0.000 \\

OTVelo & 1.874 $\pm$ 0.262 & 0.582 $\pm$ 0.039 & 1.972 $\pm$ 0.144 & 0.622 $\pm$ 0.034 & 2.878 $\pm$ 0.361 & 0.647 $\pm$ 0.015 & 1.940 $\pm$ 0.144 & 0.577 $\pm$ 0.022 & 1.847 $\pm$ 0.277 & 0.591 $\pm$ 0.042 & 1.942 $\pm$ 0.006 & 0.603 $\pm$ 0.002 \\

TIGON & 1.409 $\pm$ 0.294 & \best{0.063 $\pm$ 0.022} & \second{1.224 $\pm$ 0.053} & \best{0.044 $\pm$ 0.003}  & \second{1.840 $\pm$ 0.293} & \best{0.066 $\pm$ 0.039} & \second{0.957 $\pm$ 0.190} & \best{0.017 $\pm$ 0.009} & \second{1.394 $\pm$ 0.294} & \best{0.066 $\pm$ 0.021} & \best{0.666 $\pm$ 0.363} & \best{0.012 $\pm$ 0.014} \\

\methodname & \best{0.789 $\pm$ 0.012} & \second{0.135 $\pm$ 0.004} & \best{0.576 $\pm$ 0.012} & \second{0.140 $\pm$ 0.004} & \best{0.842 $\pm$ 0.031} & \second{0.196 $\pm$ 0.008} & \best{0.610 $\pm$ 0.010} & \second{0.220 $\pm$ 0.006} & \best{0.694 $\pm$ 0.017} & \second{0.142 $\pm$ 0.007} & {\second{0.683 $\pm$ 0.011}} & \second{0.065 $\pm$ 0.003} \\

\bottomrule
\end{tabular}
}
\end{table}

\paragraph{\methodname effectively infers network structure of simulated biological systems.}
To evaluate how well \methodname recovers the underlying network structure, we extract the inferred graph from the first layer of the parameterized vector field (defined in \eqref{eq:dyn_ngm}) which interprets the learned weight matrix as a proxy for variable-variable (gene-gene) interactions. 
We use AP and AUROC to evaluate how closely the \methodname inferred graphs recapitulate the ground truth structure used to simulate the respective synthetic biological systems.\footnote{In this work, we do not threshold the learned graphs/structures. Instead, we use AP/AUROC metrics to evaluate structure learning performance across all possible thresholds. We leave extensions for learning thresholded structures for future work.}
We compare \methodname to a variety of baseline methods (some of which are tailored for structure learning in biological systems): RF \citep{zhang2024joint}, OTVelo \citep{Zhao2024OTVelo}, TIGON \citep{sha2024reconstructing}, dynGENIE3 \citep{huynh2018dynGENIE3}, SINCERITIES \citep{gao2018sincerities}, SCODE \citep{matsumoto2017scode}, graphical lasso (Glasso), and NGM-$[$SF$]^2$M \citep{tong2023simulation}. 
All models are trained across five random seeds. 

We show the quantitative results for this experiment and evaluation in Figure~\ref{fig:AP_AUROC_interventional} and visualizations of the inferred structures in Figure~\ref{fig:combined} (top).
We observe that \methodname exhibits higher consistency as a top performing method for inferring the underlying structure across all systems and both metrics. In comparison, certain baseline methods at best show favorable performance in specific systems and on specific metrics, but struggle to consistently perform across the board. 
 {For instance, on structure learning (Figure~\ref{fig:AP_AUROC_interventional}) RF matches \methodname on BF, SW, TF, and HSC but drops on the cyclical (CY) and long-linear (LL) systems (AP $\approx 0.32$--$0.42$ vs.\ $\approx 0.70$--$0.80$), reflecting RF's single global linear OU matrix versus \methodname's nonlinear NGM drift.}

 {This structure-level competitiveness does not carry over to the concurrent task: on dynamical inference (Table~\ref{tab:timepoint_results}) RF is consistently worse than \methodname on every system ($W_2 \approx 0.95$--$2.1$ vs.\ $\approx 0.58$--$0.84$), as a linear drift cannot capture the nonlinear stochastic population dynamics. RF is computationally cheap and effective when a global linear drift suffices, but \methodname remains a generalizable structure-learning method that also models the nonlinear, stochastic population dynamics that RF cannot.}

\vspace{-5pt}
\paragraph{\methodname effectively infers dynamics across left-out time-points in simulated biological systems.}
To evaluate dynamical inference performance, we consider a leave-one-out cross-validation experiment across discrete timepoints, where we iteratively exclude each timepoint $t_k$ for $k \in \{1, \ldots, T-2\}$ during model training, then simulate via ODE from the ground truth data at $t_{k-1}$ to $t_k$. 
We use the Wasserstein-2 distance ($W_2$) and  {maximum mean discrepancy (MMD)} to evaluate the distributional distances between the predicted and ground truth distributions at $t_k$.

We consider RF, OTVelo, and TIGON as baselines since they are designed to address both inference tasks. 
We additionally include a comparison to $\mathrm{[SF]^2M}$ as a baseline method for dynamical inference, but note that $\mathrm{[SF]^2M}$ is not capable of jointly inferring the network structure. We show quantitative results for this experiment in Table~\ref{tab:timepoint_results} and visualization of the predictions in Figure~\ref{fig:rf_trajectory} and Figure~\ref{fig:structure_flow_trajectory}.  {\methodname achieves the best $W_2$ among methods that perform the joint task across the six systems, while TIGON achieves the lowest MMD (with \methodname second on MMD).}

\begin{table}[t]
    \centering
    \caption{\textbf{Unbalanced \methodname}}
    \label{tab:unbalanced}
    \resizebox{.85\columnwidth}{!}{%
        \begin{tabular}{@{}lcccc@{}}
        \toprule
         & AUROC ↑ & AP ↑ & $W_2$ ↓ & MMD ↓ \\
        \midrule
        TIGON & 0.540 $\pm$ 0.014 & 0.427 $\pm$ 0.047 & 2.355 $\pm$ 0.591  & 0.038 $\pm$ 0.041 \\
        \methodname & 0.735 $\pm$ 0.005 & 0.358 $\pm$ 0.006 & 1.672 $\pm$ 0.532 & 0.029 $\pm$ 0.031 \\
        \textbf{U}-\methodname & \textbf{0.838 $\pm$ 0.010} & \textbf{0.509 $\pm$ 0.025} & \textbf{1.670 $\pm$ 0.438} & \textbf{0.027 $\pm$ 0.033} \\
        \bottomrule
        \end{tabular}
    }
    \vspace{-1em}
\end{table}
\vspace{-7pt}

{\paragraph{Ablations of core design choices.} Two choices central to \methodname are the group lasso regularization on the structural (FC1) layer and the weight $\alpha$ balancing the stochastic score and deterministic flow components. On the TF system, group lasso improves structure recovery up to an intermediate strength (AP $0.80\!\to\!0.92$, AUROC $0.94\!\to\!0.97$ at weight $0.04$) before over-sparsification degrades it, and the score/flow balance is best at $\alpha \in [0.1, 0.5]$ with a drop at higher score weight ($\alpha=0.8$). Full sweeps (including model capacity) are reported in Appendix~\ref{ap:Structureflow_ablations}, Tables~\ref{tab:group_lasso}, \ref{tab:alpha_ablation}, and \ref{tab:hidden_dim_ablation}.}

\paragraph{\methodname infers structure and dynamics in the unbalanced setting.}
We show a use case of \methodname in the unbalanced setting. To extend \methodname to the unbalanced setting (\textbf{U}-\methodname), we replace the balanced Sinkhorn algorithm in our EOT procedure with the unbalanced Sinkhorn divergence and incorporate a growth rate field to model proliferation-induced density shifts. We evaluate our method on a growth-based bifurcation system. Following the setup of \cite{zhang2025inferringstochasticdynamicsgrowth}, we use the underlying gene regulatory network and boolean logic of the standard BoolODE bifurcation dataset, but we impose lineage specific proliferation rates where one of the attractors has a high proliferation rate and the other has a near-zero growth rate. We observe unbalanced-\methodname yields the best performance for the joint task (Table \ref{tab:unbalanced}). We discuss further details in Appendix~\ref{ap:unbalancedSF}.

\subsection{Experiments on Real Data (Renge) System}
We consider an interventional time-series dataset of human induced pluripotent stem cells (iPSC) introduced by \citep{ishikawa2023renge}. Interventions are conducted via clustered regularly interspaced short palindromic repeats (CRISPR) knockouts on 23 transcription factors (TFs). Cell states are then sequenced via single-cell RNA-seq across 4 time bins for each CRISPR intervention. 
We process the dataset following the procedure of \cite{zhang2024joint} and select the 8 interventions with greatest observed change in population-level gene expression; see Appendix~\ref{ap:real_data_system}.

\begin{wrapfigure}[14]{r}{0.48\textwidth}
    \vspace{-2em}
    \centering
    \includegraphics[width=0.47\textwidth]{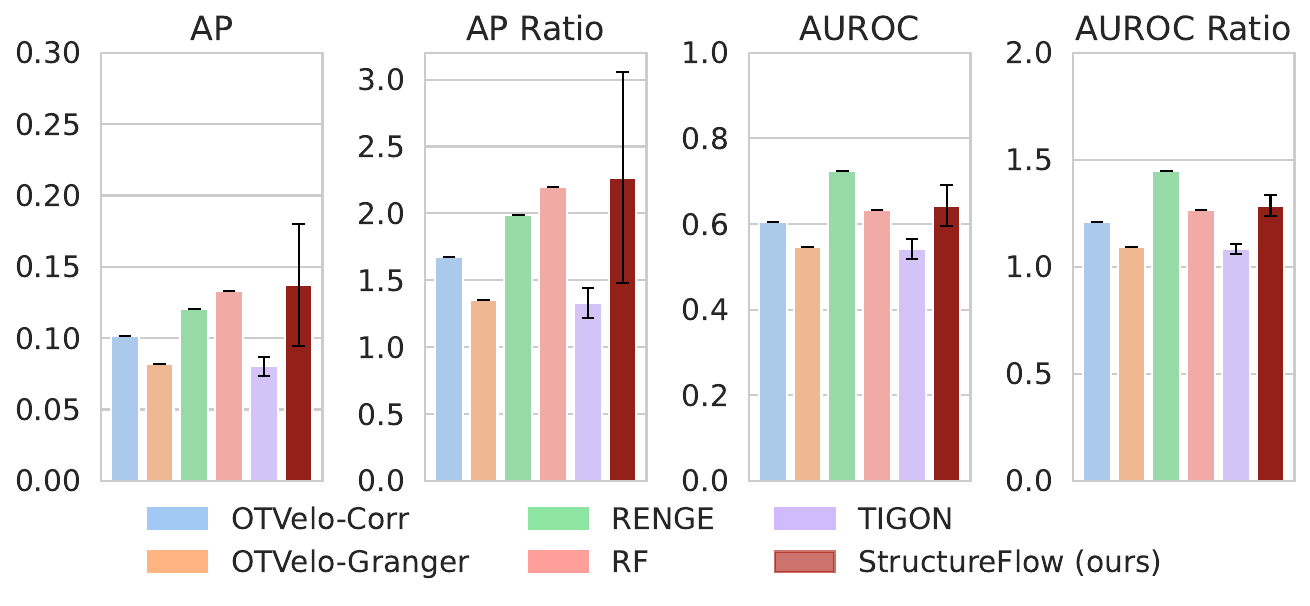}
    \caption{\textbf{\methodname achieves competitive benchmark agreement for the \textit{structure learning} tasks in the real-data (Renge) system.} We use the AP/AUROC ratio to denote structure recovery performance w.r.t. a random predictor.} 
    \label{fig:real_data_structure}
\end{wrapfigure}

\paragraph{\methodname recovers underlying structure of gene regulatory networks competitively to baselines.}

We evaluate \methodname on its ability to recover the $103 \times 103$ directed network of inferred TF-TF interactions from the processed real data.
We remark that a central challenge in evaluating structure learning performance on real biological systems is that there rarely exists a definitively known ground truth network. In this setting, we have access to an approximation of the ground-truth network for 18 genes, determined using a ChIP-seq reference; see Appendix~\ref{ap:real_data_experiments} for details. Accordingly, AP and AUROC measure agreement with this partial reference and should not be interpreted as a complete validation of the full regulatory graph.
Models are trained using all $103$ genes, but evaluation is done over $18 \times 103$ \emph{ground-truth} networks.
We show results in Figure~\ref{fig:real_data_structure} and observe that \methodname yields competitive performance to counterpart baselines.
Because performance varies across random seeds, we avoid interpreting the best seed as a definitive ranking; instead, these results indicate that \methodname is competitive with the strongest baselines under the available partial-reference benchmark.
We include additional results in Appendix~\ref{ap:additional_results}.

\paragraph{\methodname effectively models cell trajectories and improves prediction of left-out interventions (knockouts) in real-data setting.} 
We report results for dynamical inference over left out time-points and left-out knockouts on the real biological system in Table \ref{tab:renge_timepoint_results} and Table \ref{tab:renge_knockout_results}, respectively. 
In the left-out intervention (knockout) setting, we withhold all time marginals $t_k, k=[0, ..., T-1]$ for a given interventional condition during training, then evaluate performance on predicting the final timepoint.
Similar to the simulated biological systems, we use $W_2$ and ED to evaluate performance. 
OTVelo and TIGON do not model interventions, and thus are not included in the left-out intervention evaluation.
For left-out-time point dynamical inference (Table \ref{tab:renge_timepoint_results}), we observe that \methodname achieves the best $W_2$ among evaluated methods and yields competitive results on ED. 
\methodname is the best performing joint method on the left-out intervention task, and although it shows improved generalization in this setting, implying some plausibly improved learning of underlying mechanisms, we make no claim that \methodname truly models the \textit{causal} dependencies of the underlying data generative process. The inferred networks and intervention responses should therefore be treated as model-based hypotheses requiring biological validation.
We include an equivalent experiment for the simulated biological systems in Appendix \ref{ap:additional_results} Table \ref{tab:unseen_ko_boolode}.

\vspace{-0.5em}
\begin{table}[H]
\centering
\caption{\textbf{\methodname shows competitive performance for \textit{dynamical inference} of left-out marginals on the real (Renge) biological dataset.}
We show a comparison of dynamical inference methods for learning conditional population dynamics across left-out timepoints $k=1,2$ on the Renge dataset.
We report results over withheld $k$ as well as the average of the two using Wasserstein-2 ($W_2$↓) and Energy Distance (ED↓) \citep{rizzo2016energy} metrics.
Bold indicates \best{1$^{\text{st}}$ best performer} and the underline indicates \second{2$^{\text{nd}}$ best performer} for models that perform the joint inference task.
}

\vspace{0.3em}
\label{tab:renge_timepoint_results}
\resizebox{\columnwidth}{!}{
\begin{tabular}{@{}l cccccc@{}}
\toprule
 & \multicolumn{2}{c}{Timepoint 1} & \multicolumn{2}{c}{Timepoint 2} & \multicolumn{2}{c}{Average} \\
\cmidrule(lr){2-3} \cmidrule(lr){4-5} \cmidrule(lr){6-7}
 & $W_2$↓ & ED↓ & $W_2$↓ & ED↓ & $W_2$↓ & ED↓ \\
\midrule
\sfm & 5.673 $\pm$ 0.034 & 0.533 $\pm$ 0.024 & 5.841 $\pm$ 0.030 & 0.524 $\pm$ 0.024 & 5.757 $\pm$ 0.032 & 0.528 $\pm$ 0.024 \\
\midrule
RF & 10.193 $\pm$ 0.004 & 12.389 $\pm$ 0.009 & 9.242 $\pm$ 0.004 & 10.089 $\pm$ 0.008 & 9.717 $\pm$ 0.004 & 11.239 $\pm$ 0.008 \\
OTVelo & 7.106 $\pm$ 0.000 & 4.269 $\pm$ 0.000 & 6.996 $\pm$ 0.000 & 3.658 $\pm$ 0.000 & 7.051 $\pm$ 0.000 & 3.963 $\pm$ 0.000 \\
TIGON & \second{6.401 $\pm$ 0.033} & \best{0.173 $\pm$ 0.011} & \second{6.402 $\pm$ 0.002} & \best{0.125 $\pm$ 0.001} & \second{6.402 $\pm$ 0.001} & \best{0.149 $\pm$ 0.024} \\
\methodname & \best{5.589 $\pm$ 0.026} & \second{0.692 $\pm$ 0.048} & \best{5.679 $\pm$ 0.029} & \second{0.765 $\pm$ 0.025} & \best{5.634 $\pm$ 0.027} & \second{0.729 $\pm$ 0.037} \\
\bottomrule
\end{tabular}
}
\end{table}

\begin{table}[H]
\centering
\caption{\textbf{\methodname shows competitive performance for \textit{dynamical inference} of left-out interventions on the real (Renge) biological dataset.}
We show a comparison of dynamical inference methods for learning conditional population dynamics on the Renge dataset for left-out interventions.
We report results for 3 left-out gene knockout conditions using Wasserstein-2 ($W_2$↓) and Energy Distance (ED↓) metrics, as well as the average performance across conditions. \textbf{Bold} indicates the best performing method that can perform the joint task.
We provide a visualization of this result in Appendix~\ref{ap:additional_results} \cref{fig:leftout_kos_comparison}.
}
\vspace{-0.5em}
\label{tab:renge_knockout_results}
\resizebox{\columnwidth}{!}{
\begin{tabular}{@{}l cccccccc@{}}
\toprule
 & \multicolumn{2}{c}{NANOG} & \multicolumn{2}{c}{POU5F1} & \multicolumn{2}{c}{PRDM14} & \multicolumn{2}{c}{Average} \\
\cmidrule(lr){2-3} \cmidrule(lr){4-5} \cmidrule(lr){6-7} \cmidrule(lr){8-9}
 & $W_2$↓ & ED↓ & $W_2$↓ & ED↓ & $W_2$↓ & ED↓ & $W_2$↓ & ED↓ \\
\midrule
\sfm & 5.726 $\pm$ 0.015 & 0.467 $\pm$ 0.014 & 5.931 $\pm$ 0.024 & 0.826 $\pm$ 0.020 & 6.002 $\pm$ 0.043 & 0.639 $\pm$ 0.030 & 5.886 $\pm$ 0.027 & 0.644 $\pm$ 0.021 \\
\midrule
RF & 9.859 $\pm$ 0.002 & 11.492 $\pm$ 0.006 & 10.211 $\pm$ 0.002 & 12.140 $\pm$ 0.004 & 9.889 $\pm$ 0.005 & 11.461 $\pm$ 0.013 & 9.986 $\pm$ 0.003 & 11.698 $\pm$ 0.008 \\
\methodname & \textbf{5.517 $\pm$ 0.041} & \textbf{0.818 $\pm$ 0.015} & \textbf{5.717 $\pm$ 0.021} & \textbf{1.136 $\pm$ 0.039} & \textbf{5.816 $\pm$ 0.023} & \textbf{0.973 $\pm$ 0.053} & \textbf{5.683 $\pm$ 0.028} & \textbf{0.976 $\pm$ 0.036} \\
\bottomrule
\end{tabular}
}
\end{table}

\vspace{-1em}
\section{Conclusion}
\label{sec:conclusion}

\vspace{-0.5em}
In this work, we introduced \methodname, a novel simulation-free training method tailored for jointly learning the underlying network structure and the conditional population dynamics of high-dimensional stochastic systems. 
Our work highlights the benefits of addressing the joint inference task with a single model trained via score-and-flow matching and tailored for the problem --- i.e. improved performance on both tasks. 
We demonstrate this through a suite of empirical experiments and show that \methodname exhibits improved performance on several structure-learning and $W_2$ dynamical-inference metrics compared to existing approaches, which either tackle the individual tasks in isolation and/or are limited in their ability to address both problems. %
Moreover, we showcase the application of our method on a challenging real system for recovering network structure and simultaneously predicting the system response of left-out (\emph{unseen}) interventions. 

\vspace{-0.2em}
\paragraph{Limitations \& future work.} 
In this work, we focused on the problem of joint structure learning and dynamical inference for stochastic population dynamics. 
With this come non-standard challenges, e.g. evaluating structure learning performance requires knowledge of the ground truth directed graphs which define the data generative process of the dynamical system. 
To address this, we considered a suite of synthetic and biologically meaningful simulated systems \citep{pratapa2020benchmarking}, such that both tasks can be evaluated in unison, which emulate real systems. 
However, these simulated systems may be limited in their ability to fully reflect system behavior in real settings. \rev{For example, our simulated knockout conditions may not exactly reflect realistic CRISPR effects, such as imperfect knockouts. We believe exploring settings with imperfect intervention is a fruitful direction for future work.} Sensitivity analyses over the assumed diffusion scale $\sigma$, the number of timepoints, and the number of samples per timepoint are similarly valuable directions. More generally, inferred networks and predicted perturbation responses should be used as hypotheses for follow-up experiments rather than as clinical or therapeutic evidence without independent validation.

In a similar vein, evaluating structure learning performance in real-data settings requires experimental validation (which was accomplished by \citet{ishikawa2023renge} to produce what we label as the Renge dataset). Such experimental validation is costly to acquire and unrealistic in large systems. 
Thus, this experimental validation is limited to a small set of system variables, and is still at best an estimate of the ground truth network. Stronger validation remains an open challenge: this would require larger or orthogonal evidence sources, such as additional chromatin accessibility data or curated regulatory databases. This would require substantial new data generation or curation and is outside the scope of the present work.
Another limitation is that \methodname outputs unsigned edge scores. If signed interactions are needed, a sign can be estimated from the corresponding entry of the Jacobian of the learned drift, e.g. as done in \citet{qiu2022mapping}. Exploring other approaches to directly infer signed edges is a potential future direction.

We also do not consider certain aspects of physical processes that are prevalent in natural systems, such as modeling population dynamics while considering interacting particles \citep{atanackovic2024meta}
and/or \rev{structure learning of non-stationary networks \citep{lu2025netformer}, but view these directions as natural extensions for \methodname.
Incorporating biological priors (either through the choice of reference process \citep{petrovic2025curly, zhang2025learning} or by directly imposing knowledge of structural relationships on the network \citep{roohani2024predicting}) is another interesting direction for future work.
Lastly, using optimal couplings acquired using marginals across all timepoints is a directly applicable extension to \methodname, and may help improve overall performance of both structure learning and dynamical inference \citep{chen2023deepmomentummultimarginalschrodinger, theodoropoulos2025momentummultimarginalschrodingerbridge, rohbeck2025modeling}.}

\rev{We remark, \methodname is designed to be built upon and extended to all aforementioned settings. In this work, we provide a methodological and experimental foundation amenable to further development.}

\section*{Acknowledgments} The authors acknowledge funding from UNIQUE, CIFAR, NSERC, Intel, and Samsung. The research was enabled in part by computational resources provided by the Digital Research Alliance of Canada (\url{https://alliancecan.ca}), Mila (\url{https://mila.quebec}), the Province of Ontario, companies sponsoring the Vector Institute (\url{http://vectorinstitute.ai/partners/}), and NVIDIA. LA was in part supported by the University of Toronto, the Vector Institute, the Eric and Wendy Schmidt Center at the Broad Institute of MIT and Harvard, and by the NSERC Postdoctoral Fellowship, throughout this project. 

\newpage
\bibliographystyle{apalike}
\bibliography{main}

\newpage
\appendix

\section{Experimental Details}
\label{ap:expt_details}

\subsection{Neural Graphical Models for Dynamical Systems} 
\label{ap:ngm_architecture}

Unlike static graphical models, where variables are assumed to be independent and identically distributed, dynamical systems require modeling dependencies in continuous time.

Neural graphical models (NGMs) parameterize the vector field $\bm{v}(\bm{x}(t))$ using deep neural networks \citep{bellot2021neural}. Specifically, a neural dynamic structural model (NDSM) is defined as: 
$$
dx_j(t) \;=\; v^{\theta}_j(\bm{x}(t))\,dt \;+\; dw_j(t), \qquad \bm{x}(t_0)=\bm{x}_0
$$
where $v^{\theta}_j \in \mathcal F$ are analytic functions parameterized by neural networks with trainable weights $\theta \in \Theta$. Each function $v^{\theta}_j$ can be expressed as:
$$
v^{\theta}_j(\bm{x}) \;=\; \psi\!\Big(\psi\big(\cdots \psi(\,\bm{x}\,\theta_j^{A})\,\theta_j^{1}\cdots\big)\,\theta_j^{K}\Big),
$$
where $\theta_j^{A}\!\in\!\mathbb{R}^{d\times h}$ is the first (graph) layer, $\theta_j^{k}\!\in\!\mathbb{R}^{h\times h}$ for $k=1,\dots,K-1$ and $\theta_j^{K}\!\in\!\mathbb{R}^{h\times 1}$ are the deeper layers, and $\psi$ is an activation function. Directed edges in the inferred graph correspond to non-zero partial derivatives of these networks, and indicate model-based direct structural dependencies under the NGM assumptions between variables. The adjacency can be inferred if:
$$
G_{j i} \neq 0 \;\;\iff\;\; \big\|\partial_{i}\,v^{\theta}_j\big\|_{L_2} > 0
$$

Now to recover $G$, structure learning can be formulated as a penalized optimization problem:
$$
\arg\min_{v^\theta}\; R_n(\theta) \;=\; \min_{v^\theta}\; \frac{1}{n}\sum_{m=1}^{n}\big\|\bm{x}(t_m)-\hat{\bm{x}}(t_m)\big\|_2^{2},
$$
$$
\text{subject to}\;\; d\bm{x}(t)=\bm{v}^{\theta}(\bm{x}(t))\,dt \;\; \text{ and }\;\; \rho(\theta)\le \eta,
$$
where $\rho(\theta)$ is a regularization term. In \citep{bellot2021neural} this is often implemented using group lasso (GL) or adaptive group lasso (AGL):
$$
\rho_{\text{GL}}(\theta) \;=\; \lambda_{\text{GL}} \sum_{j=1}^{d}\sum_{i=1}^{d} \big\|\big(\theta_j^{A}\big)_{i,:}\big\|_2,
\qquad
\rho_{\text{AGL}}(\theta) \;=\; \lambda_{\text{AGL}} \sum_{j=1}^{d}\sum_{i=1}^{d} \frac{\big\|\big(\theta_j^{A}\big)_{i,:}\big\|_2}{\big\|\big(\hat{\theta}_j^{A}\big)_{i,:}\big\|_2^{\gamma}}.
$$
Here, $\lambda_{\text{AGL}}, \lambda_{\text{GL}}$ control the regularization strength, and $\big(\hat{\theta}_j^{A}\big)_{i,:}$ is an initial estimate of the parameters.

Given an estimate of $\,\bm{v}^{\theta}$, NGMs can be extended to irregularly sampled and non-linear time series data by numerically computing the forward trajectory with an ODE solver:
$$
\hat{\bm{x}}(t_1), \hat{\bm{x}}(t_2), \dots, \hat{\bm{x}}(t_n) \;=\; \mathrm{ODESolve}\big(\bm{v}^{\theta},\, \bm{x}(t_0),\, t_1, t_2, \dots, t_n\big)
$$
enforcing the constraint $\,d\bm{x}(t)=\bm{v}^{\theta}(\bm{x}(t))\,dt\,$ while optimizing for $R_n(\theta)$ \citep{chen2018neural}.
While the NGM was originally developed to identify causal dependencies in deterministic systems governed by ODEs, our work extends their applicability to stochastic systems by using the probability flow formulation of the underlying SDE. We use this terminology in the technical sense of the cited NGM framework, and do not claim general causal identifiability from population snapshots. Differently from \cite{bellot2021neural}, we train an NGM which parameterizes the probability-flow ODE, whose instantaneous drift is
$$
\hat{\bm{v}}_{t}(\bm{x};\Theta) \;=\; \bm{v}^{\theta}(\bm{x}) \;-\; \frac{\sigma^{2}}{2}\,\bm{s}^{\phi}_{t}(\bm{x}),
$$
which is equivalent in marginal behavior to the original SDE but removes the stochasticity by incorporating the score function $\nabla \log p_t(\bm{x})$. This allows us to retain the NGM for structure discovery, while still modeling data generated from inherently noisy dynamics.

\subsection{Reference Fitting (RF)}
\label{example_reference_fitting}
In contrast to \methodname, Reference Fitting \citep{zhang2024joint} (RF) starts from the principle that the inferred couplings $\pi$ should minimize entropy with respect to a reference process, with the joint-optimization problem defined as:
$$\min_{A\in{\mathbb{R}^{d\times d}},b\in{\mathbb{R}^d}}\min_{\pi\in\mathcal{C(\mu,\mu')}}\sigma^2\text{KL}\left(\pi|K^{\sigma}_{(A,b)}\right)+\mathcal{R}\left(A,b\right)$$
where $\mathcal{R}$ is a regularizer term, $A$ and $b$ are the linear and constant terms, respectively, of an Ornstein--Uhlenbeck (OU) process of the form:
$$\text{d}X_t=\left(AX_t + b\right)\text{d}t\ + \sigma\text{d}B_t$$
and $K^{\sigma}_{(A,b)}$ is the transition kernel of the OU process.\\ Perturbations can then be modeled in this set-up by the linear interaction matrix $A^{(g)}$, where $g$ represents a knocked-out gene, and thus the $g$-th is zeroed out. Thus, for any gene intervention (knockout) $g$, the reference OU process takes the form:
$$\text{d}X_t=\left(A\odot M^{(g)}\right)X_t^{(g)}\text{d}t\ + \sigma\text{d}B_t$$
where $M_{ij}^{(g)} = \bm{1}_{i\neq{g}}$ is the masking matrix for intervention $g$. This allows the RF process to learn interactions which may have very low signal in the observational (wild-type) data, and thus cannot be learned from the observed dynamics alone.
The transition kernel, $K^{\sigma}_{(A,b)}$, is then approximated by separating the drift and noise terms to model the kernel's mean and covariance, respectively:
$$\mu_t = e^{tA}x_0\text{, }\Sigma_t=\sigma^2tI$$
This yields a transition kernel of the form:
$$K_t(x,x')\propto\exp{\left(-\frac{||e^{tA}x-x'||^2_2}{2\sigma^2t}\right)}$$
Running an alternating optimization over the couplings $\pi$ and reference dynamics yields both the reference process itself and couplings which correspond to this process.

\paragraph{Trajectory fitting (ODE view).}
Given \(v^{\theta}\), irregularly sampled trajectories can be obtained by numerical integration,
\[
\hat{\bm{x}}(t_1),\dots,\hat{\bm{x}}(t_n)\;=\;\mathrm{ODESolve}\big(v^{\theta},\,\bm{x}(t_0),\,t_1,\dots,t_n\big),
\]
and one may minimize a data misfit \(R_n(\theta)=\tfrac{1}{n}\sum_{i=1}^{n}\|\bm{x}(t_i)-\hat{\bm{x}}(t_i)\|_2^2\) with the above sparsity penalty \citep{chen2018neural,bellot2021neural}.

\paragraph{Stochastic dynamics via probability flow.}
While NGMs were introduced for deterministic ODEs, we extend them to stochastic systems by parameterizing the \emph{probability-flow ODE} associated with an SDE. Let the instantaneous probability-flow drift be
\[
\hat{\bm{v}}_{t}(\bm{x};\Theta) \;=\; \bm{v}^{\theta}(\bm{x}) \;-\; \tfrac{\sigma^{2}}{2}\,\bm{s}^{\phi}_{t}(\bm{x}),
\]
where \(\Theta=\{\theta,\phi\}\), \(\bm{v}^{\theta}\) is the NGM (autonomous) drift defined above, and \(\bm{s}^{\phi}_{t}(\bm{x})=\nabla_{\bm{x}}\log p_t(\bm{x})\) is a score network. This PF-ODE induces the same marginals \(p_t\) as the original SDE but removes stochasticity by absorbing it into the score term. We therefore retain the NGM’s FC1 layer \(\theta^{A}\) for \emph{structure discovery} (via GL/AGL on \(A_{ji}=\|(\theta_j^{A})_{i,:}\|_2\)) while modeling data generated by noisy dynamics. This connects structure extraction to the PF-ODE used for rollout, while the training losses remain simulation-free.

\subsection{\rev{On Theoretical Guarantees for Structure Recovery from Population Dynamics via NGMs}}
\label{ap:identify}

\paragraph{Edge Recovery in the NGM and \methodname.}
\rev{Although our vector field is parameterized by a neural network, using a sparse weight matrix as the first layer of the network provides a principled approach to encode, and likewise to recover, the underlying interaction structure of the non-linear system \citep{bellot2021neural}. For continuous-time dynamics \cite{bellot2021neural} denote directed edge(s) $x_k \rightarrow x_j$ whenever the drift $f_j$ has a \textbf{functional dependence} on $x_k$, i.e. $\partial_k f_j \neq 0$.
They show that for any analytic neural network parameterization of $f_j$, every true functional dependency must pass through the input layer and conversely, if $\partial_k f_j = 0$, there exists an equivalent parameterization where the entire k-th column of the first-layer weight matrix is zero. This justifies the application group lasso solely to the first-layer weights, because zeros in that layer correspond exactly to absent edges in the underlying continuous system. \cite{bellot2021neural} show that with enough data, adaptive group lasso recovers the correct sparsity pattern. \methodname relies on the same principle because our autonomous vector field makes use of the same formulation of the NGM.}

\paragraph{On identifiability of underlying structure via the NGM and flow matching.}

\rev{In general, the question of identifiability for directed structure is challenging, even in the simplest possible case of linear dynamics (see e.g. \cite{wang2024recovering}). Similarly, \cite{zhang2024joint} were not able to provide an identifiability result but only show well-posedness of their optimization problem.} 

\rev{To our knowledge, identifiability in the joint trajectory inference and structure learning problem from population snapshots is an open problem and is particularly challenging due to the interdependence of the learned structure. 
We thus leave a general theoretical study of identifiability to future work. 
In what follows, we reason that the NGM with group lasso recovers the true underlying structure in the idealized setting where the drift function and score can be reconstructed exactly. 
We consider autonomous dynamics, where the drift field $v$ is given by a NGM model with parameters $\theta_0$
$$ dX_t = v(X_t; \theta_0) dt + \sigma dB_t.$$
Equivalently, this can be written as a probability flow, with $u_t(x) = v(x) - D \nabla \log p_t$ and $D = \frac{1}{2} \sigma \sigma^\top$: 
$$\partial_t p_t = -\nabla \cdot (p_t(x) u_t(x)).$$
Consider the case where samples of the flow $u_t(x)$ and score $\nabla \log p_t(x)$ can be accessed exactly, where $(t, x) \sim p_t(x)$.
Writing $\hat{v}_t(x; \theta)$ to be the reconstructed vector field, parameterized as a NGM with parameter $\theta$. The corresponding probability flow is
$$ \hat{u}_t(x) = \hat{v}(x) - D \nabla \log p_t. $$
The flow matching objective with target $u_t(x)$ is thus
$$\min_\theta \mathbb{E}_{t \sim [0, 1], x \sim p_t} \| \hat{u}_t(x) - u_t(x) \|_2^2 \Longrightarrow \min_\theta \mathbb{E}_{t} \mathbb{E}_{x \sim p_t} \| \hat{v}(x; \theta) - v(x; \theta_0)\|^2. $$
In this idealized setting, flow matching amounts to least-square regression on the target vector field $v(x, \theta_0)$. 
Let $\{ x_i \}_{i = 1}^N$ a sample of size $N$ drawn following $t \sim [0, 1]$ and $x \sim p_t$. Then, consider the empirical and population risks $R_N, R$
$$ R_N(\theta) = \min_\theta \frac{1}{N} \sum_{i = 1}^N \| \hat{v}(x_i; \theta) - v(x; \theta_0) \|^2, $$
$$R(\theta) = \mathbb{E}_{t, x} \| \hat{v}(x; \theta) - v(x; \theta_0) \|^2. $$} 

\rev{Our goal is to characterize identifiability of the structural graph from samples. Then this falls into the framework considered by \cite{dinh2020consistent}. Specifically, the results of [\cite{dinh2020consistent}, Theorem 3.6] guarantees local consistency of the learned graph from minimizing (1) in the limit where $n \to \infty$.}

\subsection{Experiment Details}
\label{app:exp_details}

\paragraph{Model architectures.}
The learned flow model is a NGM as described above. The score and residual models are standard multilayer perceptrons with ReLU activations. These support time-varying and conditionally encoded inputs (knockout vectors). All networks are initialized using Gaussian weights: $\mathcal{N}(0, 0.1)$ for weights and $\mathcal{N}(0, 1 \times 10^{-2})$ for biases.

\paragraph{Note on Entropic Optimal Transport}
We note that we choose to solve the EOT problem to pair successive time marginals (\smash{$\hat{p}^{(c)}_{t_i}$} and \smash{$\hat{p}^{(c)}_{t_{i+1}}$}) independently of the other marginals \smash{$\hat{p}_t \mid t \notin \{i, i+1\}$}, not globally across all time marginals like is done in \citet{chen2023deepmomentummultimarginalschrodinger, theodoropoulos2025momentummultimarginalschrodingerbridge, rohbeck2025modeling}. \methodname is flexible however, in that one would simply swap the coupling algorithm to solve the global multi-marginal problem instead with no additional changes to training. This could be used for other means as well, such as incorporating biological priors.

\paragraph{Trajectory simulation.}
Simulations are run between pairs of discrete timepoints $(t_k, t_{k+1})$ where $k = 0, \ldots, T-1$ using 100 Euler steps. We consider two rollout strategies: (a) stochastic differential equation (SDE) sampling and (b) probability flow ODE (PF-ODE) sampling. Both start from $p_{t_k}$, the empirical distribution of cells at timepoint $t_k$, but differ in how the dynamics are integrated.  

\subparagraph{ODE simulation.}
The probability flow ODE (PF-ODE), given by
\begin{equation}
\dot{x}_t = \bm{v}(x_t) - \tfrac{\sigma^2}{2}\bm{s}^{\phi}_t(x_t), 
\quad x_0 \sim p_{t_k},
\end{equation}
describes the deterministic evolution of the density $p_t$ using our learned score $\bm{s}^{\phi}_t$ and flow $\bm{v}(x_t)$. The score corrects the flow to match the evolving density, while the deterministic rollout removes sample-path noise, yielding smoother trajectories but potentially underestimating variability.

\paragraph{Evaluation metrics.}
We report Wasserstein-2 (W2) distance, squared Maximum Mean Discrepancy (MMD$^2$), and Energy Distance \cite{rizzo2016energy} between predicted and ground truth distributions. W2 is computed using the exact Earth Mover’s Distance from the POT library with a squared Euclidean cost matrix.

MMD$^2$ is computed using a radial basis function (RBF) kernel. For each evaluation, we average over five kernel bandwidths. Each bandwidth is defined via a kernel scale parameter $\sigma \in \{0.01, 0.1, 1, 10, 100\}$. The final MMD$^2$ score is the mean of the values computed under each of these five kernel settings.

Energy distance is computed according to \cite{rizzo2016energy}. For random vectors $X,\; Y \in \mathbb{R}^d$ with laws $F,G$, the squared energy distance is 
$$
\mathcal{D}^2(F,G) = 2\mathbb{E}\|X-Y\| \; - \; \mathbb{E}\|X-X'\| \; - \; \mathbb{E}\|Y-Y'\| \; \geq 0,
$$
where $X'$ and $Y'$ are i.i.d.\ copies of $X$ and $Y$. As described, it is the excess average cross-distance between two samples after subtracting their internal (within-sample) average distances, and is zero only when the two distributions coincide.

For structure learning evaluation, we compute the Area Under the Receiver Operating Characteristic curve (AUROC) and Average Precision (AP) of the learned graph against the ground-truth. AUROC measures the trade-off between true positive and false positive rates across all classification thresholds. AP summarizes the precision-recall curve as the weighted mean of precisions achieved at each threshold.

\paragraph{Experiment protocols.}
\begin{itemize}
  \item \textbf{Leave-one-out (timepoint):} For each dataset, we iteratively exclude data from a single discrete timepoint $t_k$ for $k \in \{1, \dots, T-2\}$ from training (we only hold out the interior timepoints). The model is trained on the remaining timepoints $\{t_j : j \in \{0, \dots, T-1\} \setminus \{k\}\}$ and then used to simulate the transition from timepoint $t_{k-1}$ to $t_k$. We report the average performance across all held-out timepoints by training a separate model for each excluded $t_k$. 

  \item \textbf{Leave-one-out (knockout):} For each dataset, we select the first three gene knockouts and exclude one per trained model. Each model is then used to simulate transitions from $t_k$ to $t_{k+1}$ conditioned on the held-out intervention across discrete timepoints $k = 0, \ldots, T-2$. A separate model is trained for each held-out knockout, and results are averaged across the three excluded interventions.
\end{itemize}

\paragraph{Seed values.}
All experiments are repeated across fixed seed values. For structure learning experiments, we use 5 seeds: \{1, 2, 3, 4, 5\}. For computationally intensive experiments (leave-one-timepoint-out and leave-one-knockout-out) we use 3 seeds: \{1, 2, 3\}

\paragraph{Hardware.} 
All experiments were implemented in PyTorch and run on a MacBook Pro with an Apple M1 Pro CPU and 32GB RAM. No GPU is needed. All models train in under 10 minutes.

\subsection{Hyperparameters}
\label{app:hyperparams}

We report all hyperparameters used in training our models. Unless otherwise specified, we use the AdamW optimizer with a fixed weight decay of $1 \times 10^{-2}$.

\paragraph{Synthetic datasets.}
Models are trained for 15{,}000 steps using a batch size of 64 and a learning rate of $3 \times 10^{-3}$. Regularization and model-specific settings are provided below:

\begin{table}[h]
\centering
\caption{Hyperparameters for synthetic datasets.} 
\label{tab:hyperparams_synthetic}
\begin{tabular}{@{}lc@{}}
\toprule
\textbf{Hyperparameter} & \textbf{Value} \\
\midrule
Training steps & 15{,}000 \\
Batch size & 64 \\
Learning rate & $3 \times 10^{-3}$ \\
Weight decay & $1 \times 10^{-2}$ \\
$\alpha$ (Score term influence) & 0.1 \\
$\ell_2$ regularization & $5 \times 10^{-6}$ \\
Residual regularization & $1 \times 10^{-3}$ \\
group lasso regularization & 0.04 \\
Knockout hidden layers & [100] \\
Score model hidden layers & [100, 100] \\
Residual model hidden layers & [64, 64] \\
SDE noise scale $\sigma$ & 1.0 \\
\bottomrule
\end{tabular}
\end{table}

\paragraph{Real data.}
For the real data experiments, some hyperparameters are overriden:

\begin{table}[h]
\centering
\caption{Overriden Hyperparameters for real data (RENGE).}
\label{tab:hyperparams_renge}
\begin{tabular}{@{}lc@{}}
\toprule
\textbf{Hyperparameter} & \textbf{Value} \\
\midrule
Training steps per fold & 10{,}000 \\
Learning rate & $0.07$ \\
$\ell_2$ regularization & $1 \times 10^{-9}$ \\
$\alpha$ (Score term influence) & 0.73 \\
group lasso regularization & 0.0008 \\
Knockout hidden layers & [128] \\
Score model hidden layers & [128, 128] \\
Residual model hidden layers & [128, 128] \\
\bottomrule
\end{tabular}
\end{table}

We used Optuna to perform hyperparameter optimization, and selected configurations that maximized the area under the precision-recall curve (AUPR). The search space included key hyperparameters such as the learning rate, regularization strength, group lasso regularization, batch size, and model capacity (e.g., number of hidden units). For each trial, model performance was evaluated using cross-validation, and the trial achieving the highest mean AUPR across folds was selected as the optimal configuration. AUPR is computed only on the available ChIP-seq reference edges for the 18 TFs used in evaluation; because the available reference edges are only partial, the resulting AP/AUROC should be interpreted as benchmark agreement with known or curated edges rather than an unbiased estimate of whole-network recovery. We note that the score term was of higher importance here. This is likely due to the fact that scRNA-sequencing data is much noisier than the synthetic systems.

\subsection{Synthetic Linear System}
\label{ap:synthetic_linear_system}

We generate synthetic linear systems to evaluate structure learning performance across varying dimensionality and sparsity levels. The underlying network structure is generated using Erd\H{o}s--R\'{e}nyi random graphs \citep{erdos59a}, where edges are placed independently with probability $p$ to achieve the desired sparsity level (5\%, 20\%, or 40\% of possible edges).

For a system with $d$ variables, we construct a random adjacency matrix $\bm{A} \in \mathbb{R}^{d \times d}$ where non-zero entries are sampled to avoid weak interactions. Specifically, edge weights are drawn from the union of two uniform distributions: $\mathcal{U}(-1.0, -0.5) \cup \mathcal{U}(0.5, 1.0)$, ensuring all edges have magnitude $\geq 0.5$. Positive and negative edges occur with equal probability, representing activating and inhibiting interactions respectively. The linear SDE governing the system dynamics is:
\begin{equation}
d\bm{x}_t = \bm{A}\bm{x}_t dt + \sigma d\bm{B}_t
\end{equation}
where $\bm{x}_t \in \mathbb{R}^d$ represents the system state, $\bm{A}$ encodes the linear interactions, and $\sigma = 0.1$ is the noise level. We simulate trajectories over $T=5$ time points, generating $N=1000$ samples per time point. The initial conditions are sampled from a multivariate Gaussian distribution $\mathcal{N}(\bm{0}, 0.5\bm{I})$.

\subsection{BoolODE}
\label{ap:boolode}

\begin{figure}[t]
    \centering
    \includegraphics[width=0.85\textwidth]{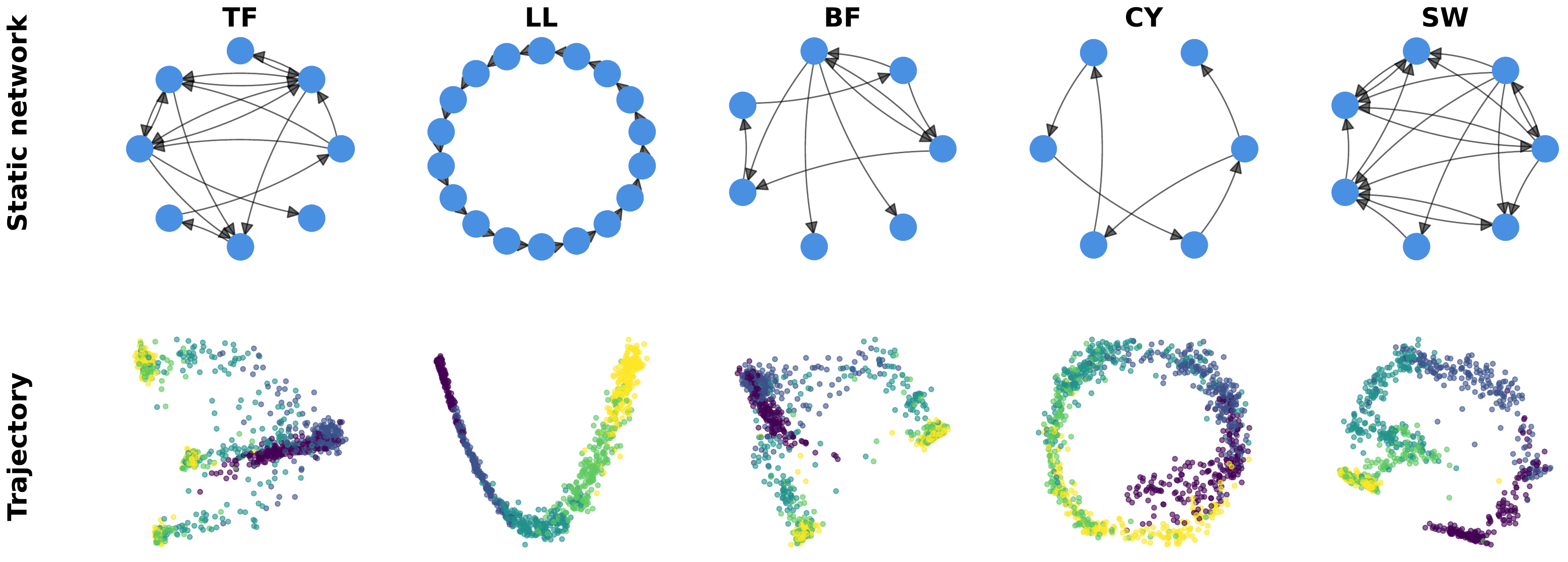}
    \caption{\textbf{BoolODE synthetic biological systems.} Network topologies and example trajectory visualizations for the six synthetic biological systems used in our experiments: trifurcating (TF), cyclical (CY), long linear (LL), swirling (SW), bifurcating (BF), and hematopoietic stem cell (HSC) systems. Figure is sourced from \citet{zhang2024joint}.}
    \label{fig:boolode_systems}
\end{figure}

BoolODE was employed to simulate 1000 cells independently for each trajectory. To obtain time-resolved snapshots, we divided the simulation into T=5 discrete intervals. Expression outputs from simulations were log-transformed for use in subsequent analysis. Knockout trajectories were produced by altering Boolean rules so that the targeted gene retained only self-activation, while its initial expression level was set to zero.

To evaluate performance across different conditions, we considered gene expression values across across the following conditions for each dataset type:
\begin{itemize}
\item HSC: Gata1, Fli1, Fog1, Eklf, Scl, Gfi1, EgrNab, cJun
\item TF, BF, CY, SW, and LL: observational (wild-type), g2, g3, g4, g5, g6, g7, g8
\end{itemize}
Since we considered 1000 cell expression trajectories, each measured across 800 continuous time and pseudo-time values, and divided across eight conditions, we were left with 100,000 cell measurements per condition. This was then further discretized into T=5 time bins.

\subsection{Single-cell CRISPR perturbation time-series}
\label{ap:real_data_system}

\begin{figure}[h]
    \centering
    \begin{subfigure}[b]{0.48\textwidth}
        \centering
        \includegraphics[width=\textwidth]{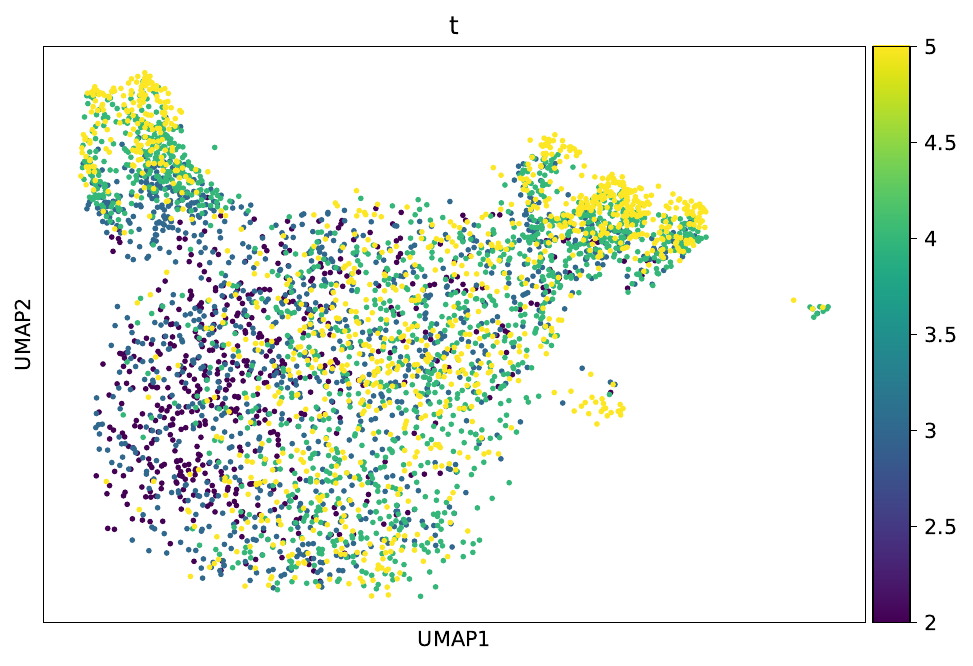}
        \caption{Time (Days 2-5)}
        \label{figs:umap_time}
    \end{subfigure}
    \hfill
    \begin{subfigure}[b]{0.48\textwidth}
        \centering
        \includegraphics[width=\textwidth]{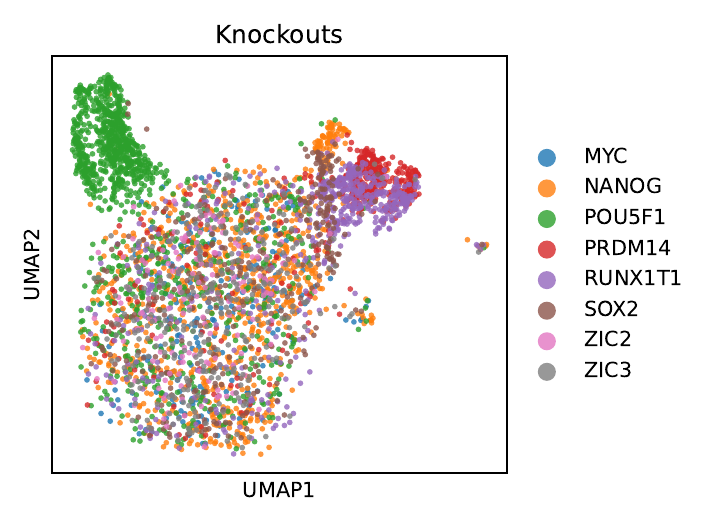}
        \caption{Interventions (knock-outs)}
        \label{figs:scCRISPR_knockouts_umap}
    \end{subfigure}
    
    \caption{\textbf{Visualization of real (Renge) dataset.} (a) UMAP visualization showing temporal progression of wildtype (unconditional) data across days 2-5, with colors indicating time points. (b) UMAP visualization of intervention conditions showing trajectories under different gene knock-out perturbations, with colors indicating the specific knocked-out gene.}
    \label{fig:umaps_renge_data}
\end{figure}

Following \citet{zhang2024joint}, we use raw count data from \citet{ishikawa2023renge}, retrieved from the Gene Expression Omnibus database (accession GSE213069). Zhang et al. use the following procedure: 
\begin{itemize}[noitemsep, topsep=0pt]
  \item Columns corresponding to gRNAs were removed.
  \item Counts were normalized using the \texttt{scanpy.pp.normalize\_total} function with default options.
  \item The data was log-transformed.
  \item Prior to dimensionality reduction, highly variable genes were selected using \seqsplit{scanpy.pp.highly\_variable\_genes}.
\end{itemize}
We considered the set of 103 transcription factors that \citet{zhang2024joint} uses and included only cells that received a single knockout. To construct the ChIP-seq reference, we obtained experimental binding information from ChIP-Atlas \citet{oki2018chip} for the following TFs (within a 1kb window): Chd7, Ctnnb1, Dnmt1, Foxh1, Jarid2, Kdm5b, Med1, Myc, Nanog, Nr5a2, Pou5f1, Prdm14, Sall4,
Sox2, Tcf3, Tcf7l1, Ubtf, Znf398 with a 1kb window. We then further reduced the knockouts to a group of 8 most prominent ones as referenced in \cite{zhang2024joint}. We show a visualization of the dataset in UMAP space in Figure \ref{fig:umaps_renge_data}.

\subsection{\rev{Extension to Anisotropic Diffusion}}
\label{ap:anisotropic}

\rev{For clarity of exposition, Sections~\ref{sec:prelims} and~\ref{sec:method}
assumed isotropic diffusion $\bm{\sigma} = \sigma\mathbf{I}$. The framework extends
naturally to anisotropic diffusion $\bm{\sigma} \in \mathbb{R}^{d\times d}$ with
only minor changes to the conditional targets and the EOT cost.}

\paragraph{\rev{Anisotropic conditional targets.}}
\rev{Let $\bm{\Sigma} = \bm{\sigma}\bm{\sigma}^\top$. The Brownian bridge
conditioned on endpoints $\bm{z} = (\bm{x}_0, \bm{x}_1)$ has conditional score
\begin{equation*}
\nabla_{\bm{x}} \log p_t(\bm{x} \mid \bm{z}) =
\frac{1}{t(1-t)}\, \bm{\Sigma}^{-1}
\bigl( t\bm{x}_1 + (1-t)\bm{x}_0 - \bm{x} \bigr),
\end{equation*}
while the conditional probability flow $\bm{v}^\circ_t(\bm{x}\mid\bm{z})$ retains
the same form as in the isotropic case, since the displacement field does not
depend on $\bm{\Sigma}$.}

\paragraph{\rev{Anisotropic EOT coupling.}}
\rev{The squared Euclidean cost in Eq.~(6) is inappropriate when $\bm{\Sigma}$ is
non-isotropic. The correct EOT problem uses a Mahalanobis-type cost,
\begin{equation*}
\pi^\star = \arg\min_{\pi \in \Pi(q_0, q_1)}\;
\tfrac{1}{2}\,\mathbb{E}_\pi \bigl[(\bm{x}-\bm{y})^\top \bm{\Sigma}^{-1}
(\bm{x}-\bm{y})\bigr]
\;+\; \varepsilon\, \mathrm{KL}(\pi \,\|\, q_0 \otimes q_1),
\end{equation*}
which still admits efficient solution via Sinkhorn iterations after a
preconditioning step.}

\paragraph{\rev{Reduction to diagonal diffusion.}}
\rev{Beyond these direct extensions, linear SDEs with arbitrary diffusion can be
reduced to an equivalent SDE with diagonal diffusion via a change of basis, as
formalized below.}

\begin{lemma}
\rev{Up to a change of basis, any linear SDE is equal in law to a linear SDE
with diagonal diffusion. Concretely, take $\diff \bm{X}_t = \bm{A}\bm{X}_t \diff t
+ \bm{\sigma}\,\diff \bm{B}_t$ with drift $\bm{A} \in \mathbb{R}^{d\times d}$ and
diffusion $\bm{\sigma} \in \mathbb{R}^{d\times d}$. Then there exists an
orthogonal $\bm{U} \in \mathbb{R}^{d\times d}$ such that
$\bm{Y}_t = \bm{U}^\top \bm{X}_t$ obeys a linear SDE with drift matrix
$\bm{U}^\top \bm{A} \bm{U}$ and a diagonal diffusion matrix $\bm{\Lambda}$.}
\end{lemma}

\begin{proof}
\rev{Let $\bm{D} = \bm{\sigma}\bm{\sigma}^\top / 2$. By the spectral theorem,
$\bm{D} = \bm{U}\bm{\Lambda}^2\bm{U}^\top / 2$ for some orthogonal $\bm{U}$
and $\bm{\Lambda} = \mathrm{diag}(\bm{\lambda}) \succeq 0$. Consider the two
SDEs
\begin{align*}
(\mathrm{a})\quad & \diff\bm{X}_t = \bm{A}\bm{X}_t\,\diff t
  + \bm{\sigma}\,\diff\bm{B}_t, \\
(\mathrm{b})\quad & \diff\bm{X}_t = \bm{A}\bm{X}_t\,\diff t
  + \bm{U}\bm{\Lambda}\bm{U}^\top\,\diff\bm{B}_t.
\end{align*}
The generator of~(a) is
$(\mathcal{L}f)(\bm{x}) = \langle \partial_{\bm{x}} f(\bm{x}),\bm{A}\bm{x}\rangle
+ \tfrac{1}{2}\mathrm{tr}\bigl(\bm{\sigma}^\top \partial_{\bm{x}\bm{x}} f(\bm{x})\,\bm{\sigma}\bigr)$,
and that of~(b) is
$(\tilde{\mathcal{L}}f)(\bm{x}) = \langle \partial_{\bm{x}} f(\bm{x}),\bm{A}\bm{x}\rangle
+ \tfrac{1}{2}\mathrm{tr}\bigl(\bm{U}\bm{\Lambda}\bm{U}^\top \partial_{\bm{x}\bm{x}} f(\bm{x})\,
\bm{U}\bm{\Lambda}\bm{U}^\top\bigr)$. Since
$\mathrm{tr}(\bm{U}\bm{\Lambda}\bm{U}^\top \partial_{\bm{x}\bm{x}} f\,\bm{U}\bm{\Lambda}\bm{U}^\top)
= \mathrm{tr}(\bm{\sigma}^\top \partial_{\bm{x}\bm{x}} f\,\bm{\sigma})$, the
generators coincide and the SDEs are equal in law.

Orthogonal transformations preserve Brownian increments, so
$\bm{U}^\top\,\diff\bm{B}_t = \diff\tilde{\bm{B}}_t$ for a Brownian motion
$\tilde{\bm{B}}_t$. Substituting into~(b), we obtain
$\diff\bm{X}_t = \bm{A}\bm{X}_t\,\diff t
+ \bm{U}\bm{\Lambda}\,\diff\tilde{\bm{B}}_t$, and the result follows.}
\end{proof}

\section{Implementation Details}
\label{ap:implementation_details}

\subsection{\methodname Algorithm}
\begin{algorithm}[H]
\caption{\methodname Training Loop}
\label{alg:sf2m}
\begin{algorithmic}[1] %
    \Require Training data across conditions \(c\), hyperparameters \(\alpha, \sigma, \lambda_{\text{GL}}\), number of steps \(N\), \(d\) variables
    \State Initialize parameters \(\Theta = \{\theta, \phi\}\), Optimizer \(Opt\)
    \For{step \(n=1\) to \(N\)}
        \State $\bm{M}^{(c)} \leftarrow M^{(c)}_{ji} = 0 \text{ if } i=c, j \neq c, M^{(c)}_{ji} = 1 \text{ otherwise} \textbf{ for all } j, i \in {1,...,d}\times {1,...,d}$
        \State $\bm{k}^{(c)} \leftarrow k^{(c)}_j = 1 \text{ if } j \text{ perturbed under } c, \text{ else } k^{(c)}_j = 0 \textbf{ for all } j$
        \State Sample \((\bm{x}_t, t, \bm{s}_t, \bm{v}_t^{\circ})\) from SB for condition \(c\)
        \State \(\bm{s}^{\phi}_t \leftarrow \bm{s}_t^{\phi}(\bm{x}_t|\bm{k}^{(c)})\)
        \State $ \hat{\bm{v}}_t \leftarrow 
        \bm{v}^{\theta_j}(\bm{x}_t \mid \bm{M}^{(c)})
        - \tfrac{\bm{\sigma}^2}{2}\,
        \bm{s}^{\phi}_t\!\left(\bm{x}_t \mid \bm{k}^{(c)}\right) $
        \Comment{Using $ \bm{M}^{(c)} \odot \theta^A_j $}
        \State \(\mathcal{L} \leftarrow \sum_c \mathbb{E}\left[\left(1-\alpha\right)\left\Vert\hat{\bm{v}}_t(\bm{x}_t;\Theta |\bm{M}^{(c)},\bm{k}^{(c)}) -\bm{v}^{\circ}_t\right\Vert^2+\alpha\left\Vert\bm{s}_t^{\phi}(\bm{x}_t|\bm{k}^{(c)})-\bm{s}_t\right\Vert^2\right]\)
        \State \(\bm{g} \leftarrow \nabla_{\Theta} \mathcal{L}\)
        \State Update parameters \(\Theta \leftarrow Opt(\Theta, \bm{g})\)
        \State \( \theta^A_j \leftarrow \mathrm{Prox}_{\,\eta\,\lambda_{\mathrm{GL}} \|\cdot\|_{\text{group}}}\!\big(\theta^A_j\big)\) \Comment{group-wise proximal update on first layer}
    \EndFor
\end{algorithmic}
\end{algorithm}

\subsection{Gene Regulatory Network Inference Baselines}

\rev{All baselines were trained with the same pre-processing steps, train-test splits, and evaluation metrics. In addition, we experimented with the hyper-parameters of the baseline methods. For methods that considered the same BoolODE dataset(s) \citep{zhang2024joint, Zhao2024OTVelo}, we used the respective hyper-parameters reported in the work. For other methods, we experimented with hyper-parameters settings, but found that the hyper-parameters set by the authors of the respective baseline methods performed generally well.}

\paragraph{SCODE:}
Sparse COding for Differential Equations (SCODE) from \citeSupp{matsumoto2017scode} is a method for reconstructing GRNs from single-cell time series by fitting a linear ODE system to gene expression trajectories. The method infers a sparse gene interaction matrix such that simulated expression profiles recapitulate observed temporal dynamics. In the synthetic case, we take the transposed expression matrix and pseudotime vector that is given to us in both the observational and interventional settings and write them to disk in a format compatible with SCODE. We then execute the model via the Ruby + R wrapper developed by \citeSupp{matsumoto2017scode}, and read the mean interaction matrix from the resulting output files, which gives us an adjacency matrix. SCODE does not work with interventional data so we do not expect an increase in performance. 
 
\paragraph{SINCERITIES:} SINgle CEll Regularized Inference using TIme-stamped Expression profileS \citep{gao2018sincerities} is an R-based algorithm for inferring directed GRNs from time-stamped cross-sectional single-cell data. It quantifies how each gene's expression distribution shifts between consecutive timepoints and formulates GRN inference as a ridge-regularized linear regression on those distributional changes; edge signs (activation vs. repression) are assigned via partial correlation between gene pairs.

\paragraph{GLASSO:} Graphical Lasso is a classical method for inferring sparse undirected graphical models from high dimensional data. GLASSO estiamtes the precision (inverse covariance) matrix under an $l_1$ penalty to enforce sparsity, such that nonzero entries correspond to conditional dependencies between genes. This method does not work with time-series data. We first standardize the gene expression matrix, and choose the regularization parameter via cross-validation. The estimated precision matrix is negated (higher values correspond to stronger interactions), and the diagonal is set to zero. Our result is an adjacency matrix with symmetric edge weights that represent partial correlations between genes. 

\paragraph{dynGENIE3:} This method from \cite{huynh2018dynGENIE3} is an extension of the GENIE3 algorithm \cite{huynh-thu2010genie3} to reconstruct GRNs from time-series single-cell or bulk gene expression data. The method frames network inference as a feature selection problem, where for each gene, a random forest regression model is \rev{trained} to predict its expression at each time point as a function of all other genes' expressions at earlier time points. The regulatory strength is aggregated over all of the trees and time lags, and yields a directed adjacency matrix. We aggregated our expression matrix \texttt{adata.X}, which is an annData object, by time points (mean expression per time) and store it in a list \texttt{X}, and keep the corresponding points in \texttt{t}. The dynGENIE3 model is then called with these two parameters as input and we save the resulting adjacency matrix. \rev{By averaging across all cells at each time $t$ we effectively create pseudo-bulk expressions, then stack these per-timepoint means into a matrix so as to be a single trajectory. This approach is corroborated by \cite{moscardo2025multi} who use a nearly identical procedure for dynGENIE3.}

\paragraph{RENGE:} RENGE \cite{ishikawa2023renge} can use both temporal and interventional data. Renge uses pseudotime trajectories from single-cell data, and CRISPR knockout information, with a two-step inference strategy, that involves a regression model to predict expression as a function of genes, and network aggregation to aggregate the inferred effects across all perturbations and time points. This gives us a gene $\times$ gene matrix based on both the endogenous dynamics and systematic perturbation present in the dataset. The main experiment from \cite{ishikawa2023renge} was what was used in our paper and in \cite{zhang2024joint}. We use the model that \cite{ishikawa2023renge} provide in their codebase and do further experimentation, using that matrix as a baseline. 

\paragraph{TIGON:} Trajectory Inference with Growth via Optimal transport and Neural networks (TIGON) \citep{sha2024reconstructing} models time-series scRNA-seq as a continuum of cell states whose density evolves under a reaction-transport dynamic. It represents a time-dependent cell-state density represented by two fields: a velocity field that models state transitions, and a growth field that accounts for birth/death and mass change. TIGON integrates the learned dynamics using ODE integration per time step, and learns to minimize the Wasserstein-Fisher-Rao distance. At inference, TIGON learns to reconstruct the continuous trajectories and velocities for individual cells, estimates time-resolved growth maps, and can learn a gene-regulatory structure from the Jacobian of its flow. 
\rev{TIGON uses either a PCA layer or an autoencoder to first encode the ambient data. This is a step for the ``dimensionless-solver'' used within TIGON, which is optimized to operate in 10 dimensional space. All of the other baselines, and \methodname, are trained directly in the ambient space. Thus, in order to provide a comparative setting, we train TIGON in the ambient space as well and forgo the PCA/autoencoder step. We did not encounter challenges with applying TIGON to the high dimensional real (Renge) biological system. TIGON uses Gaussian mixture models to represent $p_t$ (equivalent to our Eq. \ref{eq:fpe}).
We also train TIGON in PCA space, and then convert to the ambient space using the following formula: $J_X = W^T J_z W$ where $J_z$ is the latent Jacobian. We note no difference in performance for structure learning.} 

\paragraph{OTVelo:} 
\rev{OTVelo \cite{Zhao2024OTVelo} estimates gene velocity from cell trajectories reconstructed via optimal transport. The method proceeds in two stages. The first stage is trajectory inference where it aligns cell across consecutive time-points using Fused Gromov--Wasserstein (FGW) optimal transport, which uses both expression state and local geometric structure. The second part uses OT pairings to estimate the gene velocities via two different approaches. We consider two versions of this model, OTVelo-Corr and OTVelo-Granger. OTVelo-Corr uses time-lagged correlation analysis that generalizes the standard time-lagged correlation to distributions of cells matched by the OT pairings. OTVelo-Granger, on the other hand, is a regularized linear regression model (Elastic Net) where the velocity of a target gene at time $t+1$ is predicted from the velocities of regulating gene at time $t$. Given that we discretize all of our data into bins, and OTVelo tests their model on the same datasets, simulated using BoolODE, we apply the model from \cite{Zhao2024OTVelo} directly in our pipeline.}

\begin{table}[t]
\centering
\caption{ {Assumptions and capabilities of StructureFlow and baseline methods. 
We indicate whether each method uses interventional (knockout) data, requires 
discrete time labels, relies on pseudo-trajectories or OT pairings to recover 
temporal correspondence, applies a dimensionality-reduction step (PCA/autoencoder) 
before inference, and whether it addresses the \emph{joint} structure-and-dynamics 
task or only one of the two.}}
\label{tab:baseline-assumptions}
\footnotesize
\setlength{\tabcolsep}{4pt}
\renewcommand{\arraystretch}{1.2}
 {
\begin{tabular}{lccccc}
\toprule
\textbf{Method} & \textbf{Interventions} & \textbf{Time labels} & \textbf{Pseudo-traj.} & \textbf{PCA/AE} & \textbf{Joint task} \\
\midrule
\textbf{StructureFlow (ours)} & Yes (masks) & Yes & No & No & Yes \\
$[\text{SF}]^2$M           & Cond.\ input & Yes & No & No & No (dynamics only) \\
RF                          & Yes (mask on $A$) & Yes & No & No & Yes (linear) \\
TIGON                       & No & Yes & No & Optional & Yes (Jacobian) \\
OTVelo                      & No & Yes & OT pairings & No & Yes \\
dynGENIE3                   & No & Yes & Pseudo-bulk & No & Structure only \\
SINCERITIES                 & No & Yes & No & No & Structure only \\
SCODE                       & No & Yes (pseudotime) & No & No & Structure only \\
Glasso                      & No & No & No & No & Structure only (undirected) \\
RENGE                       & Yes & Yes (pseudotime) & No & No & Structure only \\
PCMCI (adapted)             & Both & Yes & OT pseudo-traj. & No & Structure only \\
\bottomrule
\end{tabular}}
\end{table}

\subsection{Unbalanced \methodname}
\label{ap:unbalancedSF}
\rev{In order to properly model cell proliferation and death, we formalize the extension of \methodname to biological systems where probability mass is not conserved. We generalize the probability flow formulation to include a growth term, derived from Unbalanced Optimal Transport (UOT).}

\paragraph{\rev{Unbalanced Optimal Transport}}
\rev{In the standard \methodname framework, couplings $(x_0, x_1) \sim \pi(x_0, x_1)$ are acquired via a balanced Sinkhorn solver, which enforces strict mass conservation $(\sum \pi_i = \frac{1}{N})$. To model cell proliferation and death, we replace this with UOT. The generality of our method allows us to easily incorporate unbalancedness into \methodname by rescaling source and
target measures \cite{eyring2024unbalancednessneuralmongemaps}. We compute the coupling $\pi^*$ between empirical snapshots $\hat{\rho}_{t_k}$ and $\hat{\rho}_{t_{k+1}}$ by minimizing the unbalanced Sinkhorn Divergence \cite{wang2025jointvelocitygrowthflowmatching, chizat2019unbalancedoptimaltransportdynamic}:
$$\pi^* = \arg \min_{\pi \in \mathbb{R}_+^{N \times M}} \sum_{i,j}C_{ij}\pi_{ij} \; + \epsilon H(\pi) \; + \rho[KL(\pi_{\#0} |\hat{\rho}_{t_k}) \; + KL(\pi_{\#1}|\hat{\rho}_{t_{k+1}})]$$
where $\epsilon H(x)$ is the entropic regularization coefficient, $\rho$ is the marginal relaxation parameter, and $\pi_{\#0}$ are row sums which control the amount of mass transported away from each starting cell. If $\pi_{\#0} < \hat{\rho}_{t_k}$ then the transport plan used less mass than was available and if $\pi_{\#1} > \hat{\rho}_{t_{k+1}}$ then the transport plan delivered more mass than was originally available relative to the source.}

\paragraph{\rev{Stochastic Dynamics with Growth}}
\rev{To account for cellular growth, we adopt the formulation proposed in \cite{zhang2025inferringstochasticdynamicsgrowth}, where the population density $\rho_t(x)$ evolves according to a Fokker--Planck equation with a source term: 
$$\partial_t \rho_t(x) = - \nabla \cdot (\rho_t(\bm{x}_t) \bm{v}_t(\bm{x}_t)) + \textstyle\frac{1}{2}\nabla \cdot (\sigma^2 \nabla \rho_t(\bm{x}_t)) \; +g_t(x)\rho_t(x)$$
where $\bm{v}_t(\bm{x})$ represents the drift term and \smash{$\frac{1}{2}\nabla \cdot (\sigma^2 \nabla \rho_t(\bm{x}_t))$} is the diffusion term. Unlike \cref{eq:fpe} where probability mass is conserved, the total mass $M_t = \int \rho_t(x)dx$ varies over time according to the net growth rate of the system. We model the joint evolution where the local change in log-density is governed by both the divergence of the flow and the local growth rate, defined as:
$$\frac{d}{dt}\log\rho_t(x_t) = g_t(x_t) - \nabla \cdot u_t(x_t)$$ 
where $u_t(x)$ is defined in \cref{eq:PFODE}. This formulation then allows us to disentangle the vector field $v_t$ from the growth field, so that the density increase is due to the growth field, and the vector field only models transitions.
Lastly to learn the growth field $g_\phi(x)$, we adopt the simulation-free strategy proposed in \cite{wang2025jointvelocitygrowthflowmatching}. Instead of solving a differential equation for mass evolution during training, we interpret the row-sums of the unbalanced coupling matrix as static proxies for net proliferation accumulated along a trajectory. For a source sample $x_0^i$, the target growth $G_i$ is defined by the log-marginal of the transport plan $\pi^*$:
$$G_i = \log(\sum_j\pi^*_{ij})$$
This value represents the total probability mass at $t_1$ attributed to the lineage of $x_0^i$. If the sum is greater than 1, the cell has proliferated, otherwise the cell has undergone apoptosis (died). To learn the growth network, we regress the neural network $g_\phi(x,t)$ to the static target across the interpolated trajectory:
$$\mathcal{L}_{\text{growth}}(\phi) = \mathbb{E}_{t,(x_0^i,x_1^j)\sim \pi^*}[\|g_\phi(x_t)-G_i\|^2]$$}

\section{Novelty Statement}
\label{ap:novelty}

\rev{This work introduces key methodological and empirical contributions to joint structure learning and dynamics inference:}

\rev{\textbf{Decomposition of Stochastic Dynamics.} We decompose the probability flow into an autonomous drift $\bm{v}(\bm{x})$ capturing system structure and a time-dependent score function $\bm{s}_t(\bm{x})$ modeling stochasticity. This aligns with biophysical reality where regulatory mechanisms are time-invariant.}

\rev{\textbf{Conditional Dynamics via Interventions.} We incorporate interventional data through knockout conditioning masks, enabling learning from perturbational experiments and interventional trajectory inference on unseen perturbations.}

\rev{\textbf{Simulation-Free Training for High-Dimensional Systems.} Our framework extends simulation-free training to conditional settings, enabling tractable learning on high-dimensional systems (up to 500 variables) while jointly inferring structure and dynamics.}

\rev{\textbf{Comprehensive Evaluation.} We establish extensive benchmarks including scalability analysis, evaluation on six biologically plausible systems with interventional data, and validation on real single-cell CRISPR time-series data with realistic constraints (4-5 time points).}

\section{Broader Impact}
\label{ap:broader_impact}
The genome is the fundamental identity of every living thing. The expression of DNA dictates life. It is a multi-dimensional representation of the state of any living thing. The ultimate goal of our work is to enable improve our understanding of how an organism’s genes interact, allowing us to predict the trajectory of its life, and ultimately to computationally develop interventions that can positively change that trajectory. The stochastic dynamical systems we consider here are used ubiquitously in the biophysics and modelling literature. In a biological context, our vector field $\bm{v}$ can be related to the metaphorical ``Waddington's landscape''.

\section{Additional Results}
\label{ap:additional_results}

\subsection{Structure Learning on Synthetic Systems with Observational Data}
\label{ap:syn_structure_obs}

\begin{figure}[htbp]
    \centering
    \includegraphics[width=\textwidth]{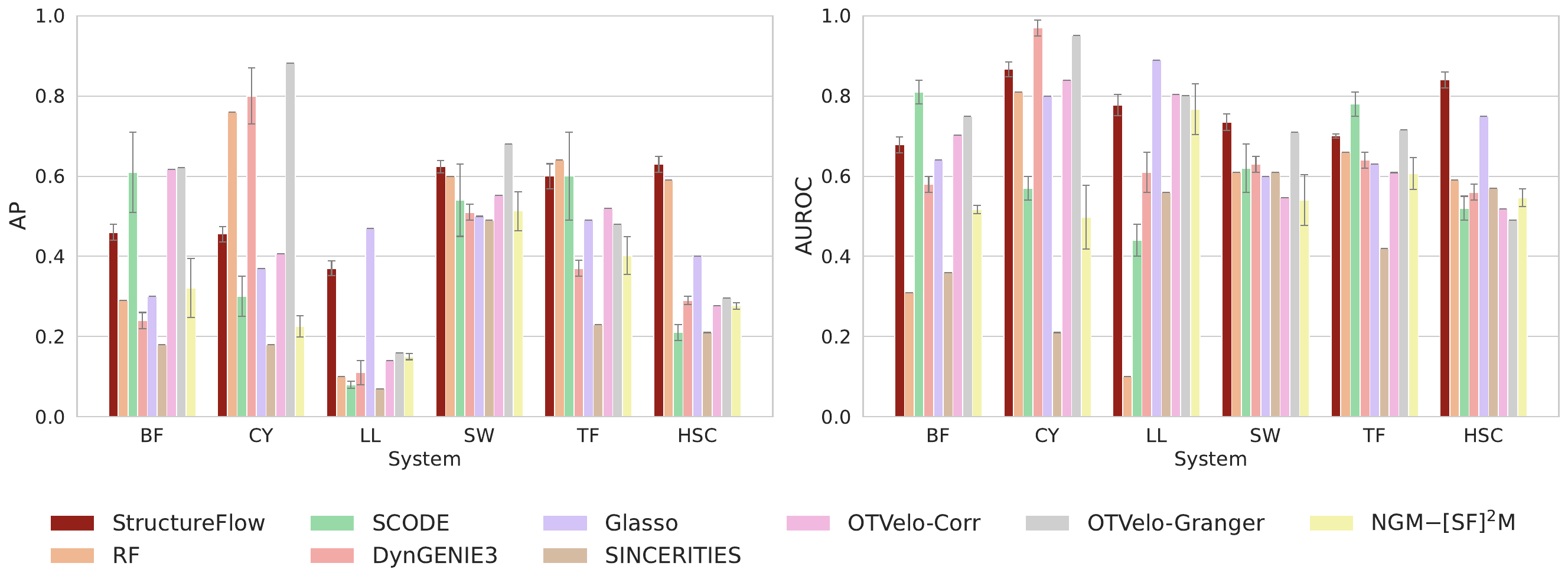}
    \caption{\textbf{Structure learning performance across synthetic systems using only observational data}. Shown are average precision (AP) and area under the ROC curve (AUROC) scores for models trained on only observational data}
    \label{fig:AP_AUROC_wildtype}
\end{figure}

In Figure~\ref{fig:AP_AUROC_wildtype}, we observe that \methodname achieves competitive results in most systems across both AP and AUROC metrics. Whereas other methods tend to fluctuate widely in performance across these synthetic systems, \methodname underperforms all other methods only on the SW system, while beating most other benchmarks otherwise.

\begin{figure*}[p] %
  \centering
  \begin{subfigure}{\textwidth}
    \centering
    \includegraphics[width=\textwidth, height=0.12\textheight, keepaspectratio]{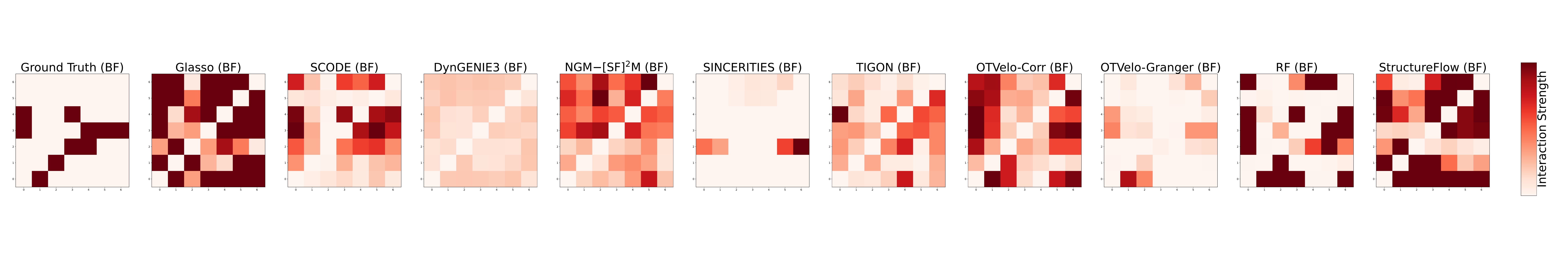}
    \caption{Dataset BF (Observational)}\label{fig:bf_obs}
  \end{subfigure}
  \vspace{2pt} %

  \begin{subfigure}{\textwidth}
    \centering
    \includegraphics[width=\textwidth, height=0.12\textheight, keepaspectratio]{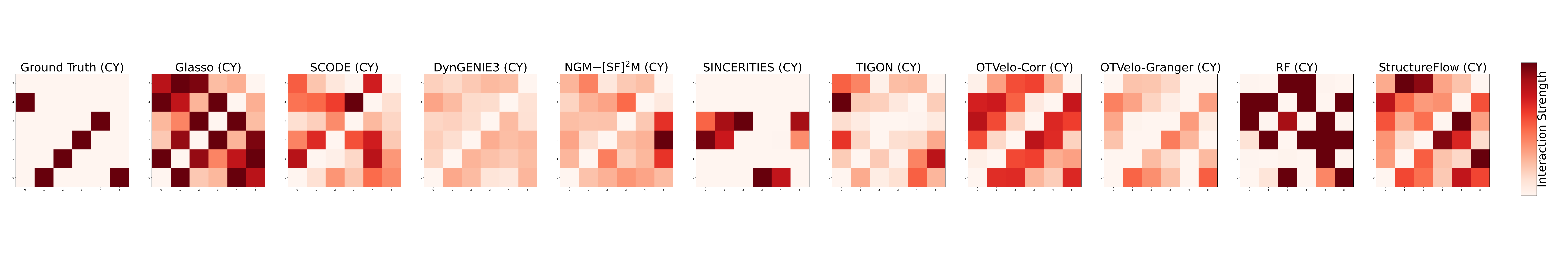}
    \caption{Dataset CY (Observational)}\label{fig:cy_obs}
  \end{subfigure}
  \vspace{2pt}

  \begin{subfigure}{\textwidth}
    \centering
    \includegraphics[width=\textwidth, height=0.12\textheight, keepaspectratio]{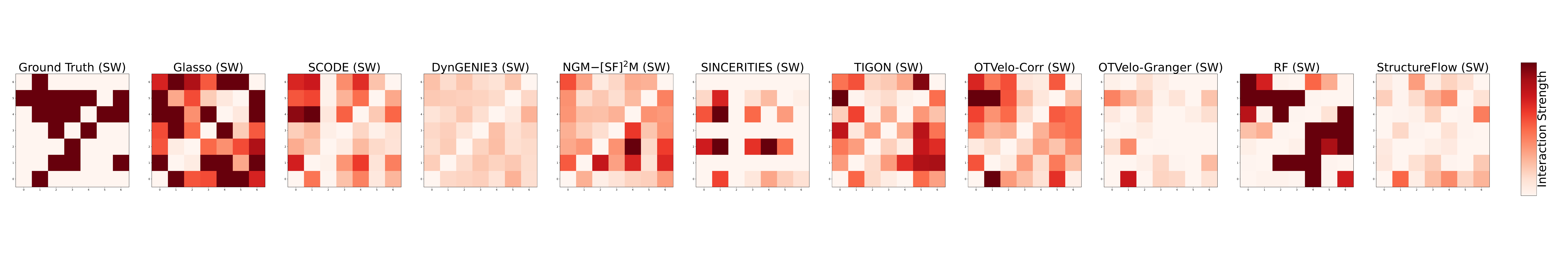}
    \caption{Dataset SW (Observational)}\label{fig:sw_obs}
  \end{subfigure}
  \vspace{2pt}

  \begin{subfigure}{\textwidth}
    \centering
    \includegraphics[width=\textwidth, height=0.12\textheight, keepaspectratio]{heatmaps/dyn-TF_wt_grid.pdf}
    \caption{Dataset TF (Observational)}\label{fig:tf_obs}
  \end{subfigure}
  \vspace{2pt}

  \begin{subfigure}{\textwidth}
    \centering
    \includegraphics[width=\textwidth, height=0.12\textheight, keepaspectratio]{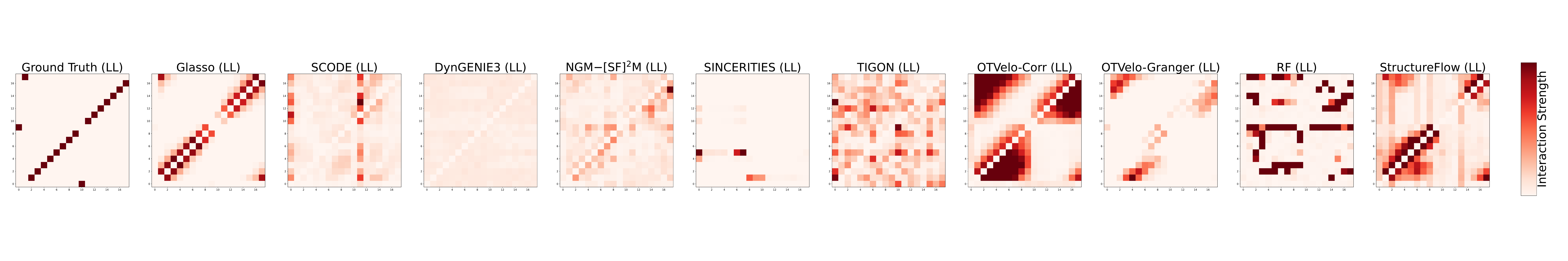}
    \caption{Dataset LL (Observational)}\label{fig:ll_obs}
  \end{subfigure}
  \vspace{2pt}

  \begin{subfigure}{\textwidth}
    \centering
    \includegraphics[width=\textwidth, height=0.12\textheight, keepaspectratio]{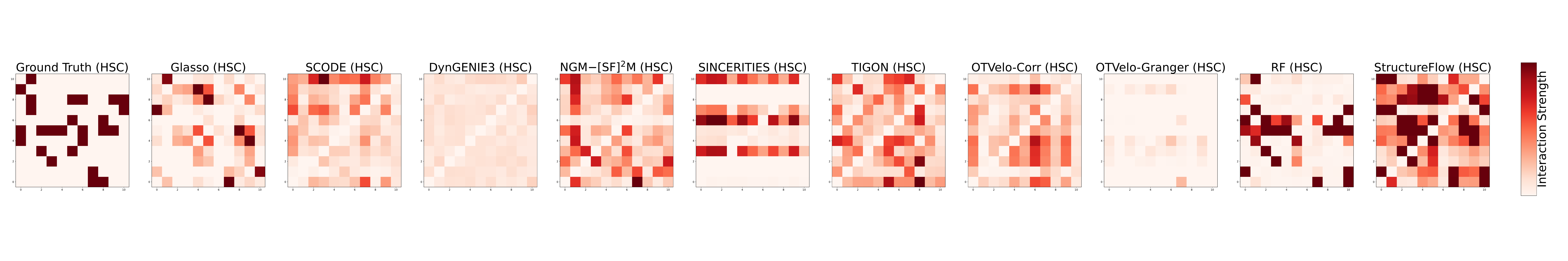}
    \caption{Dataset HSC (Observational)}\label{fig:hsc_obs}
  \end{subfigure}

  \caption{Six of the synthetic datasets with only observational information. Refer to panels (a)–(f) when discussing individual datasets.}
  \label{fig:all_heatmaps_obs}
\end{figure*}

\begin{figure*}[p]
  \centering
  \begin{subfigure}{\textwidth}
    \centering
    \includegraphics[width=\textwidth, height=0.12\textheight, keepaspectratio]{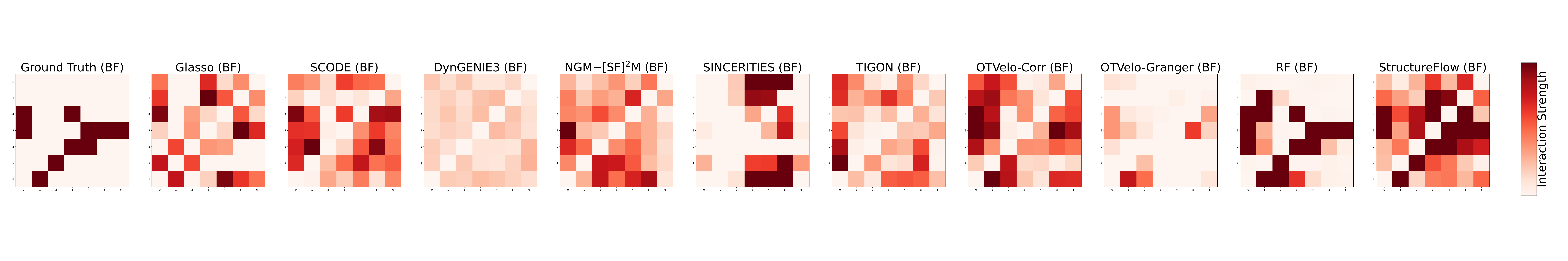}
    \caption{Dataset BF (Observational + Interventional)}\label{fig:bf_int}
  \end{subfigure}
  \vspace{2pt}

  \begin{subfigure}{\textwidth}
    \centering
    \includegraphics[width=\textwidth, height=0.12\textheight, keepaspectratio]{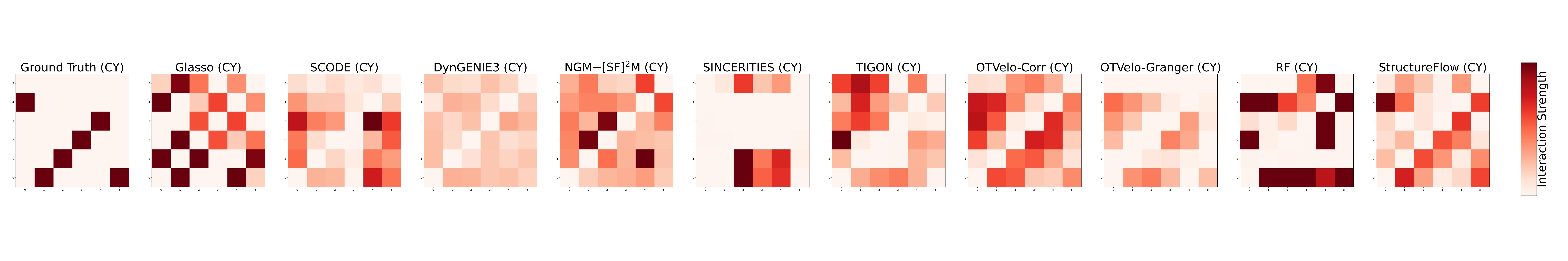}
    \caption{Dataset CY (Observational + Interventional)}\label{fig:cy_int}
  \end{subfigure}
  \vspace{2pt}

  \begin{subfigure}{\textwidth}
    \centering
    \includegraphics[width=\textwidth, height=0.12\textheight, keepaspectratio]{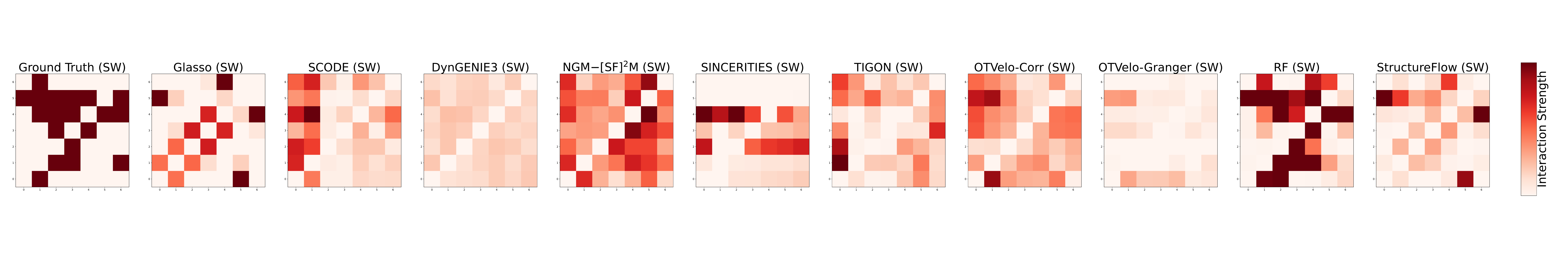}
    \caption{Dataset SW (Observational + Interventional)}\label{fig:sw_int}
  \end{subfigure}
  \vspace{2pt}

  \begin{subfigure}{\textwidth}
    \centering
    \includegraphics[width=\textwidth, height=0.12\textheight, keepaspectratio]{heatmaps/dyn-TF_full_grid.pdf}
    \caption{Dataset TF (Observational + Interventional)}\label{fig:tf_int}
  \end{subfigure}
  \vspace{2pt}

  \begin{subfigure}{\textwidth}
    \centering
    \includegraphics[width=\textwidth, height=0.12\textheight, keepaspectratio]{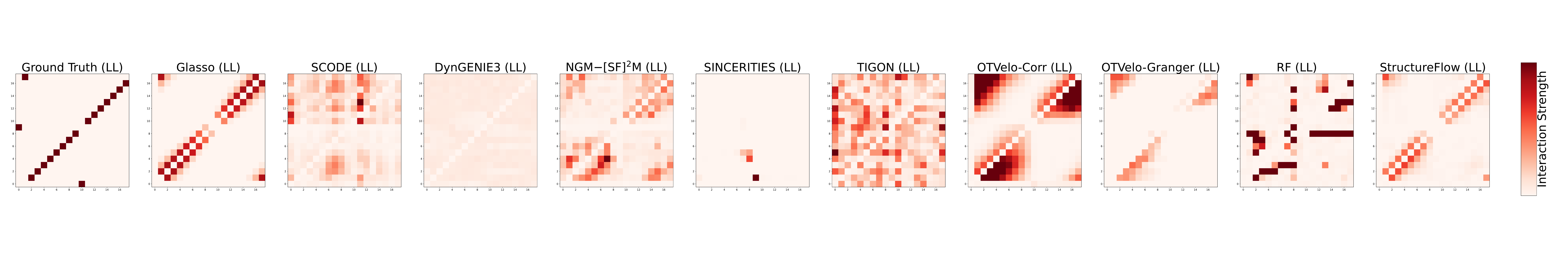}
    \caption{Dataset LL (Observational + Interventional)}\label{fig:ll_int}
  \end{subfigure}
  \vspace{2pt}

  \begin{subfigure}{\textwidth}
    \centering
    \includegraphics[width=\textwidth, height=0.12\textheight, keepaspectratio]{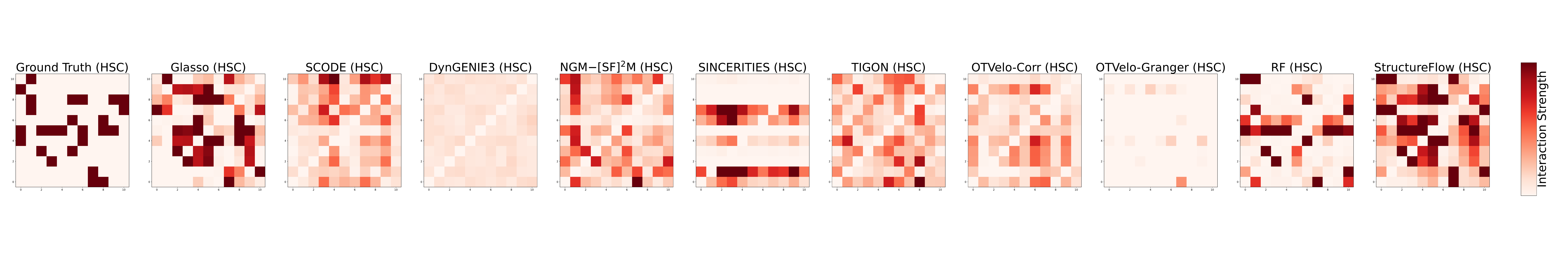}
    \caption{Dataset HSC (Observational + Interventional)}\label{fig:hsc_int}
  \end{subfigure}

  \caption{Six of the synthetic datasets with the addition of interventional information. Refer to panels (a)–(f) when discussing individual datasets.}
  \label{fig:all_heatmaps_int}
\end{figure*}

\subsection{Dynamical Inference of Conditional Population Dynamics on Synthetic Systems using \sfm}
\label{ap:syn_mlp_trajectory_inference}

\begin{figure}[htbp]
    \centering
    \includegraphics[width=0.8\textwidth]{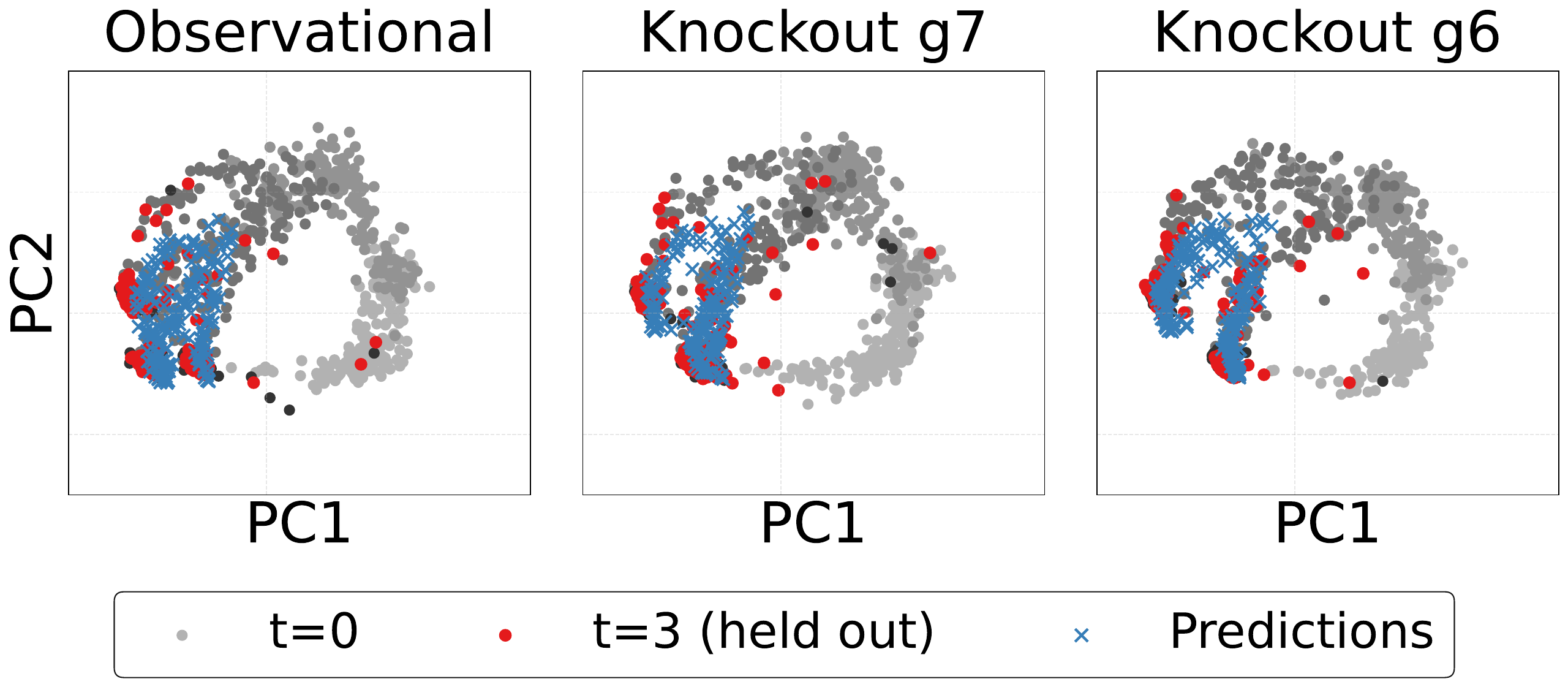}
    \caption{\textbf{$\mathrm{[SF]^2M}$ trajectory inference visualization.} Performance on observational (wild-type) data and two interventions selected for their diverse trifurcating paths. These examples illustrate the trajectory inference problem, where timepoint 3 was excluded during training, and the model simulates with the ODE from timepoint 2 to timepoint 3. Ground truth data is shown in grey, with progressively darker shades as time increases, except for timepoint 3 which is highlighted in red. Blue represent the model's predictions.}
    \label{fig:mlp_trajectory}
\end{figure}

We note in Figure~\ref{fig:mlp_trajectory} that $\text{[SF]}^2\text{M}$ predictions cluster around the synthetic systems' ground-truth trajectories across both observational and gene knockout settings. This is a result of training across both unconditional and conditional data.

\subsection{Prediction to Unseen Interventions (Knock-outs) on Synthetic Systems}
\label{ap:syn_kos_preds}

\begin{figure}[htbp]
    \centering
    \begin{subfigure}[b]{0.8\textwidth}
        \centering
        \includegraphics[width=\textwidth]{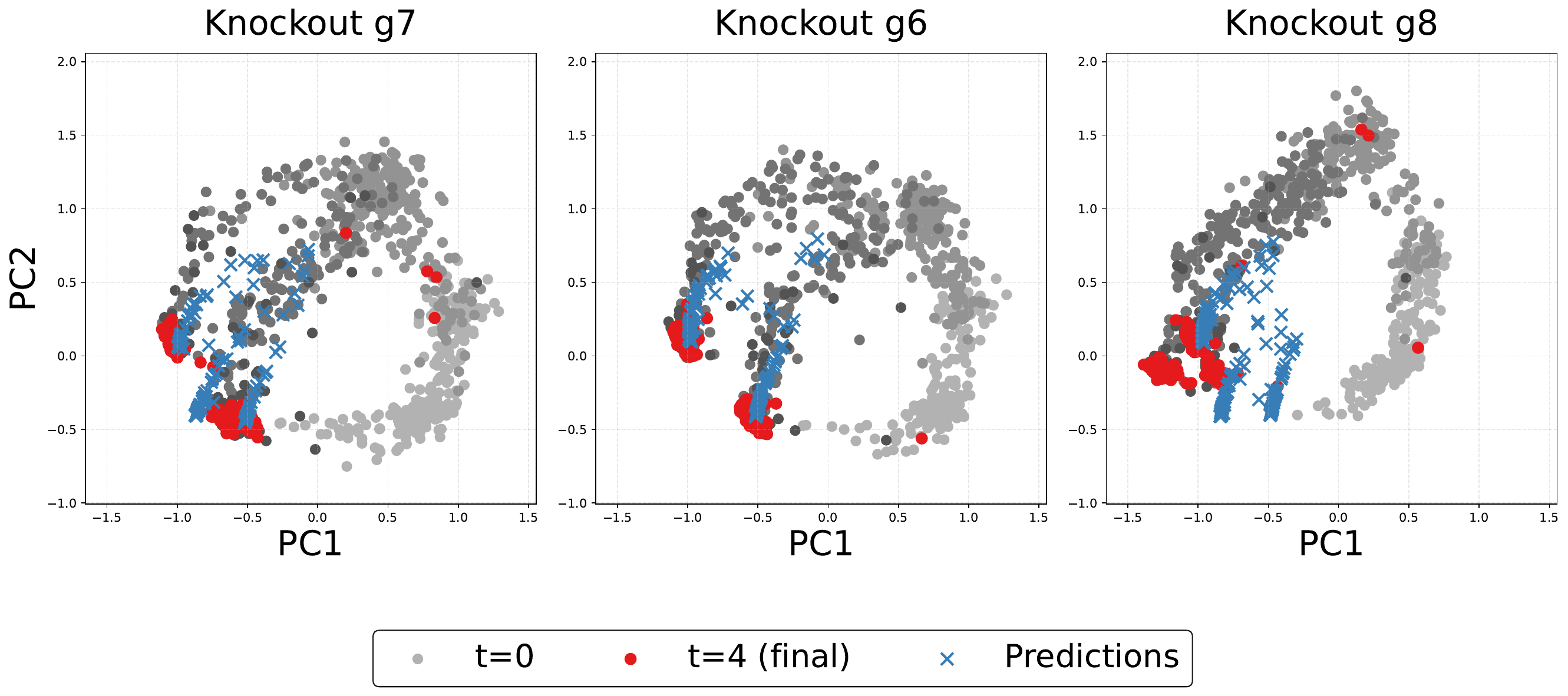}
        \caption{\textbf{$\mathrm{[SF]^2M}$ (no structure)}}
        \label{fig:mlp_leftout_kos}
    \end{subfigure}
    \hfill
    \begin{subfigure}[b]{0.8\textwidth}
        \centering
        \includegraphics[width=\textwidth]{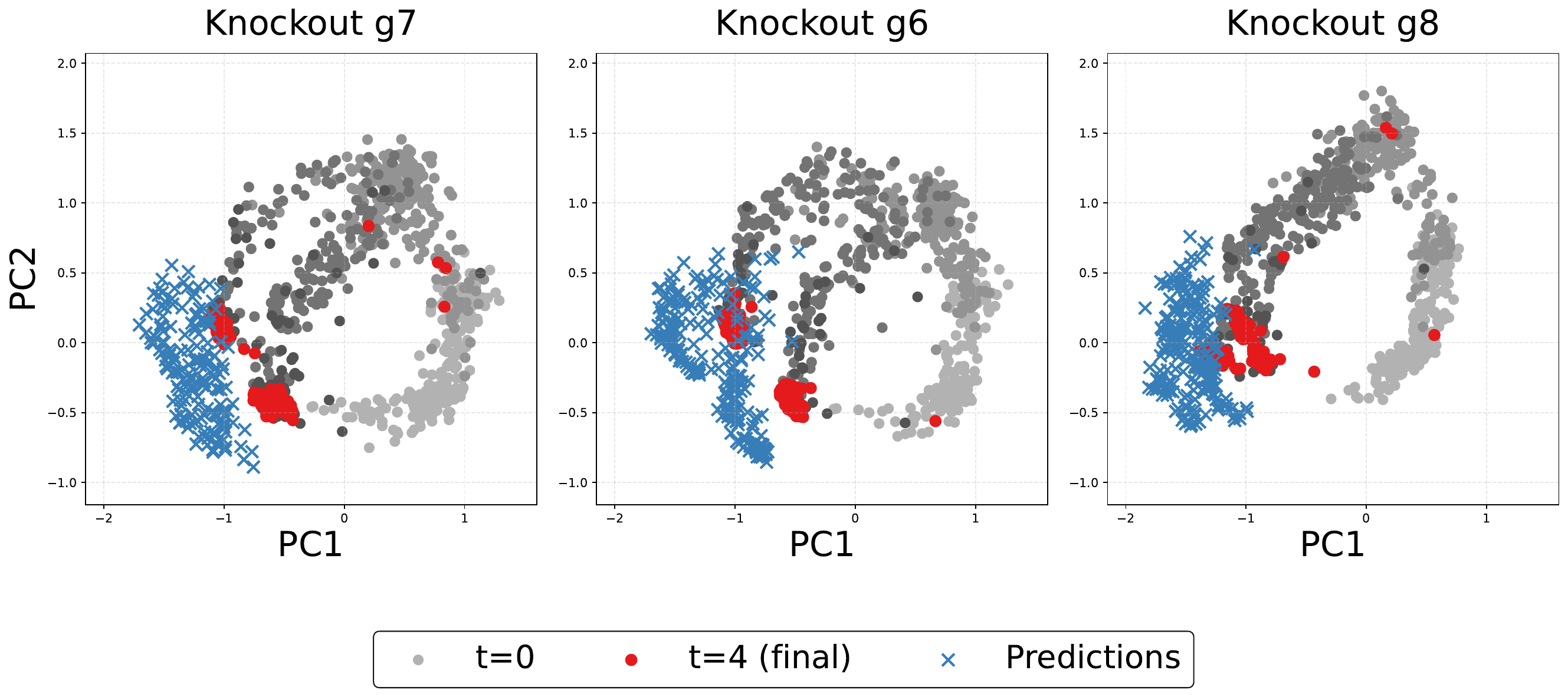}
        \caption{\textbf{RF}}
        \label{fig:rf_leftout_kos}
    \end{subfigure}
    \hfill
    \begin{subfigure}[b]{0.8\textwidth}
        \centering
        \includegraphics[width=\textwidth]{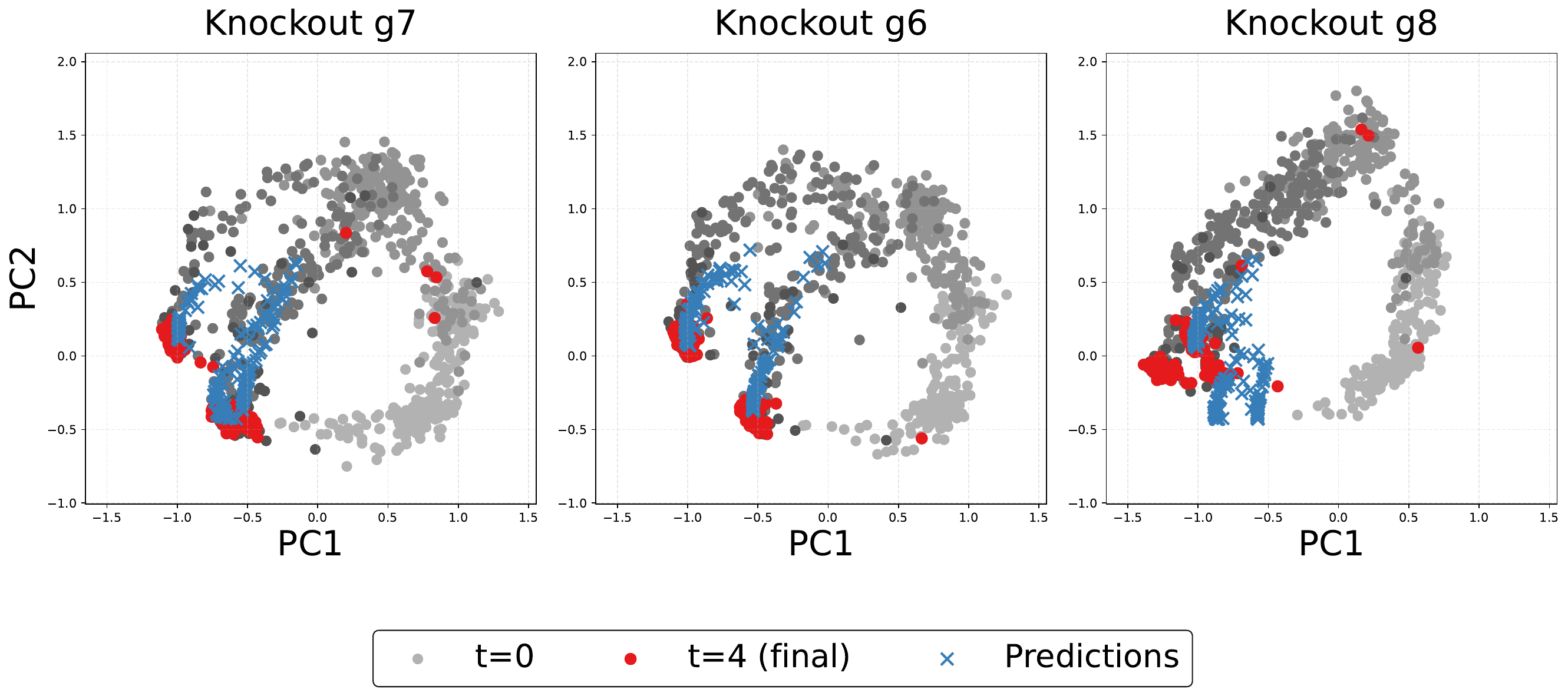}
        \caption{\textbf{\methodname (ours)}}
        \label{fig:structureflow_leftout_kos}
    \end{subfigure}
    
    \caption{\textbf{Comparison of trajectory inference performance for left-out interventions (knock-outs).} We show PCA visualizations of model predictions for interventions that were excluded during training. Ground truth data is shown in grey, with model predictions overlaid. This demonstrates each method's ability to generalize to unseen perturbational conditions.}
    \label{fig:leftout_kos_comparison}
\end{figure}

Table \ref{ap:knockout_results} shows performance of \methodname and baselines on the leave-one-knockout-out trajectory inference task.  {\methodname improves over RF on $W_2$ for the TF and CY systems, while RF remains stronger on several MMD comparisons and the Curated (HSC) system.}

\begin{table}[th]
\centering
\caption{\textbf{Comparison of left-out intervention (knock-out) prediction on simulated biological systems.} For each dataset, we report Wasserstein-2 ($W_2$) and MMD metrics. Lower is better. 
}
\label{ap:knockout_results}
\resizebox{\columnwidth}{!}{
\begin{tabular}{@{}lcccccccccccc@{}}
\toprule
 & \multicolumn{2}{c}{TF} & \multicolumn{2}{c}{CY} & \multicolumn{2}{c}{LL} & \multicolumn{2}{c}{HSC} & \multicolumn{2}{c}{BF} & \multicolumn{2}{c}{SW} \\
\cmidrule(lr){2-3}
\cmidrule(lr){4-5}
\cmidrule(lr){6-7}
\cmidrule(lr){8-9}
\cmidrule(lr){10-11}
\cmidrule(lr){12-13}
 & $W_2$↓ & MMD↓ & $W_2$↓ & MMD↓ & $W_2$↓ & MMD↓ & $W_2$↓ & MMD↓ & $W_2$↓ & MMD↓ & $W_2$↓ & MMD↓ \\
\midrule
\sfm & 0.964 $\pm$ 0.016 & 0.103 $\pm$ 0.003 & 0.967 $\pm$ 0.028 & 0.277 $\pm$ 0.017 & 1.082 $\pm$ 0.095 & 0.157 $\pm$ 0.010 & 0.886 $\pm$ 0.011 & 0.104 $\pm$ 0.003 & 1.138 $\pm$ 0.031 & 0.220 $\pm$ 0.005 & 1.404 $\pm$ 0.051 & 0.358 $\pm$ 0.015 \\

\midrule
RF & 1.062 $\pm$ 0.000 & 0.098 $\pm$ 2.8e-8 & 1.889 $\pm$ 0.000 & 0.329 $\pm$ 1.5e-8 & 1.887 $\pm$ 0.005 & 0.232 $\pm$ 6.5e-4 & 0.984 $\pm$ 3.4e-10 & 0.065 $\pm$ 1.7e-8 & 1.176 $\pm$ 0.000 & 0.134 $\pm$ 1.1e-8 & 1.285 $\pm$ 0.000 & 0.284 $\pm$ 3.4e-8 \\
\methodname & 1.012 $\pm$ 0.010 & 0.110 $\pm$ 4.1e-3 & 0.924 $\pm$ 0.036 & 0.250 $\pm$ 1.8e-2 & 1.123 $\pm$ 0.037 & 0.141 $\pm$ 6.1e-3 & 0.852 $\pm$ 0.011 & 0.120 $\pm$ 4.2e-3 & 1.079 $\pm$ 0.016 & 0.213 $\pm$ 7.1e-3 & 1.159 $\pm$ 0.032 & 0.330 $\pm$ 1.1e-2 \\
\bottomrule
\label{tab:unseen_ko_boolode}
\end{tabular}
}
\end{table}

\subsection{OT-CFM Ablation on Synthetic (Trifurcating) System}
\label{ap:ot_cfm_ablation}

In Table \ref{tab:otcfm_ablation} we remove the stochastic noise term from the conditional flow matching objective, reducing the formulation to an ODE that models the drift component. This corresponds to the classical flow matching objective \cite{lipman2023fm}, where the dynamics align deterministic trajectories. \methodname is trained to model the process as a Schr\"odinger Bridge, where the forward-backward SDE explicitly models both drift and diffusion. From the results, we observe that discarding the stochastic component (OT-CFM) substantially degrades both classification (AUROC/AUPR) and distributional alignment metrics ($W_2$), highlighting that noise modeling contributes meaningfully to the structure learning task. For dynamical systems such as cellular trajectories observed across time points, modeling the process as a Schr\"odinger Bridge rather than an ODE provides a more accurate representation of developmental variability and uncertainty.

\begin{table}[H]
\centering
\caption{\textbf{Comparison of \methodname with and without stochastic component on the TF synthetic system.} Results show AUROC, AUPR, and distributional distance metrics ($W_2$, MMD).
}
\label{tab:otcfm_ablation}
\begin{tabular}{@{}lcccccc@{}}
\toprule
 & AUROC ↑ & AUPR ↑ & $W_2$ ↓ & MMD ↓ \\
\midrule
\textbf{\methodname} & \textbf{0.968 $\pm$ 0.005} & \textbf{0.932 $\pm$ 0.011} & \textbf{0.795 $\pm$ 0.007} & 0.136 $\pm$ 0.002 \\
\textbf{\methodname (OT-CFM)} & 0.791 $\pm$ 0.013 & 0.634 $\pm$ 0.037 & 0.846 $\pm$ 0.022 & \textbf{0.062 $\pm$ 0.004} \\
\bottomrule
\end{tabular}
\end{table}

\subsection{Scaling and runtime of various baselines and our method}
\label{ap:structureflow_scaling}
\rev{Here we provide an additional experiment comparing training time of all methods on the BoolODE \textbf{TF} system. For this experiment, all methods are run on the same device (GPU, Quadro RTX 6000), but we note \methodname can be efficiently trained using a CPU. 
We only report parameter counts for neural network-based methods.
For \methodname and the flow matching baseline, we use ($\cdot$) to denote the additional time used to compute EOT. We use the same number of timepoints (5), dimensionality (8), and number of cells (8000).  We add flow matching as a reference point to \methodname, where we remove regularization terms, the score model, and replace the NGM backbone with a traditional MLP backbone. 
}
 {The comparison of wall-clock timing, together with the synthetic scaling study (Figure~\ref{fig:linear_scaling}) and the Renge runtimes (Table~\ref{tab:runtime_renge}), use a common data setup and were run on a standard desktop workstation.
  For baselines, hyperparameters were set to author-recommended default values.
  We report parameter counts as a proxy for memory footprint, with \methodname being small and training on CPU within 32\,GB of RAM.
}
\begin{table}[H]
\centering
\caption{\rev{Model wall-clock time on the Trifurcating (\textbf{TF}) BoolODE system. We include parameter counts for all neural network-based methods. We use ($\cdot$) to denote runtime of EOT.}}
\label{tab:runtime_boolode}
    \begin{tabular}{@{}lcl@{}}
    \toprule
    Time [s] & Model & Parameter Size \\
    \midrule
    263 + (3.3) & \methodname & 20716 \\
    1511.18 & TIGON & 43140 \\
    82.01 & RF & - \\
    3228.61 & NGM-NODE & 8008 \\
    116.17 + (3.3) & Flow matching & 8008 \\
    208.58 + (3.3) & $\text{[SF]}^2\text{M}$ & 14892 \\
    131.22 & OTVelo & - \\
    5771.41 & SDFL & - \\
    \bottomrule
    \end{tabular}
\end{table}

\subsection{Additional Ablations of Hyperparameters}
\label{ap:Structureflow_ablations}
\rev{In order to assess the stability of our method we vary three hyperparameters that are crucial to the performance of our model: $\alpha$, which varies the weight of the score and flow term; group lasso regularization which determines how sparse the adjacency matrix recovered by the autonomous flow term is; and the size of the hidden dimensions in our flow module (We use one layer in our case).}
\begin{table}[h]
    \centering
    \caption{\rev{Ablation over the group lasso regularization term.}}
    \label{tab:group_lasso} %
    \resizebox{\columnwidth}{!}{%
    \begin{tabular}{lcccccccc}
        \toprule
        group lasso Reg. & 0.0 & 0.0001 & 0.001 & 0.01 & 0.04 & 0.1 & 0.2 & 0.5 \\
        \midrule
        AUROC & $0.943\pm0.004$ & $0.946\pm0.004$ & $0.944\pm0.004$ & $0.952\pm0.002$ & \textbf{0.971$\pm$0.008} & $0.853\pm0.032$ & $0.851\pm0.017$ & $0.879\pm0.016$ \\
        AP & $0.804\pm0.014$ & $0.816\pm0.015$ & $0.805\pm0.014$ & $0.833\pm0.008$ & \textbf{0.918$\pm$0.026} & $0.677\pm0.032$ & $0.688\pm0.015$ & $0.708\pm0.021$ \\
        $W_2$ & $0.7623\pm0.0047$ & $0.7623\pm0.0030$ & \textbf{0.7608$\pm$0.0050} & $0.7665\pm0.0077$ & $0.7886\pm0.0048$ & $0.8329\pm0.0049$ & $0.8398\pm0.0059$ & $0.8414\pm0.0058$ \\
        \bottomrule
    \end{tabular}%
    }
\end{table}

\begin{table}[h]
    \centering
    \caption{\rev{Ablation over $\alpha$. $\alpha = 1$ pertains to only using the score model. $\alpha = 0$ pertains to only using flow model.}}
    \label{tab:alpha_ablation}
    \begin{tabular}{lcccc}
        \toprule
        $\alpha$ & 0.1 & 0.2 & 0.5 & 0.8 \\
        \midrule
        AUROC & \textbf{0.971$\pm$0.008} & $0.965\pm0.007$ & \textbf{0.971$\pm$0.009} & $0.937\pm0.005$ \\
        AP & $0.917\pm0.025$ & $0.886\pm0.026$ & \textbf{0.918$\pm$0.040} & $0.851\pm0.011$ \\
        $W_2$ & 0.7886$\pm$0.0048 & \textbf{0.7814$\pm$0.0007} & 0.7887$\pm$0.0035& 0.8085$\pm$0.0042 \\
        \bottomrule
    \end{tabular}
\end{table}

\begin{table}[h]
    \centering
    \caption{\rev{Ablation over size of hidden dimension (\# of neurons) per layer for the NGM flow model.}}
    \label{tab:hidden_dim_ablation}
    \resizebox{\columnwidth}{!}{
    \begin{tabular}{lccccc}
        \toprule
        Hidden dimension for NGM & 10 & 50 & 100 & 200 & 256 \\
        \midrule
        AUROC & $0.925\pm0.010$ & $0.964\pm0.004$ & $0.971\pm0.008$ & $0.969\pm0.003$ & \textbf{0.975$\pm$0.003} \\
        AP & $0.832\pm0.020$ & $0.910\pm0.013$ & $0.917\pm0.025$ & $0.916\pm0.014$ & \textbf{0.932$\pm$0.012} \\
        $W_2$ & 0.7950$\pm$0.0024 & \textbf{0.7853$\pm$0.0026} & 0.7886$\pm$0.0048 & 0.7972$\pm$0.0055 & 0.7858$\pm$0.0208 \\
        \bottomrule
    \end{tabular}
    }
\end{table}

\subsection{Assessing the stability of the score and flow network in \methodname}
\label{ap:structure_stable}
\rev{Here we test two settings where we try freezing the weights of one of the modules. We jointly train $s_\phi$ and $v_\theta$ for $K \in \{10000,15000\}$ iterations, then freeze the parameters of $s_\phi$ and continue training only $v_\theta$ for an additional $M = 5000$ iterations. We do the same experiment but freeze the flow term instead. In both settings our performance stays constant for structure learning, but is more sensitive for dynamics.}
\begin{table}[!th]
    \centering
    \caption{\rev{Ablation over freezing flow and score model post-training when models pre-trained for $10000$ iterations.}}
    \begin{tabular}{lccc}
        \toprule
        Freeze model & AP & AUROC & $W_2$ \\
        \midrule
        score & 0.9011 & 0.9681 & $1.5462 \pm 0.0520$ \\
        flow  & 0.9109 & 0.9700 & $0.9366 \pm 0.1992$ \\
        \bottomrule
    \end{tabular}
\end{table}

\begin{table}[!th]
    \centering
    \caption{\rev{Ablation over freezing flow and score model post-training when models pre-trained for $15000$ iterations.}}
    \begin{tabular}{lccc}
        \toprule
        Freeze model & AP & AUROC & $W_2$ \\
        \midrule
        score & 0.9019 & 0.9668 & $1.5200 \pm 0.0569$ \\
        flow  & 0.8843 & 0.9650 & $0.9337 \pm 0.2181$ \\
        \bottomrule
    \end{tabular}
\end{table}

\subsection{Scaling and runtime of various baselines and our method on Renge data}
\label{ap:structureflow_scaling_real}
\begin{table}[H]
\centering
\caption{\rev{Model wall-clock time on the Real (Renge) dataset. We include parameter counts for all neural network-based methods. We note that for \methodname and $[SF]^2M$ the OT pairing was negligble and thus not included.}}
\label{tab:runtime_renge}
    \begin{tabular}{@{}lcl@{}}
    \toprule
    Time [s] & Model & Parameter Size \\
    \midrule
    319.31 $\pm$ 21.29 & \methodname & 1440646 \\
    1542.88 $\pm$ 26.67 & TIGON (PCA) & 43140 \\
    14112.49 & TIGON (Ambient) & 409448 \\
    338.12 $\pm$ 0.78 & RF & - \\
    101.31 $\pm$ 1.43 & $\text{[SF]}^2\text{M}$ & 175310  \\
    12.21 $\pm$ .125 & OTVelo & - \\
    \bottomrule
    \end{tabular}
\end{table}

\subsection{Structure Experiments for Real Data}
\label{ap:real_data_experiments}
\begin{figure*}[t]
    \centering
    \begin{subfigure}[t]{0.42\textwidth}
        \centering
        \includegraphics[width=\linewidth]{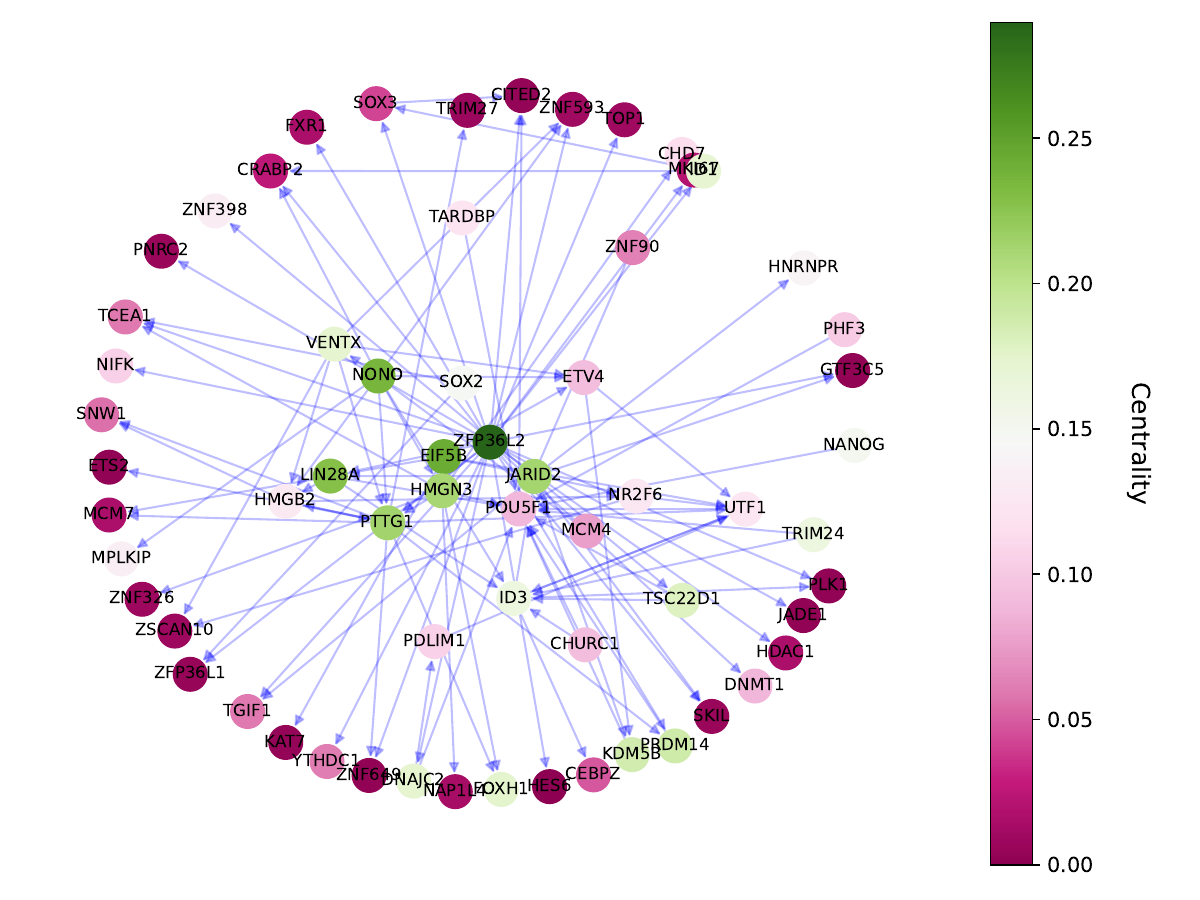}
        \caption{\methodname inferred regulatory network graph.}
        \label{fig:structflow-graph}
    \end{subfigure}
    \hfill
    \begin{subfigure}[t]{0.56\textwidth}
        \centering
        \includegraphics[width=\linewidth]{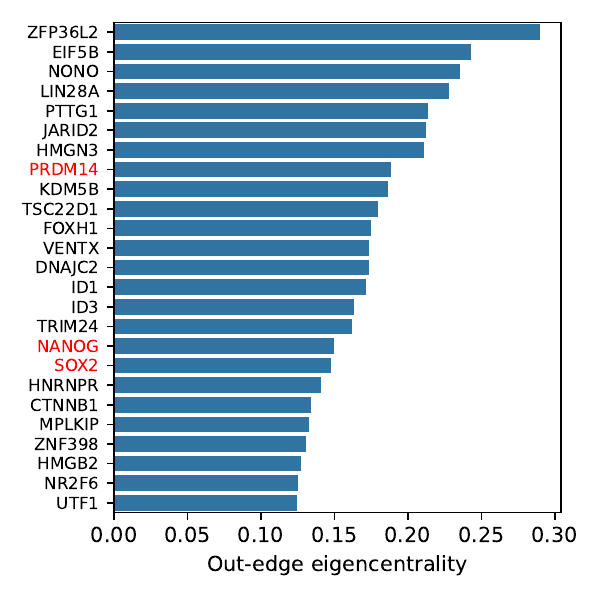}
        \caption{Out-edge eigencentrality scores for the \methodname-inferred network.}
        \label{fig:structflow-centrality}
    \end{subfigure}
    
    \vspace{1em} %

    \begin{subfigure}[b]{0.32\textwidth}
        \centering
        \includegraphics[width=\linewidth]{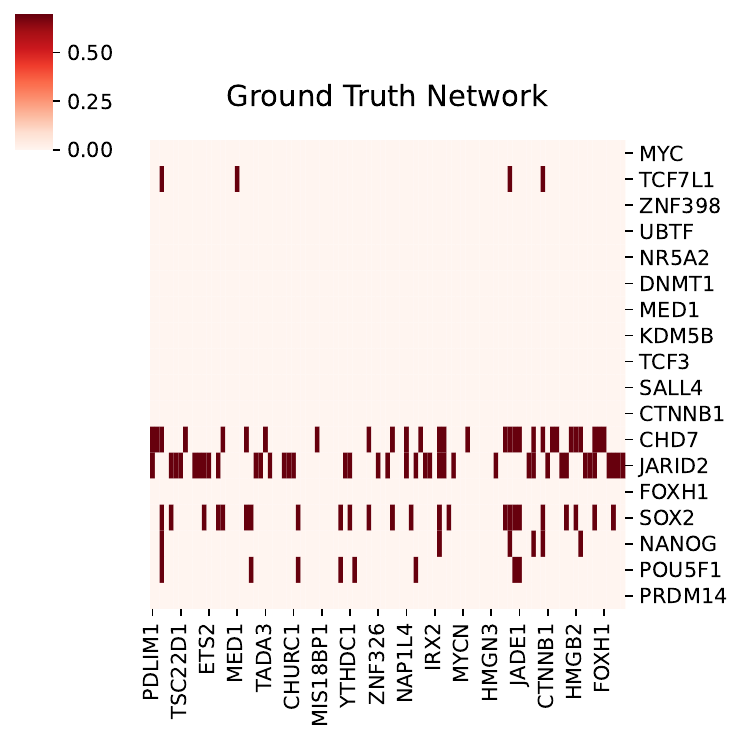}
        \caption{Ground truth regulatory network (heatmap).}
        \label{fig:gt-heatmap}
    \end{subfigure}
    \hfill
    \begin{subfigure}[b]{0.32\textwidth}
        \centering
        \includegraphics[width=\linewidth]{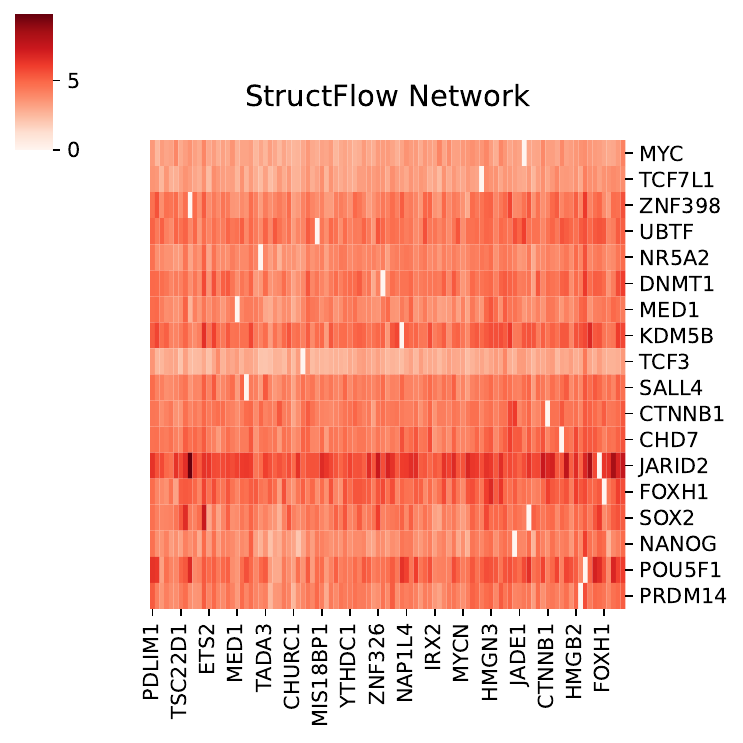}
        \caption{\methodname-inferred network (no threshold).}
        \label{fig:structflow-heatmap}
    \end{subfigure}
    \hfill
    \begin{subfigure}[b]{0.32\textwidth}
        \centering
        \includegraphics[width=\linewidth]{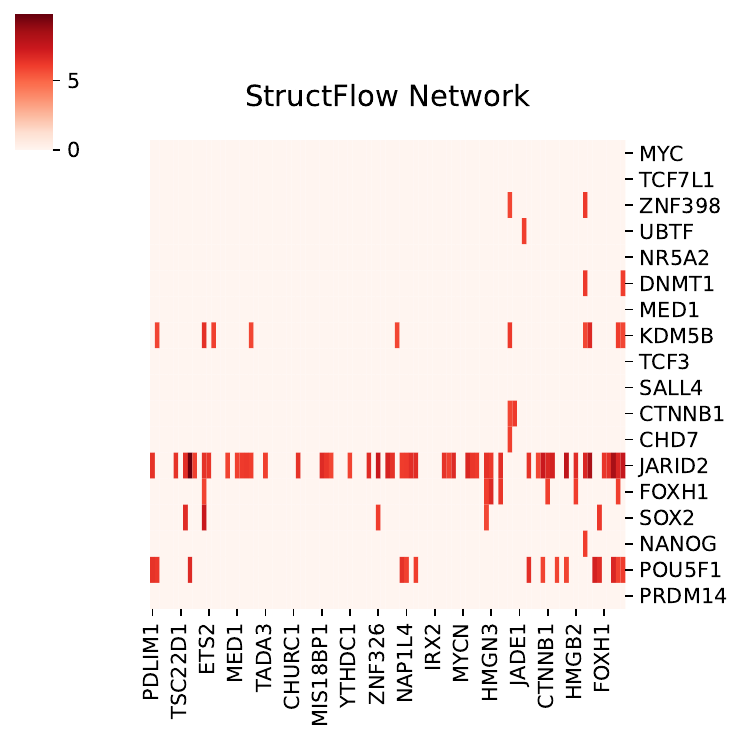}
        \caption{\methodname-inferred network (99th quantile threshold).}
        \label{fig:structflow-heatmap-thresh}
    \end{subfigure}
    
    \vspace{0.5em}
    \caption{
        \textbf{Analysis of gene regulatory network inference results.}
        \textbf{(a)} \methodname graph representation of the inferred gene regulatory network.
        \textbf{(b)} The top 25 Centrality scores (out-edge eigencentrality) for nodes in the \methodname-inferred network.
        \textbf{(c)} Ground truth regulatory network shown as a heatmap. This network is from the experimental binding information from the ChIP-atlas.
        \textbf{(d)} \methodname-inferred network adjacency matrix (no thresholding).
        \textbf{(e)} \methodname-inferred network thresholded at the 99th quantile to highlight the strongest edges.
    }
    \label{fig:structflow-summary}
    \vspace{-10pt}
\end{figure*}

\clearpage

\subsection{\rev{Evaluating StructureFlow with Irregularly Sampled Time-Points}}
\label{ap:irregular_timepoints}

\rev{We evaluate \methodname and Reference Fitting on the TF BoolODE system using population snapshots from non-uniform timepoints. To accomplish this, we bin the continuous timepoint data into randomly non-uniformly spaced time bins with a standard deviation of 30\% of the time bin width (whereas our original experiments bin into discrete equally spaced timebins of $t=1,2,3,\ldots$). We run this experiment over 3 seeds, where the seeding randomizes the time bin spacing (making the data non-deterministic between runs, and thus RF is non-deterministic in this specific setting).}

\begin{table}[H]
\centering
\caption{\rev{Comparison of \methodname and RF on irregularly sampled time-points for the TF system.}}
\label{tab:irregular_timepoints}
\begin{tabular}{@{}lcc@{}}
\toprule
Metric & \methodname & RF \\
\midrule
GRN Graph AP & $0.8501 \pm 0.0348$ & $0.9614 \pm 0.0282$ \\
GRN Graph AUROC & $0.9326 \pm 0.0180$ & $0.9812 \pm 0.0099$ \\
Trajectory $W_2$ & $0.8491 \pm 0.0168$ & $1.4060 \pm 0.0665$ \\
Trajectory MMD & $0.1408 \pm 0.0022$ & $0.1977 \pm 0.0152$ \\
\bottomrule
\end{tabular}
\end{table}

\subsection{\rev{Comparison with Symbolic Distribution Flow Learning (SDFL)}}
\label{ap:sdfl_comparison}

\rev{We compare \methodname to Symbolic Distribution Flow Learning (SDFL) \citep{dakhmouche2025robust}, which tackles the population dynamics problem through learning explicit symbolic equations. While SDFL provides interpretable outputs through per-dimension equations, it requires significant computation time to run and cannot explicitly recover a learned matrix structure for the structure learning task. We evaluate SDFL on the trifurcating (TF) system for the trajectory inference task and report results below (we include \methodname results from Table \ref{tab:timepoint_results} for easy comparison). Note due to the method's training instability, we were unable to get 3 consecutive successful seeds, and therefore report only 1 seed result.}

\begin{table*}[t]
\centering
\caption{\textbf{\methodname  {achieves the best $W_2$ among joint methods} for \textit{dynamical inference} on simulated biological systems.} 
Shown is a comparison of dynamical inference methods for learning conditional population dynamics across synthetic biological systems. 
We include \sfm for comparison, but note that \sfm does not infer underlying network structure (i.e. cannot address joint task).
Colors indicate \best{1$^{\text{st}}$ best performer} and \second{2$^{\text{nd}}$ best performer} for models that perform the joint inference task.
}
\vspace{0.3em}
\label{tab:timepoint_results_full}
\resizebox{\textwidth}{!}{
\begin{tabular}{@{}lcccccccccccc@{}}
\toprule
 & \multicolumn{2}{c}{\textbf{TF}} & \multicolumn{2}{c}{\textbf{CY}} & \multicolumn{2}{c}{\textbf{LL}} & \multicolumn{2}{c}{ {\textbf{SW}}} & \multicolumn{2}{c}{\textbf{BF}} & \multicolumn{2}{c}{ {\textbf{HSC (Curated)}}} \\
\cmidrule(lr){2-3} \cmidrule(lr){4-5} \cmidrule(lr){6-7} \cmidrule(lr){8-9} \cmidrule(lr){10-11} \cmidrule(lr){12-13}
 & $W_2$↓ & MMD↓ & $W_2$↓ & MMD↓ & $W_2$↓ & MMD↓ & $W_2$↓ & MMD↓ & $W_2$↓ & MMD↓ & $W_2$↓ & MMD↓ \\
\midrule
\sfm & 0.761 $\pm$ 0.014 & 0.134 $\pm$ 0.005 & 0.517 $\pm$ 0.015 & 0.153 $\pm$ 0.002 & 0.847 $\pm$ 0.053 & 0.190 $\pm$ 0.010 &  {0.572 $\pm$ 0.008} &  {0.215 $\pm$ 0.005} & 0.660 $\pm$ 0.007 & 0.140 $\pm$ 0.004 &  {0.664 $\pm$ 0.011} &  {0.083 $\pm$ 0.003} \\
\midrule
RF & \second{1.376 $\pm$ 0.000} & 0.191 $\pm$ 0.000 & 2.086 $\pm$ 0.000 & 0.326 $\pm$ 0.000 & 2.120 $\pm$ 0.007 & 0.296 $\pm$ 0.000 &  {1.371 $\pm$ 0.000} &  {0.290 $\pm$ 0.000} & 1.495 $\pm$ 0.000 & 0.242 $\pm$ 0.000 &  {0.950 $\pm$ 0.000} &  {0.066 $\pm$ 0.000} \\
OTVelo & 1.874 $\pm$ 0.262 & 0.582 $\pm$ 0.039 & 1.972 $\pm$ 0.144 & 0.622 $\pm$ 0.034 & 2.878 $\pm$ 0.361 & 0.647 $\pm$ 0.015 &  {1.940 $\pm$ 0.144} &  {0.577 $\pm$ 0.022} & 1.847 $\pm$ 0.277 & 0.591 $\pm$ 0.042 &  {1.942 $\pm$ 0.006} &  {0.603 $\pm$ 0.002} \\
TIGON & 1.409 $\pm$ 0.294 & \best{0.063 $\pm$ 0.022} & \second{1.224 $\pm$ 0.053} & \best{0.044 $\pm$ 0.003}  & \second{1.840 $\pm$ 0.293} & \best{0.066 $\pm$ 0.039} & \second{0.957 $\pm$ 0.190} & \best{0.017 $\pm$ 0.009} & \second{1.394 $\pm$ 0.294} & \best{0.066 $\pm$ 0.021} & \best{0.666 $\pm$ 0.363} & \best{0.012 $\pm$ 0.014} \\
\methodname & \best{0.789 $\pm$ 0.012} & \second{0.135 $\pm$ 0.004} & \best{0.577 $\pm$ 0.012} & \second{0.140 $\pm$ 0.004} & \best{0.842 $\pm$ 0.031} & \second{0.196 $\pm$ 0.008} &  {\best{0.610 $\pm$ 0.010}} &  {\second{0.220 $\pm$ 0.006}} & \best{0.694 $\pm$ 0.017} & \second{0.142 $\pm$ 0.007} &  {\best{0.683 $\pm$ 0.011}} &  {\second{0.068 $\pm$ 0.003}} \\
\bottomrule
\end{tabular}
}
\vspace{-10pt}
\end{table*}

\begin{table}[H]
\centering
\caption{\rev{Comparison of dynamical inference performance with SDFL on the TF system.}}
\label{tab:sdfl_comparison}
\begin{tabular}{@{}lcc@{}}
\toprule
Method & $W_2$↓ & MMD↓ \\
\midrule
SDFL & $3.4880 $ & $0.0785 $ \\
\methodname & $0.789 \pm 0.012$ & $0.135 \pm 0.004$ \\
\bottomrule
\end{tabular}
\end{table}

\subsection{\rev{Full Results for Renge (real) Data Results}}
\label{ap:renge_timepoint_full_results}
We provide the full results across held-out timepoints 1 and 2 for the \textit{dynamical inference} of left-out marginals on the real (Renge) biological dataset.

\begin{table}[H]
\centering
\caption{\textbf{\methodname shows competitive performance for \textit{dynamical inference} of left-out marginals on the real (Renge) biological dataset.}
We show a comparison of dynamical inference methods for learning conditional population dynamics across left-out timepoints $k=1,2$ on the Renge dataset.
We report results over withheld $k$ as well as the average of the two using Wasserstein-2 ($W_2$↓) and Energy Distance (ED↓) \citep{rizzo2016energy} metrics.
Bold indicates \best{1$^{\text{st}}$ best performer} and the underline indicates \second{2$^{\text{nd}}$ best performer} for models that perform the joint inference task.
}

\vspace{0.3em}
\label{tab:renge_timepoint_results_full_precision}
\resizebox{\columnwidth}{!}{
\begin{tabular}{@{}l cccccc@{}}
\toprule
 & \multicolumn{2}{c}{Timepoint 1} & \multicolumn{2}{c}{Timepoint 2} & \multicolumn{2}{c}{Average} \\
\cmidrule(lr){2-3} \cmidrule(lr){4-5} \cmidrule(lr){6-7}
 & $W_2$↓ & ED↓ & $W_2$↓ & ED↓ & $W_2$↓ & ED↓ \\
\midrule
\sfm & 5.673 $\pm$ 0.034 & 0.533 $\pm$ 0.024 & 5.841 $\pm$ 0.030 & 0.524 $\pm$ 0.024 & 5.757 $\pm$ 0.032 & 0.528 $\pm$ 0.024 \\
\midrule
RF & 10.193 $\pm$ 0.004 & 12.389 $\pm$ 0.009 & 9.242 $\pm$ 0.004 & 10.089 $\pm$ 0.008 & 9.717 $\pm$ 0.004 & 11.239 $\pm$ 0.008 \\
OTVelo & 7.106 $\pm$ 0.000 & 4.269 $\pm$ 0.000 & 6.996 $\pm$ 0.000 & 3.658 $\pm$ 0.000 & 7.051 $\pm$ 0.000 & 3.963 $\pm$ 0.000 \\
TIGON & \second{6.401 $\pm$ 0.033} & \best{0.173 $\pm$ 0.011} & \second{6.402 $\pm$ 0.002} & \best{0.125 $\pm$ 0.001} & \second{6.402 $\pm$ 0.001} & \best{0.149 $\pm$ 0.024} \\
\methodname & \best{5.589 $\pm$ 0.026} & \second{0.692 $\pm$ 0.048} & \best{5.679 $\pm$ 0.029} & \second{0.765 $\pm$ 0.025} & \best{5.634 $\pm$ 0.027} & \second{0.729 $\pm$ 0.037} \\
\bottomrule
\end{tabular}
}
\end{table}

\begin{table}[H]
\centering
\caption{\textbf{\methodname shows competitive performance for \textit{dynamical inference} of left-out interventions on the real (Renge) biological dataset.}
We show a comparison of dynamical inference methods for learning conditional population dynamics on the Renge dataset for left-out interventions.
We report results for 3 left-out gene knockout conditions using Wasserstein-2 ($W_2$↓) and Energy Distance (ED↓) metrics, as well as the average performance across conditions. \textbf{Bold} indicates the best performing method that can perform the joint task.
We provide a visualization of this result in Appendix~\ref{ap:additional_results} \cref{fig:leftout_kos_comparison}.
}
\vspace{0.2em}
\label{tab:renge_knockout_results_full_precision}
\resizebox{\columnwidth}{!}{
\begin{tabular}{@{}l cccccccc@{}}
\toprule
 & \multicolumn{2}{c}{NANOG} & \multicolumn{2}{c}{POU5F1} & \multicolumn{2}{c}{PRDM14} & \multicolumn{2}{c}{Average} \\
\cmidrule(lr){2-3} \cmidrule(lr){4-5} \cmidrule(lr){6-7} \cmidrule(lr){8-9}
 & $W_2$↓ & ED↓ & $W_2$↓ & ED↓ & $W_2$↓ & ED↓ & $W_2$↓ & ED↓ \\
\midrule
\sfm & 5.726 $\pm$ 0.015 & 0.467 $\pm$ 0.014 & 5.931 $\pm$ 0.024 & 0.826 $\pm$ 0.020 & 6.002 $\pm$ 0.043 & 0.639 $\pm$ 0.030 & 5.886 $\pm$ 0.027 & 0.644 $\pm$ 0.021 \\
\midrule
RF & 9.859 $\pm$ 0.002 & 11.492 $\pm$ 0.006 & 10.211 $\pm$ 0.002 & 12.140 $\pm$ 0.004 & 9.889 $\pm$ 0.005 & 11.461 $\pm$ 0.013 & 9.986 $\pm$ 0.003 & 11.698 $\pm$ 0.008 \\
\methodname & \textbf{5.517 $\pm$ 0.041} & \textbf{0.818 $\pm$ 0.015} & \textbf{5.717 $\pm$ 0.021} & \textbf{1.136 $\pm$ 0.039} & \textbf{5.816 $\pm$ 0.023} & \textbf{0.973 $\pm$ 0.053} & \textbf{5.683 $\pm$ 0.028} & \textbf{0.976 $\pm$ 0.036} \\
\bottomrule
\end{tabular}
}
\end{table}

\subsection{PCMCI Adapted to Snapshot Settings}
\label{ap:pcmci}
\begin{table}[H]
\centering
\caption{ {\textbf{\methodname achieves higher AUPRC than the adapted PCMCI baseline in the snapshot 
setting.} We adapt PCMCI to the snapshot regime by constructing pseudo-longitudinal 
trajectories sampled from OT couplings between successive marginals. Results are
reported on the trifurcating (TF) and cyclical (CY) BoolODE systems, in both the
observational (obs) and observational + interventional (obs+int) settings.}}
\label{tab:pcmci_comparison}
\begin{tabular}{@{}llcc@{}}
\toprule
Model & Dataset & AUPRC observational & AUPRC obs. + interventional \\
\midrule
PCMCI       & TF & 0.29 & 0.36 \\
\methodname & TF & \best{0.60} & \best{0.92} \\
\midrule
PCMCI       & CY & 0.22 & 0.21 \\
\methodname & CY & \best{0.47} & \best{0.80} \\
\bottomrule
\end{tabular}
\end{table}

\end{document}